\documentclass{article}

\usepackage[preprint, nonatbib]{neurips_2025}

\usepackage[utf8]{inputenc} 
\usepackage[T1]{fontenc}    
\usepackage{hyperref}       
\usepackage{url}            
\usepackage{booktabs}       
\usepackage{amsfonts}       
\usepackage{nicefrac}       
\usepackage{microtype}
\usepackage{lipsum}
\usepackage{fancyhdr}  
\usepackage[numbers]{natbib}
\usepackage{graphicx}
\usepackage{subcaption}
\usepackage{algorithm}
\usepackage{algpseudocode}
\usepackage{amsmath}
\usepackage{cleveref}
\usepackage{bm}
\graphicspath{{media/}}
\usepackage{textcomp}  
\usepackage{scalerel}  
\usepackage{colortbl}
\usepackage{multirow}
\usepackage{array}
\usepackage{tabularx}
\usepackage{booktabs}
\usepackage[toc]{appendix}
\usepackage{minitoc}
\usepackage{titletoc}
\usepackage{amssymb}
\usepackage{caption}
\usepackage{subcaption}
\usepackage{makecell}
\usepackage{framed} 
\usepackage{enumitem}
\usepackage{array,multirow,graphicx}
\usepackage{float}
\usepackage{mdframed}
\usepackage{listings} 
\usepackage[table]{xcolor}
\usepackage[most]{tcolorbox/tcolorbox}
\usepackage{threeparttable}
\usepackage{etoolbox}

\makeatletter
\newcommand{\myfnsymbol}[1]{%
  \expandafter\@myfnsymbol\csname c@#1\endcsname
}
\newcommand{\@myfnsymbol}[1]{%
  \ifcase #1
  \or 1
  \or 2
  \or \TextOrMath{\textasteriskcentered}{*}
  \or \TextOrMath{\textdagger}{\dagger}
  \fi
}
\newcommand{\affiliationA}{\@myfnsymbol{1}}
\newcommand{\affiliationB}{\@myfnsymbol{2}}
\newcommand{\equalcontributor}{\@myfnsymbol{3}}
\newcommand{\correspondingA}{\@myfnsymbol{4}}
\makeatother

\definecolor{bestcolor}{RGB}{255,200,200}    
\definecolor{secondcolor}{RGB}{166,202,240} 

\newcommand{\markbest}[1]{\cellcolor{bestcolor}#1}
\newcommand{\marksecond}[1]{\cellcolor{secondcolor}#1}

\newcommand{\captionbest}[1]{{\colorbox{bestcolor}{\textcolor{black}{#1}}}}
\newcommand{\captionsecond}[1]{{\colorbox{secondcolor}{\textcolor{black}{#1}}}}

\renewcommand{\arraystretch}{1.4} 
\newcolumntype{Y}{>{\centering\arraybackslash}m{3.5cm}} 
\newcolumntype{Z}{>{\centering\arraybackslash}m{2.8cm}} 
\newcolumntype{W}{>{\centering\arraybackslash}m{3cm}}   
\newcolumntype{V}{>{\centering\arraybackslash}m{2.2cm}} 

\title{SpatialViz-Bench: A Cognitively-Grounded Benchmark for Diagnosing Spatial Visualization in MLLMs}

\author{%
  Siting Wang\textsuperscript{1,2,3}\And
  Minnan Pei\textsuperscript{1,2}\And
  Luoyang Sun\textsuperscript{1,2,3}\And
  Cheng Deng\textsuperscript{4,\correspondingA}\And
  Yuchen Li\textsuperscript{5}\And
  Kun Shao\textsuperscript{4}\And
  Zheng Tian\textsuperscript{6}\And
  Haifeng Zhang\textsuperscript{1,\correspondingA}\And
  Jun Wang\textsuperscript{7,\correspondingA}\AND
  \normalfont
  \textsuperscript{1}The Key Laboratory of Cognition and Decision Intelligence for Complex Systems\\  
  Institute of Automation, Chinese Academy of Sciences\\
  \textsuperscript{2}University of Chinese Academy of Sciences, Beijing, China \\
  \textsuperscript{3}AI Lab, The Yangtze River Delta\\
  \textsuperscript{4}Huawei Noah’s Ark, UK \\
  \textsuperscript{5}Shanghai Jiao Tong University \\
  \textsuperscript{6}	School of Creativity and Art, ShanghaiTech University\\
  \textsuperscript{7}University College London
}

\makeatletter

\pretocmd{\@maketitle}{\begingroup\renewcommand{\arraystretch}{0.9}}{}{}
\apptocmd{\@maketitle}{\endgroup}{}{}

\patchcmd{\@maketitle}
  {\hfil\linebreak[0]\hfil}
  {\hspace{0.2em}\linebreak[0]\hspace{0.2em}}
  {}{\PackageWarning{neurips}{Patch And spacing failed}}

\patchcmd{\@maketitle}
  {\hfil\linebreak[4]\hfil}
  {\hspace{0.2em}\linebreak[4]\hspace{0.2em}}
  {}{\PackageWarning{neurips}{Patch AND spacing failed}}

\patchcmd{\@maketitle}
  {{\LARGE\bf \@title\par}}
  {{\fontsize{17}{21}\selectfont\bf \@title\par}}
  {}{\PackageWarning{neurips}{Patch title fontsize failed}}

\newcommand{\neurips@tightstrut}{%
  \patchcmd{\@maketitle}{24\p@}{12\p@}{}{}%
}
\neurips@tightstrut\neurips@tightstrut\neurips@tightstrut\neurips@tightstrut

\makeatother

\begin{document}

\renewcommand{\thefootnote}{\myfnsymbol{footnote}}
\maketitle
\footnotetext[4]{Correspondence to \href{mailto:davendw49@gmail.com}{Cheng Deng}, \href{mailto:haifeng.zhang@ia.ac.cn}{Haifeng Zhang}, \href{mailto:jun.wang@cs.ucl.ac.uk}{Jun Wang}.}%

\setcounter{footnote}{0}
\renewcommand{\thefootnote}{\arabic{footnote}}

\vspace{-2em}
\begin{abstract}
Humans can imagine and manipulate visual images mentally, a capability known as \textit{spatial visualization}. 
While many multi-modal benchmarks assess reasoning on visible visual information, the ability to infer unseen relationships through spatial visualization remains insufficiently evaluated as a spatial skill. 
This reliance on publicly sourced problems from IQ tests or math competitions risks data contamination and compromises assessment reliability.
To this end, we introduce \textbf{\textit{SpatialViz-Bench}}, a comprehensive multi-modal benchmark for \textit{spatial visualization} with \emph{12} tasks across \emph{4} sub-abilities, comprising \emph{1,180} programmatically generated problems, a scalable framework that allows for expansion to ensure fair and continuously reliable evaluations. 
Our evaluation of \emph{27} Multi-modal Large Language Models (MLLMs) reveals wide performance variations, demonstrates the benchmark's strong discriminative power, and uncovers counter-intuitive findings: Chain-of-Thought (CoT) prompting paradoxically degrades accuracy on open-source models.
Through statistical and qualitative analysis of error types, SpatialViz-Bench demonstrates that state-of-the-art MLLMs exhibit deficiencies in \textit{spatial visualization} tasks, thereby addressing a significant lacuna in the field.
The benchmark data and evaluation code are publicly available. 
\footnote[1]{\begin{tabular}[t]{@{}l@{ }l@{}}
Data: & \url{https://huggingface.co/datasets/PLM-Team/Spatial-Visualization-Benchmark} \\
Code: & \url{https://github.com/wangst0181/SpatialViz-Bench}
\end{tabular}}
\end{abstract}

\section{Introduction}
\begin{figure}[t]
    \centering
    \includegraphics[width=1\linewidth]{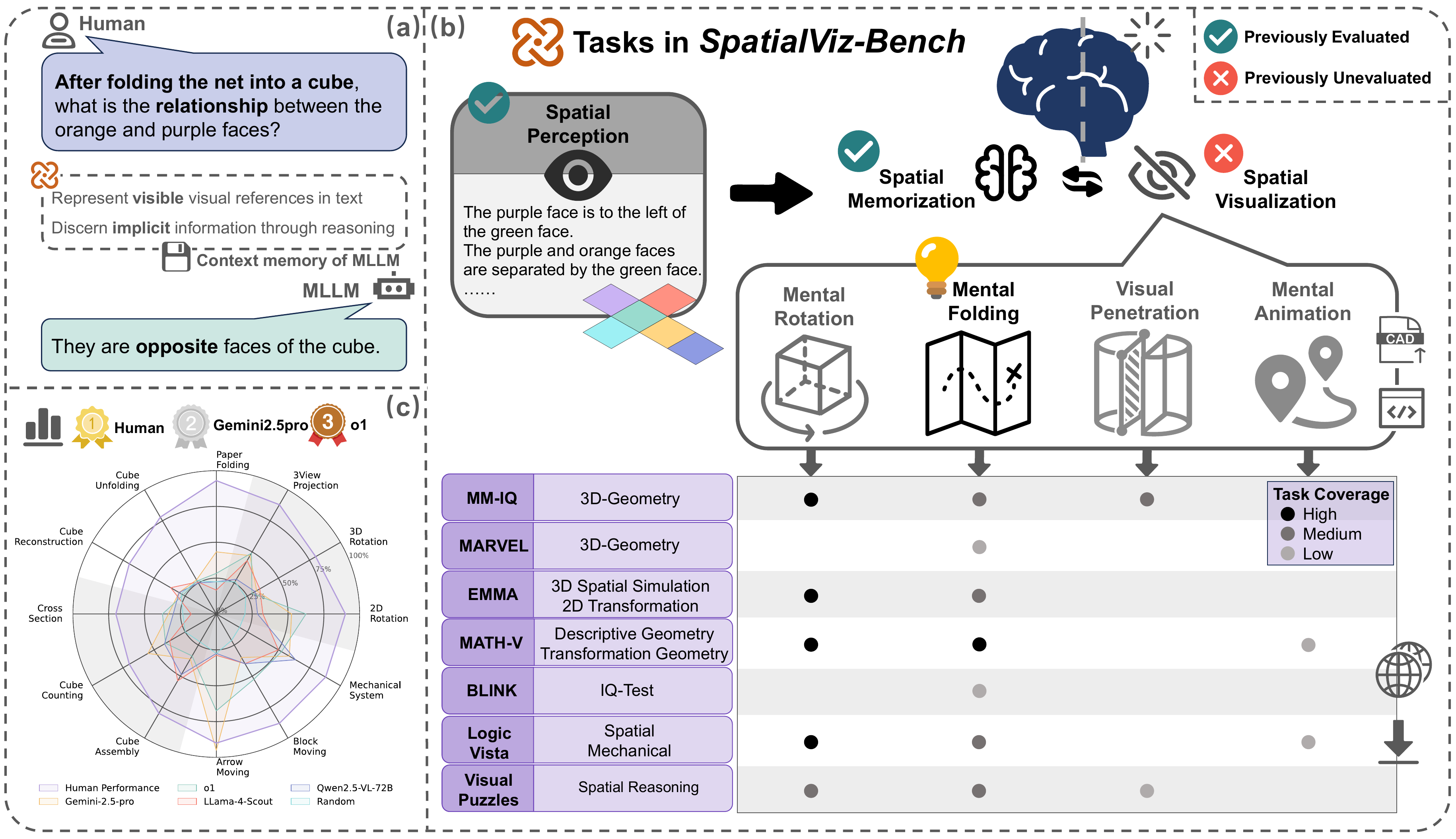}
    \caption{\textbf{Overview of SpatialViz-Bench.} (a) presents a representative task instance. (b) unfolds the reasoning behind (a): perceiving visible cues to infer unseen relationships via iterative visualization and memorization. The table highlights a systematic gap: unlike perception, \textit{spatial visualization} remains a largely unassessed blind spot in prior benchmarks (indicated by lighter colors). (c) displays zero-shot accuracy revealing significant gaps against human performance.}
    \label{fig:comp}
    \vspace{-2em}
\end{figure}

\begin{figure}[t]
    \centering
    \includegraphics[width=0.96\linewidth]{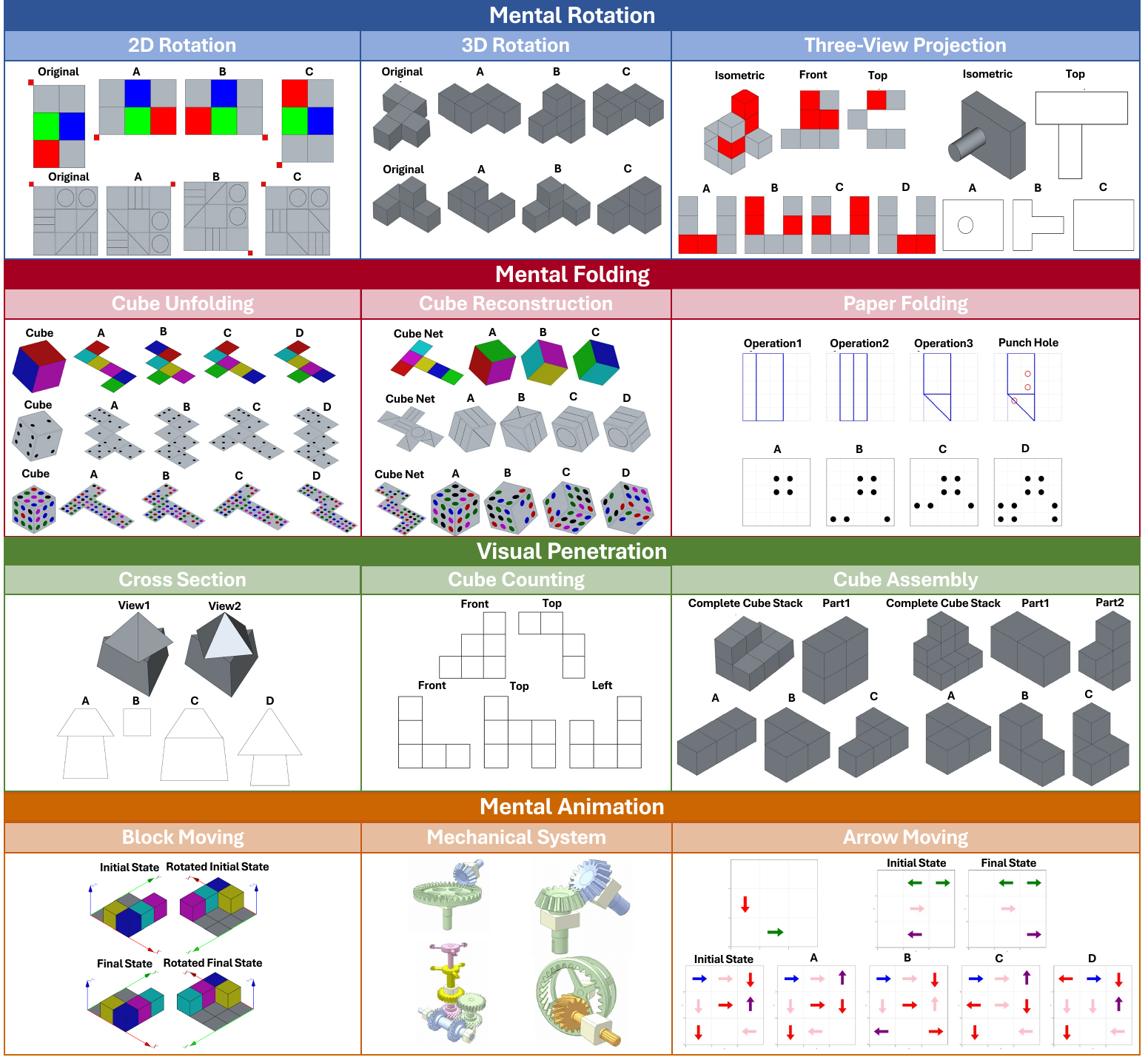}
    \caption{\textbf{Overview of Tasks in SpatialViz-Bench.} SpatialViz-Bench evaluates 4 spatial sub-abilities, mental rotation, mental folding, visual penetration, and mental animation, via 3 tasks each (12 tasks total). Each task has 2–3 difficulty levels of 40–50 cases, yielding 1,180 question–answer pairs.}
    \label{fig:Taks}
    \vspace{-1em}
\end{figure}

Large Language Models (LLMs) have demonstrated strong capabilities in complex reasoning, and the integration of Vision Transformers (ViTs) has given them ``eyes'', extending these abilities into the multimodal domain. While many tasks focus on \textit{visible} information, real-world challenges in fields like architectural design and medical-image–assisted surgery often demand the ability to mentally construct and manipulate \textit{unseen} structures, a capability in which existing MLLMs still struggle. To bridge this gap, \textit{spatial visualization} must be abstracted and assessed through targeted evaluations that isolate it from confounding factors, like a well-designed physics exam tests fundamental principles. 
However, current evaluations rely heavily on web-sourced problems, risking data leakage and inconsistent formulations, underscoring the need for a procedurally generated, standardized benchmark to ensure fair and reliable assessment.

This cognitive faculty for mental manipulation is known as \textbf{\textit{spatial visualization}}, which was first identified by Thurstone in his work on primary mental abilities~\citep{thurstone1938}. Successfully performing spatial visualization tasks relies on two other fundamental spatial abilities: \textit{Spatial perception}~\citep{thurstone1950}, which aims to perceive external spatial information and relationships, and \textit{spatial memorization}~\citep{memory}, which requires temporarily storing transformation information mentally without accessing physical objects.

Despite their importance as dedicated spatial-reasoning challenges, \textit{spatial visualization} tasks are often buried under broader categories like mathematical or logical reasoning, appearing as multimodal puzzles or 3D geometry problems. This categorization obscures the evaluation of \textit{spatial visualization} as a distinct capability and focuses on ``solving'' a problem rather than driving research toward core spatial abilities. Moreover, most examples are drawn from publicly available sources, online IQ tests, administrative exams, and math contests, which risks overlap between training and evaluation data and undermines reliability. The scarcity of items per subskill also magnifies random error, while heterogeneous formats make it hard to distinguish true reasoning failures from misinterpretation. Consequently, even with potential pretraining exposure, performance remains poor. State-of-the-art systems score just \emph{27.64} on 3D Geometry in MM-IQ~\citep{mmiq} and \emph{26.00} on Descriptive Geometry in MathVision~\citep{mathvision}.
Beyond task difficulty, the modern paradigm of pretraining on vast, scraped internet data fundamentally challenges evaluation validity~\citep{data_contamination}, a problem exacerbated by proprietary datasets that make auditing for contamination impossible. This fundamental challenge calls for a new generation of benchmarks with dynamically updatable test banks to ensure persistent evaluation integrity~\citep{surveyllmbench}. 

To address these shortcomings, we introduce \textit{\textbf{SpatialViz-Bench}}, a novel benchmark designed to formally evaluate the \textit{spatial visualization} capabilities of MLLMs, comprising a framework of \emph{4} key sub-abilities(mental rotation, mental folding, visual penetration, and mental animation) from which \emph{12} targeted tasks are designed for comprehensive assessment. Inspired by benchmarks like CLEVR~\citep{clevr}, a diagnostic benchmark for \textit{spatial perception}, which uses Blender~\citep{blender} for data generation, we developed a pipeline that integrates Python with FreeCAD~\citep{freecad} for the programmatic generation of novel test cases, enabling scalable task expansion while effectively preventing data contamination by dynamically updating the test bank through randomized generation. We employ standardized question templates to minimize errors arising from varied instructions. Furthermore, programmatic generation allows us to control task difficulty precisely and to create distractors with explanations systematically.

Models with strong \textit{spatial visualization} skills can serve as an \textbf{efficient internal world model}, providing a foundational capability for various downstream applications. This allows a model to run fast, lightweight internal “what-if” scenarios (e.g., "what happens if I rotate this object?", “if this gear turns clockwise, which way will the connected gear move?”) to predict the outcome of actions. This is far more efficient than the current alternative of invoking large, diffusion-based video generation models to explicitly render a future state.

The main contributions of our work can be listed as follows:


\begin{itemize}[leftmargin=1.2em]
\vspace{-0.6em}
\item We introduce \textit{\textbf{SpatialViz-Bench}}, the first benchmark to formally establish a comprehensive and challenging evaluation framework for \textit{spatial visualization}, a core yet long-overlooked cognitive ability. It is grounded in cognitive science and assesses \emph{4} key sub-abilities through \emph{12} distinct tasks, resulting in a total of \emph{1,180} examples across parameter-controlled difficulty levels.
\item We establish a scalable and trustworthy programmatic generation methodology for \emph{11} of our tasks. This approach not only enables continuous expansion of tasks but also sets a new standard for fair evaluation by preventing data contamination through dynamic updates to the test bank.
\item We systematically evaluate \emph{27} MLLMs, with top scores from Gemini-2.5-pro (\emph{44.66}\%) and o1 (\emph{41.36}\%). These results demonstrate the benchmark's challenge and high discriminative power, revealing a significant capability gap to human performance.
\item We conduct a diagnostic analysis revealing that model failures stem primarily from fundamental Perceptual and Spatial Transformation deficits, rather than from high-level reasoning, which offers a clear direction for future improvements.
\end{itemize}

\section{Related Works}
\textbf{Current Landscape in Spatial Reasoning Benchmarks} The evaluation of spatial reasoning in MLLMs has largely concentrated on abilities tied to directly observable information. Benchmarks for \textit{spatial perception}, the ability to identify and interpret spatial relationships from visual input, are the most established. Existing benchmarks like What’sUp~\citep{whatsup}, Blink~\citep{blink}, and SpatialRGPT-bench~\citep{spatialrgpt} assess how models understand object- or camera-centric relationships, relative distances, sizes, and positions. 
Progress has also been made in evaluating \textit{spatial memorization}, with video-based benchmarks like VCBench~\citep{vcbench} and VSI-bench~\citep{vsibench} challenging models to track objects in dynamic scenes. 
These efforts have built a foundation for assessing a type of spatial reasoning that relies on explicit visual information and applies a model's world knowledge to interpret what is perceived. However, they largely neglect the advanced capability of \textit{spatial visualization}, the ability to infer implicit visual-spatial information through transformation of structures derived from visible inputs, leaving a significant gap in the current evaluation landscape.

\textbf{Evaluation of Spatial Visualization} Evaluating \textit{spatial visualization} presents challenges regarding data contamination, obscured categorization, and narrow task coverage. A primary concern is contamination from public sources~\citep{defining}, a risk programmatic generation seeks to mitigate, as seen in the LEGO-Puzzles benchmark~\citep{lego}. Furthermore, \textit{spatial visualization} is often subject to obscured categorization, subsumed under broader domains like mathematical or logical reasoning in general benchmarks (e.g., MM-IQ~\citep{mmiq}, MathVision~\citep{mathvision}), which diverts focus from it as a core ability. Concurrently, specialized datasets exhibit narrow task coverage, focusing on single sub-skills like mental rotation (SPARE3D~\citep{spare3d}, CLEVR-MRT~\citep{clevrmrt}) or specific tasks like paper folding (SRBench~\citep{mind}). \citet{mentalmodelinglimited} also assess mental modeling, utilizing distinct organizational frameworks, such as relative spatial perspectives.

\begin{figure}[t]
    \centering
    \includegraphics[width=1\linewidth]{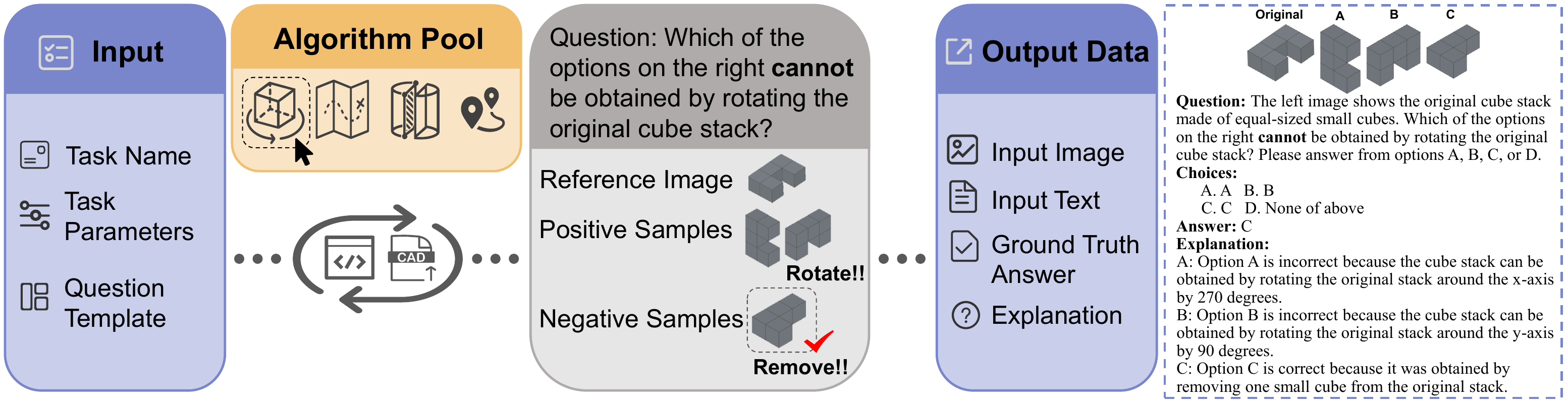}
    \caption{\textbf{The programmatic generation pipeline of a data instance.} We constructed the dataset using an programmatic generation system that integrates Python with FreeCAD, enabling precise control of difficulty, systematic generation of distractor options, and programmatic recording of explanations for incorrect choices.}
    \label{fig:DataPipeline}
    \vspace{-1.5em}
\end{figure}

\section{SpatialViz-Bench}
\subsection{Spatial Visualization}
\textbf{\textit{Spatial visualization}} is a core component of human cognitive systems and a critical capability for deployment in downstream applications. Research into this ability began with Thurstone~\citep{thurstone1938}, who defined it as performing mental operations on visual images and identified it as one of the key spatial factors: \textit{spatial perception}, \textit{spatial visualization}, and mental rotation~\citep{thurstone1950}.

Building on this foundation, we establish a cognitive framework that decomposes spatial visualization tasks into two phases: \textbf{observing visible information} and \textbf{ discerning implicit information}. The former requires basic \textit{spatial perception}, while the latter demands an alternation between \textit{\textbf{spatial visualization}} (mentally manipulating images to find implicit information) and \textit{\textbf{spatial memorization}} (temporarily storing visuospatial information)~\citep{memory}.

Our benchmark's design is guided by 4 core sub-abilities: 
1) \textbf{mental rotation}: Mentally representing and rotating objects while maintaining their features; 
2) \textbf{mental folding}: Mentally folding 2D patterns into 3D objects or vice versa~\citep{mentalfolding}; 
3) \textbf{visual penetration}: Imagining the internal structure of an object from its external features~\citep{visualpenetrative};
4) \textbf{mental animation}: Mentally visualizing the motion of components within a system~\citep{mentalanimation}.

\subsection{Overview of SpatialViz-Bench}
Stemming from an availability-driven collection, current web-sourced benchmarks containing \textit{spatial visualization} tasks lack standardization and a cognitive theory basis, resulting in inconsistent tasks and incomplete coverage. We counter this with a systematic, ability-centric methodology: we use a hierarchical framework based on cognitive principles to guide new task design and employ a unified input format with standardized templates to reduce confounds and enable fine-grained error analysis.

Based on our cognitive framework, we propose \textbf{\textit{SpatialViz-Bench}} to comprehensively evaluate the \textit{spatial visualization} capabilities of MLLMs. It is organized around \emph{4} core sub-abilities—mental rotation, mental folding, visual penetration, and mental animation—with \emph{3} assessment tasks designed for each, totaling \emph{12} tasks. Each task includes \emph{2} to \emph{3} difficulty levels, with each level containing \emph{40} or \emph{50} test cases, comprising \emph{1,180} question-answer pairs in total, mostly with image-based options to focus on visual reasoning. Further details on the dataset characteristics are provided in Appendix~\ref{characteristic}.

\begin{table}[t]
\centering
\normalsize
\caption{A Compact Summary of Spatial Reasoning Tasks.}
\label{tab:task_summary_compact}
\resizebox{\linewidth}{!}{%
    \renewcommand{\arraystretch}{1} 
    \setlength{\tabcolsep}{4pt} 
    \begin{tabular}{c c l l l}
        \toprule
        \textbf{Category} & \textbf{Task Name} & \textbf{Core Objective} & \textbf{Negative Samples} & \textbf{Difficulty Scaling} \\
        \midrule
        
        \multirow{3}{*}[-3.5ex]{\makecell[c]{\textbf{Mental}\\\textbf{Rotation}}}
        & 2D Rotation & Identify correct 2D rotation & \makecell[l]{Mirroring;\\ internal pattern rotation} & Non-centrally symmetric patterns \\
        \cmidrule(l){2-5}
        & 3D Rotation & Identify correct 3D rotation & \makecell[l]{View mirroring\\ cube removal} & Larger assemblies \\
        \cmidrule(l){2-5}
    & \makecell[c]{Three-View \\Projection} & Select left view from projections & \makecell[l]{Wrong view substitution; \\view flipping;\\ line deletion} & \makecell[l]{Real engineering parts\\ (DeepCAD~\citep{deepcad})} \\
        \midrule

        \multirow{3}{*}[-3.5ex]{\makecell[c]{\textbf{Mental}\\\textbf{Folding}}}
        & Paper Folding & Predict unfolded hole pattern & \makecell[l]{Hole mirroring, addition,\\ deletion, or relocation} & \makecell[l]{More folds;\\ larger grid;\\ more holes} \\
        \cmidrule(l){2-5}
        & Cube Unfolding & Select correct 2D net from view & \makecell[l]{Swapping face colors;\\ rotating internal patterns} & Asymmetric/dot patterns on faces \\
        \cmidrule(l){2-5}
        & Cube Reconstruction & \makecell[l]{Select 3D view from net; \\  Find opposite face} & Mirroring the correct 3D view & Follows Cube Unfolding \\
        \midrule

        \multirow{3}{*}[-3.5ex]{\makecell[c]{\textbf{Visual}\\\textbf{Penetration}}} 
        & Cross-Section & Identify cross-section of solid & Altered geometric proportions & \makecell[l]{3-solid composites;\\ oblique slicing} \\
        \cmidrule(l){2-5}
        & Cube Counting & Infer total cube count from views & Options from min/max math bounds & \makecell[l]{2 to 3 views;\\ larger assemblies} \\
        \cmidrule(l){2-5}
        & Cube Assembly & Find complementary part of split stack & Add/remove cubes from correct part & \makecell[l]{Larger stacks;\\ 3-part splits} \\
        \midrule

        \multirow{3}{*}[-3.5ex]{\makecell[c]{\textbf{Mental}\\\textbf{Animation}}} 
        & Arrow Moving & \makecell[l]{Predict final state\\ or movement sequence} & Incorrect endpoint from same start & \makecell[l]{Multiple arrows;\\ interaction rules} \\
        \cmidrule(l){2-5}
        & Block Moving & Predict final state with gravity & Incorrect final states & \makecell[l]{Higher complexity;\\ longer sequences} \\
        \cmidrule(l){2-5}
        & Mechanical System & Understand motion propagation & Incorrect motion outcomes & More system modules \\
        \bottomrule
    \end{tabular}
}
\vspace{-1em}
\end{table}

\subsection{Construction of SpatialViz-Bench}
\label{4.2}

\textit{SpatialViz-Bench} is constructed through a combination of programmatic generation and manual design. For 11 of the tasks, we used a programmatic system integrating Python with FreeCAD~\citep{freecad} (see~\autoref{fig:DataPipeline}). By explicitly utilizing cognitive load parameters rather than heuristics, such as aligning rotational complexity (global object vs. internal pattern rotation) with mental transformation steps~\citep{mentalrotation3D}, our programmatic framework ensures precise difficulty control, while employing controlled randomness to enhance diversity and generate distractor options with explanations for deep diagnostics.
Notably, the Three-View Projection task (Level 1) uses fixed DeepCAD~\citep{deepcad} models, but we programmatically generate novel distractors (e.g., random line deletion, view flipping) to ensure novelty.
Conversely, the Mechanical System task (1/12) was manually designed, as programmatic, physically-consistent generation was technically difficult. Using representative public simulations as a reference, experts designed all questions from scratch. These visual-based questions probe dynamic motion propagation (e.g., rotational dynamics from a single image), testing visual simulation rather than caption recall or theoretical derivation.

This combined methodology, leveraging both programmatic generation and the vast pool of public simulations for expert-driven question design, supports a dynamically updated test bank that proactively mitigates data contamination. A task summary is presented in~\autoref{tab:task_summary_compact}, with detailed generation processes, algorithmic pseudocode, and illustrative examples deferred to Appendix \ref{task_construction}, \ref{pseudocode} and \ref{examples}.
\section{Evaluation}
\subsection{Evaluation Setup}
\label{5.1}

\begin{table}[t]
\centering
\caption{Comparison of open-source model performances. Tasks: 2D Rotation (2DR), 3D Rotation (3DR), Three-View Projection (3VP), Paper Folding (PF), Cube Unfolding (CU), Cube Reconstruction (CR), Cross-Section (CS), Cube Counting (CC), Cube Assembly (CA), Arrow Moving (AM), Block Moving (BM), Mechanical System (MS). The \captionbest{first} and \captionsecond{second} highest accuracy of MLLMs are marked in red and blue, with open-source and closed-source models marked separately.}
\label{tab:opensource}
\resizebox{\linewidth}{!}{%
    \renewcommand{\arraystretch}{1.4}
    \begin{tabular}{l|c c|c c c c|c c c c|c c c c|c c c c}
    \toprule
    \multirow{2}{*}{\textbf{Model}} & \multicolumn{2}{c|}{\textbf{Overall}} & \multicolumn{4}{c|}{\textbf{Mental Rotation}} & \multicolumn{4}{c|}{\textbf{Mental Folding}} & \multicolumn{4}{c|}{\textbf{Visual Penetration}} & \multicolumn{4}{c}{\textbf{Mental Animation}} \\
    \cline{2-19}
     & {w/o CoT} & {w/ CoT} & {2DR} & {3DR} & {3VP} & {Avg} & {PF} & {CU} & {CR} & {Avg} & {CS} & {CC} & {CA} & {Avg} & {AM} & {BM} & {MS} & {Avg}\\
    \midrule
    Human  & - & 82.46 & 90.00 & 79.16 & 87.50 & 85.56 & 93.75 & 75.00 & 72.92 & 80.56 & 72.92 & 70.83 & 82.50 & 75.42 & 90.00 & 87.50 & 87.50 & 88.33 \\
    \midrule
    Random & - & 25.08 & 23.75 & 27.50 & 31.00 & 27.69 & 19.17 & 20.00 & 25.83 & 21.67 & 30.00 & 25.00 & 30.00 & 28.12 & 28.75 & 16.25 & 25.00 & 23.33 \\
    \midrule
    Qwen2.5-72B-Instruct(Text-only) & - & 25.86 & 15.00 & 35.00 & 15.00 & 21.67 & 23.33 & 16.67 & 26.67 & 22.22 & 20.00 & 33.33 & 45.00 & 31.25 & 25.00 & 30.00 & 30.00 & 28.33 \\

    \hline
    \multicolumn{19}{c}{\textbf{Open Source MLLMs}} \\
    
    \hline
    \multicolumn{19}{c}{3B} \\
    \hline
    SAIL-VL-1.5-2B & 29.32 & 24.15 & 22.50 & 22.50 & 22.00 & 22.31 & 20.00 & 27.50 & 20.00 & 22.50 & 24.17 & 26.67 & 32.50 & 27.19 & 21.25 & 25.00 & 27.50 & 24.58 \\
    InternVL3-2B & - & 26.19 & 16.25 & 33.75 & 31.00 & 27.31 & 22.50 & 25.83 & 25.00 & 24.44 & 20.00 & 30.83 & 30.00 & 26.56 & 18.75 & 32.50 & 30.00 & 27.08 \\
    Deepseek-VL2-tiny(3B) & 29.58 & 21.36 & 17.50 & 22.50 & 27.00 & 22.69 & 21.67 & 20.83 & 19.17 & 20.56 & 20.83 & 22.50 & 18.75 & 20.94 & 18.75 & 21.25 & 25.00 & 21.67 \\
    Qwen2.5-VL-3B-Instruct & 30.17 & 26.10 & 20.00 & 18.75 & 21.00 & 20.00 & 25.00 & 25.83 & 21.67 & 24.17 & \marksecond{25.83} & 23.33 & 30.00 & 25.94 & \markbest{35.00} & 30.00 & 42.50 & 35.83 \\
    \hline
    \multicolumn{19}{c}{7B} \\
    \hline
    Qwen2.5-VL-7B-Instruct & 30.76 & 27.97 & 25.00 & 16.25 & 29.00 & 23.85 & \markbest{34.17} & 21.67 & 30.00 & \markbest{28.61} & 16.67 & 36.67 & 28.75 & 27.19 & 22.50 & 23.75 & 51.25 & 32.50 \\
    Qwen2.5-Omni-7B & 31.44 & 27.29 & 22.50 & 20.00 & 29.00 & 24.23 & 25.00 & 27.50 & 20.00 & 24.17 & 20.83 & 33.33 & 27.50 & 27.19 & 31.25 & 30.00 & 45.00 & 35.42 \\
    SAIL-VL-1.6-8B & 29.15 & 25.00 & 18.75 & 21.25 & 25.00 & 21.92 & 28.33 & 25.00 & 18.33 & 23.89 & 21.67 & 19.17 & 23.75 & 21.25 & 25.00 & 35.00 & 45.00 & 35.00 \\
    InternVL3-8B & 30.25 & 30.08 & 20.00 & \marksecond{38.75} & 28.00 & 28.85 & 28.33 & 23.33 & 25.00 & 25.56 & 15.83 & \marksecond{40.83} & 38.75 & 30.94 & 30.00 & 30.00 & 51.25 & 37.08 \\
    \hline
    \multicolumn{19}{c}{16B} \\
    \hline
    Kimi-VL-A3B-Instruct(16B) & 32.37 & 23.90 & 16.25 & 30.00 & 36.00 & 28.08 & 25.83 & 20.00 & 26.67 & 24.17 & 21.67 & 5.00 & 28.75 & 17.19 & 15.00 & 31.25 & 37.50 & 27.92 \\
    Kimi-VL-A3B-thinking(16B) & - & 28.14 & 13.75 & 20.00 & 25.00 & 20.00 & 23.33 & 24.17 & 26.67 & 24.72 & 25.00 & 36.67 & 25.00 & 29.38 & 30.00 & \marksecond{43.75} & 47.50 & \marksecond{40.42} \\
    Deepseek-VL2-small(16B) & 25.17 & 25.17 & \marksecond{31.25} & 16.25 & 26.00 & 24.62 & 22.50 & 25.00 & 26.67 & 24.72 & 9.17 & 35.00 & 35.00 & 25.31 & 26.25 & 23.75 & 28.75 & 26.25 \\
    \hline
    \multicolumn{19}{c}{32B} \\
    \hline
    Deepseek-VL2(27B) & 30.08 & 28.31 & 25.00 & 33.75 & 30.00 & 29.62 & \marksecond{31.67} & 25.00 & 22.50 & 26.39 & 18.33 & 39.17 & 28.75 & 28.75 & 26.25 & 30.00 & 31.25 & 29.17 \\
    Qwen2.5-VL-32B-Instruct & \marksecond{33.90} & 32.12 & \marksecond{31.25} & 35.00 & 38.00 & \marksecond{35.00} & 21.67 & 25.00 & 27.50 & 24.72 & \marksecond{25.83} & 36.67 & 43.75 & 34.38 & 28.75 & 27.50 & 55.00 & 37.08 \\
    InternVL3-38B & 29.75 & 30.34 & 22.50 & 33.75 & 29.00 & 28.46 & 20.83 & \marksecond{29.17} & \marksecond{30.83} & \marksecond{26.94} & 21.67 & 32.50 & 41.25 & 30.63 & 25.00 & 30.00 & \marksecond{56.25} & 37.08 \\
    \hline
    \multicolumn{19}{c}{72B} \\
    \hline
    Qwen2.5-VL-72B-Instruct & \markbest{35.00} & \marksecond{33.31} & 28.75 & 31.25 & 28.00 & 29.23 & 22.50 & 20.00 & 30.00 & 24.17 & \markbest{30.00} & \markbest{41.67} & \marksecond{48.75} & \markbest{39.06} & 27.50 & 40.00 & \markbest{63.75} & \markbest{43.75} \\
    QvQ-72B-preview & - & 28.14 & 21.25 & 30.00 & 31.00 & 27.69 & 16.67 & 19.17 & 27.50 & 21.11 & \markbest{30.00} & 22.50 & 32.50 & 27.81 & 25.00 & \markbest{50.00} & 43.75 & 39.58 \\
    InternVL3-78B & 32.29 & 29.75 & 25.00 & 25.00 & 34.00 & 28.46 & 19.17 & 25.00 & 22.50 & 22.22 & 20.83 & 40.00 & \marksecond{48.75} & \marksecond{35.00} & 23.75 & 41.25 & 41.25 & 35.42 \\
    \hline
    \multicolumn{19}{c}{108B} \\
    \hline
    Llama-4-Maverick-17B-128E-Instruct & - & 31.78 & 20.00 & \markbest{40.00} & \marksecond{40.00} & 33.85 & 16.67 & \marksecond{29.17} & 29.17 & 25.00 & 19.17 & 35.00 & 47.50 & 32.19 & \markbest{35.00} & 40.00 & 42.50 & 39.17 \\
    LLama-4-Scout-17B-16E-Instruct & - & \markbest{34.24} & \markbest{32.50} & 35.00 & \markbest{43.00} & \markbest{37.31} & 16.67 & \markbest{32.50} & \markbest{36.67} & \markbest{28.61} & 17.50 & 37.50 & \markbest{53.75} & 34.06 & 28.75 & 40.00 & 50.00 & 39.58 \\

    \hline
    \multicolumn{19}{c}{\textbf{Closed Source MLLMs}} \\
    \hline
    
    GPT-4o & 30.76 & 31.10 & 32.50 & 27.50 & 33.00 & 31.15 & 29.17 & 15.83 & 30.00 & 25.00 & 19.17 & 40.83 & 40.00 & 32.50 & 22.50 & 32.50 & \markbest{60.00} & 38.33 \\
    o1 & - & \marksecond{41.36} & \markbest{62.50} & 28.75 & \markbest{49.00} & \markbest{46.92} & 28.33 & \markbest{34.17} & 26.67 & 29.72 & \markbest{37.50} & 40.83 & 33.75 & 37.81 & \marksecond{67.50} & \markbest{52.50} & 52.50 & \marksecond{57.50} \\
    Claude-3.5-sonnet & 26.86 & 32.54 & 31.25 & 25.00 & 45.00 & 34.62 & 20.83 & 22.50 & \marksecond{31.67} & 25.00 & 22.50 & 35.83 & \markbest{46.25} & 33.44 & 37.50 & 31.25 & 52.50 & 40.42 \\
    Claude-3.7-sonnet & - & 33.90 & 32.50 & \markbest{36.25} & 44.00 & 38.08 & 18.33 & 26.67 & 29.17 & 24.72 & 24.17 & 30.83 & \marksecond{43.75} & 31.56 & 66.25 & 28.75 & 43.75 & 46.25 \\
    Gemini-2.5-flash & - & 36.86 & 42.50 & 30.00 & 35.00 & 35.77 & 26.67 & 30.00 & \markbest{40.83} & \marksecond{32.50} & 30.00 & 38.33 & 28.75 & 32.81 & \marksecond{67.50} & 33.75 & 48.75 & 50.00 \\
    Gemini-2.5-pro & - & \markbest{44.66} & \marksecond{52.50} & 32.50 & \marksecond{47.00} & \marksecond{44.23} & \markbest{43.33} & \marksecond{31.67} & 30.00 & \markbest{35.00} & \marksecond{33.33} & \marksecond{55.00} & 36.25 & \markbest{42.19} & \markbest{95.00} & 35.00 & \marksecond{58.75} & \markbest{62.92} \\
    Doubao-1-5-vision-pro\quad\quad\quad\quad\quad\quad\quad & \markbest{37.54} & 33.31 & 7.50 & \marksecond{35.00} & 45.00 & 30.38 & \marksecond{31.67} & 23.33 & 29.17 & 28.06 & 30.00 & \markbest{55.83} & 30.00 & \marksecond{39.69} & 22.50 & \marksecond{37.50} & 47.50 & 35.83 \\
    Qwen-VL-max & \marksecond{36.10} & 32.03 & 23.75 & 26.25 & 33.00 & 28.08 & 24.17 & 17.50 & \marksecond{31.67} & 24.44 & 26.67 & 47.50 & 42.50 & 38.44 & 26.25 & 36.25 & 55.00 & 39.17 \\
    
    \bottomrule
    \end{tabular}
}
\vspace{-1em}
\end{table}

\textbf{Models} We conducted comprehensive experiments on a diverse range of MLLMs, including 8 closed-source and 19 open-source models. For \textbf{closed-source MLLMs}, we evaluated models from 5 major providers, including OpenAI series (GPT-4o~\citep{GPT-4o}, o1~\citep{o1}), Gemini series (Gemini-2.5-flash, Gemini-2.5-pro~\citep{gemini-2.5}), Claude series (Claude-3.5-sonnet~\citep{claude-3.5}, Claude-3.7-sonnet~\citep{claude-3.7}), Qwen-VL-max~\citep{qwen-vl}, and Doubao-1.5-vision-pro~\citep{doubao}. For \textbf{open-source MLLMs}, we assessed Qwen2.5-VL series~\citep{qwen2.5vl}, QvQ~\citep{qvq-72b-preview}, Qwen-Omni~\citep{qwen2.5omni}, InternVL-3 series~\citep{internvl3}, Deepseek-VL2 series~\citep{deepseekvl2}, SAIL-VL series~\citep{sail}, Kimi-VL-A3B series~\citep{kimivl} and LLama-4 series~\citep{llama4}. For \textbf{text-only LLM}, we used Qwen2.5-72B-Instruct~\citep{qwen2.5}.

\textbf{Setting} For a rigorous evaluation, all experiments were performed in a zero-shot setting~\citep{emma, mathvision}, comparing model performance under two prompting schemes: (1) CoT, where prompts were designed to encourage models to output their reasoning process before the final answer, and (2) Direct Answering (non-CoT), where prompts solicited the answer directly (see Appendix~\ref{D.1}). This methodology enabled us to not only assess the accuracy of responses but also gain deeper insights into the models' underlying reasoning mechanisms across our benchmark tasks.

\textbf{Metric Design} To evaluate models handling multimodal inputs and generating textual outputs, with most options presented as images, we formatted all tasks as Multiple-Choice Answer (MCA) with one correct answer. Option and reference images were integrated into a unified visual input. For questions where answers could be expressed as simple text, we also provided a text-based answer format (detailed in Appendix~\ref{extraction rule}). Model performance was assessed using accuracy, based on the match between predicted and ground-truth answers. This standardized approach ensures consistent evaluation across tasks and enables fair comparison of multimodal understanding across models. A comparative analysis of performance on both formats is provided in Appendix~\ref{ComparisonFormat}.

\textbf{Human Baseline} Our human baseline was established with 8 graduate students from mechanical engineering and computer science, selected for their strong spatial reasoning backgrounds. Each participant solved a 72-problem subset under strict conditions designed to be analogous to MLLM evaluation: no external aids (e.g., scratch paper) were allowed, but time was unlimited. This protocol isolates intrinsic spatial visualization abilities for a fair comparison.

\subsection{Evaluation Results}
\label{5.2}
\begin{figure}[t]
    \centering
    \includegraphics[width=1\linewidth]{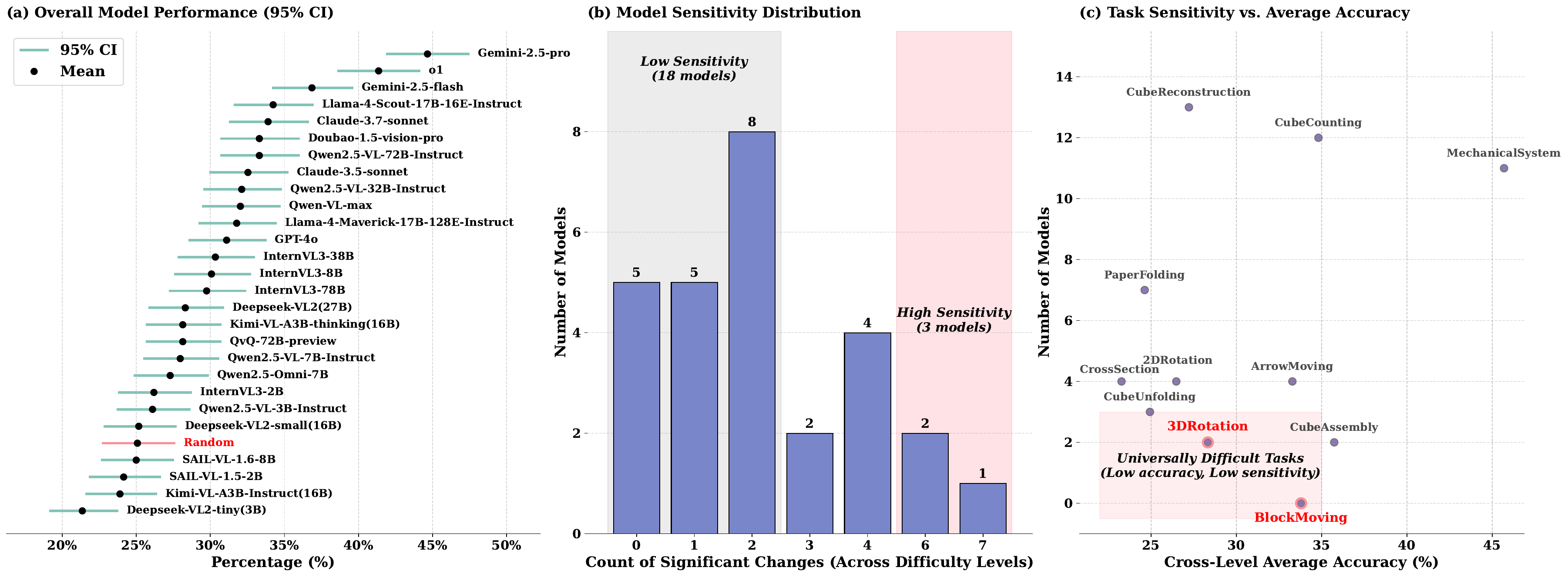}
    \caption{\textbf{Statistical Analysis of Model Performance, Difficulty Sensitivity, and Task Discriminability.} (a) presents the overall model performance with 95\% Wilson confidence intervals. (b) shows the distribution of model sensitivity to difficulty gradients. (c) provides a task-centered analysis of difficulty sensitivity, revealing how difficulty levels differentiate model capabilities across tasks.}
    \label{fig:combined_analysis}
    \vspace{-1em}
\end{figure}

This section first establishes the performance gaps between different models and then, through a CoT ablation study, investigates the impact of explicit reasoning to identify the core abilities required for advanced spatial reasoning.

\subsubsection{Main Results}
\textbf{Tasks in SpatialViz-Bench are Vision-Dependent and Reasoning-Intensive} As the textual input alone is insufficient, visual input is essential for problem-solving, making the benchmark highly vision-dependent. We empirically validated this claim by evaluating a powerful text-only LLM (Qwen2.5-72B-Instruct). As detailed in \autoref{tab:opensource}, the text-only model achieved a total accuracy of 25.86\%, which is negligibly different from the random-chance baseline (25.08\%), quantitatively proving that the visual modality is indispensable. Most options are image-based, requiring precise visual analysis rather than simple matching, thereby increasing reasoning complexity. For both humans and MLLMs, these tasks demand multi-step spatial transformations and inferences that mirror complex CoT processes.


\textbf{Performance Gaps Reveal a Statistically Validated Hierarchy of MLLMs} All evaluated models performed well below the human baseline (82.46\%), underscoring the benchmark's difficulty. Our analysis, now supported by 95\% Wilson confidence intervals (CIs) (as shown in \autoref{fig:combined_analysis}), confirms this performance hierarchy is statistically robust. The top performer, Gemini-2.5-pro (44.66\%, CI: [41.85\%, 47.51\%]), demonstrates capabilities irrefutably above the random baseline (25.08\%, CI: [22.69\%, 27.64\%]), as their CIs do not overlap. More importantly, this analysis provides solid statistical backing for the critical capability gap between proprietary and open-source models. The CI for Gemini-2.5-pro shows no overlap with that of the top open-source model, LLaMA-4-Scout (34.24\%, CI: [31.58\%, 36.99\%]), confirming this $\sim$10\% performance delta is significant. Conversely, the CIs help group statistically similar models into "performance tiers"; for example, the CIs for LLaMA-4-Scout and Qwen2.5-VL-72B-Instruct (35.00\%, CI: [30.67\%, 36.04\%]) highly overlap, making their performance statistically indistinguishable. This statistically validated discriminative power highlights significant room for improvement.

\textbf{Core 3D Visualization Tasks Reveal Common Model Failures} Models with higher overall accuracy generally perform well across individual tasks. Most models show near-random accuracy on core 3D tasks like 3D Rotation, Cube Unfolding \& Reconstruction, indicating common and severe perceptual and visualization limitations in 3D space. Both proprietary models perform well on the Arrow Moving task, with Gemini-2.5-pro even surpassing human performance, while most of open-source models perform at near-random levels. This suggests that, despite its relatively low visual complexity, the task requires advanced reasoning—such as understanding object-centered motion—which open-source models still lack. In most cases, model performance matched our expected difficulty levels, though some discrepancies with human perception offer valuable insights for refining task design and guiding future research. Additional evaluation results and task-specific analysis are provided in Appendix \ref{intra-category_comparisons}.

\textbf{Difficulty Collapse Only Visible in Top-Tier Models} We first validated our intended difficulty gradient (DG) against human performance and hypothesized models would show similar scaling. However, data reveals a widespread "performance floor" at L0; 10 models showed $\le$1 significant DG, while the top-performing Gemini-2.5-pro was most sensitive (7 DGs) (\autoref{fig:combined_analysis}.b). From a task-centric perspective (\autoref{fig:combined_analysis}.c), three tasks induced a significant DG in 11 or more models. Notably, the stark DG contrast between CubeReconstruction (12 models) and its symmetric counterpart CubeUnfolding (1 model) suggests models better reason about symmetry from unfolded views. Conversely, BlockMoving (0 DGs) proved challenging at both levels, rendering any drop statistically invisible. Critically, on 3DRotation, the only two models exhibiting a DG were the top-two performers (Gemini-2.5-pro, o1). This confirms our core claim: only top-tier models achieved non-random L0 accuracy, and thus were the only ones capable of showing a statistically significant collapse at L1.

\subsubsection{CoT Prompting Ablation Study}
\begin{table}[t]
\centering
\caption{Statistical significance analysis of CoT prompting impact ($p < 0.05$).}
\label{tab:cot-significance}
\resizebox{0.7\linewidth}{!}{
\begin{tabular}{@{}llccc@{}}
\toprule
\textbf{Model} & \textbf{Source} & \textbf{CoT Impact} & \textbf{Significant ($p < 0.05$)} & \textbf{p-value} \\
\midrule
Kimi-VL-A3B-Instruct & Open & Negative & Yes & 0.0192 \\
Deepseek-VL2-tiny & Open & Negative & Yes & 0.0463 \\
Internvl2.5-78B & Open & Negative & Yes & 0.0368 \\
Qwen2.5-Omni-7B & Open & Negative & Yes & 0.0216 \\
Sail-VL-1.6-8B & Open & Negative & Yes & 0.0479 \\
\midrule
Claude-3.5-sonnet & Closed & Positive & Yes & 0.0007 \\
\bottomrule
\end{tabular}}
\vspace{-1.1em}
\end{table}

\begin{table}[t]
\centering
\caption{\textbf{Robustness analysis of CoT performance.} (a) Performance remains stable across different CoT prompt templates. (b) The significant performance gap between CoT and non-CoT persists across extraction rules, ruling out parsing failures as the cause of performance drops.}
\label{tab:sensitivity}

\resizebox{0.48\linewidth}{!}{%
\begin{tabular}[t]{l|cc|c}
\toprule
\multicolumn{4}{c}{\textbf{(a) Sensitivity to Prompt Variations (Accuracy \%)}} \\
\midrule
\textbf{Model} & \textbf{CoT A} & \textbf{CoT B} & \textbf{$\Delta$} \\
\midrule
Qwen2.5-VL-72B & 33.31 & 31.19 & -2.12 \\
GPT-4o & 31.10 & 30.81 & -0.29 \\
Claude-3.5-sonnet & 32.54 & 28.31 & -4.23 \\
\bottomrule
\end{tabular}%
}
\hfill 
\resizebox{0.48\linewidth}{!}{%
\begin{tabular}[t]{l|cc|c}
\toprule
\multicolumn{4}{c}{\textbf{(b) Sensitivity to Extraction Rules (Acc. Drop\%)}} \\
\midrule
\textbf{Model} & \textbf{Rule A} $\downarrow$ & \textbf{Rule B} $\downarrow$ & \textbf{$\Delta$} \\
\midrule
SAIL-VL-1.5-2B & -8.22 & -7.29 & +0.93 \\
Deepseek-VL2-3B & -5.18 & -5.01 & +0.17 \\
Kimi-VL-16B & -8.47 & -9.66 & -1.19 \\
\bottomrule
\end{tabular}%
}
\vspace{-1.1em}
\end{table}

For the non-CoT evaluation, we excluded models designed for extended reasoning (e.g., o1, Gemini-2.5 series) or those unable to adhere to the format (e.g., InternVL3-2B), proceeding only with models that could reliably provide a single-letter answer (detailed in Appendix \ref{D.1}). 

Our ablation study on Chain-of-Thought (CoT) prompting confirms a "CoT paradox," a phenomenon also noted by EMMA~\citep{emma}: CoT benefits high-performing closed-source MLLMs but often paradoxically degrades their open-source counterparts. We provide new statistical validation for this. As shown in \autoref{tab:cot-significance}, the impact is significantly positive for claude-3.5-sonnet but significantly negative for several leading open-source models.

Crucially, our analysis pinpoints where this degradation occurs. The performance loss for these open-source models is not uniform but is highly concentrated in "pure-visual" spatial tasks (e.g., 3ViewProjection, 3DRotation). This strongly supports our hypothesis: for these models, the mandate to generate explanatory text (CoT) interferes with their native visual-spatial judgment, acting as a cognitive distraction rather than an aid. In contrast, top-tier closed-source models demonstrate superior resistance to this interference, likely due to specialized RL-based reasoning training, allowing them to leverage CoT effectively.

\subsubsection{Robustness to Prompting and Extraction Strategies}
To rule out the possibility that the observed CoT degradation is an artifact of specific prompt engineering or parsing failures, we conducted a sensitivity analysis in \autoref{tab:sensitivity}. First, we tested models with an alternative CoT prompt template (detailed in Appendix~\ref{D.1}). As shown in~\autoref{tab:sensitivity}(a), the performance trends remained consistent, with Qwen2.5-VL-72B still underperforming compared to its non-CoT baseline (35.00\%). Second, we compared two distinct answer extraction rules (truncated letter matching as Rule A vs. full-format regex matching as Rule B, detailed in Appendix~\ref{extraction rule}). \autoref{tab:sensitivity}(b) reveals that the discrepancy between rules is negligible ($<1.2\%$), confirming that the negative impact of CoT (ranging from -5\% to -9\%) is a genuine reasoning failure, not a parsing error.



\subsection{Error Analysis}
This section first presents a statistical error analysis across several representative models to identify common failure modes, followed by a detailed case study of Gemini-2.5-pro to illustrate its specific reasoning processes.

\subsubsection{Statistical Error Analysis }

\begin{figure}[t]
    \centering
    \includegraphics[width=1\linewidth]{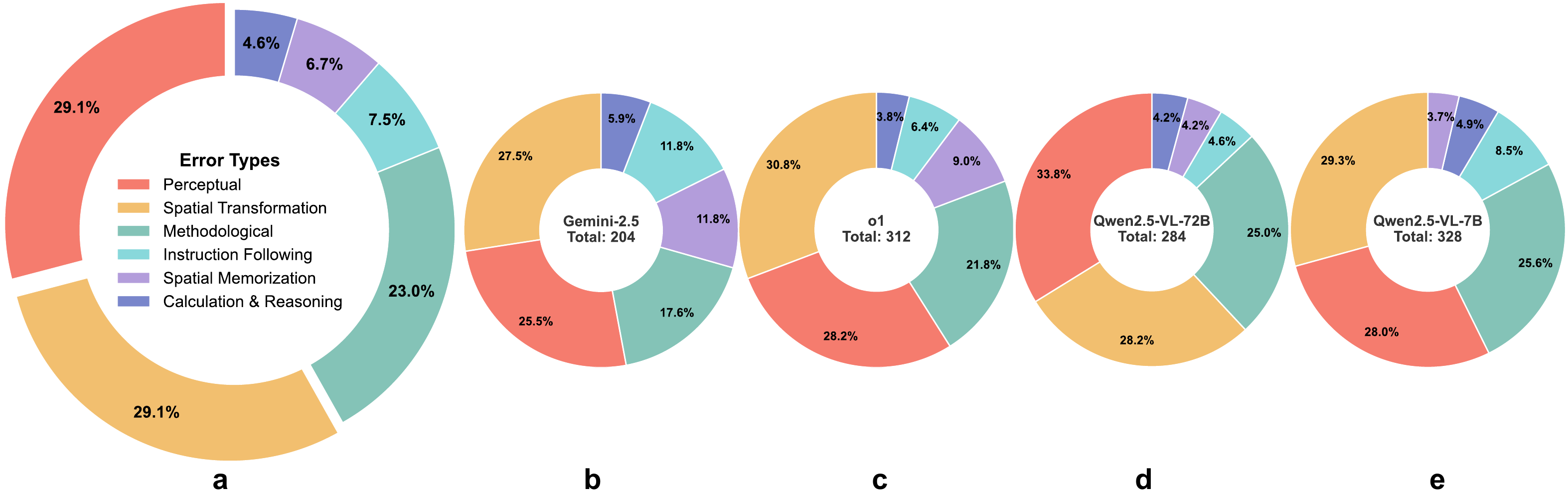}
    \caption{\textbf{Comparison of error type distributions}, with chart (a) showing the overall breakdown and charts (b-e) detailing results for specific MLLMs: (b) Gemini-2.5, (c) o1, (d) Qwen2.5-VL-72B and (e) Qwen2.5-VL-7B. Errors are classified into six categories: Perceptual, Spatial Transformation, Methodological, Instruction Following, Spatial Memorization, and Calculation \& Reasoning.}
    \label{fig:ErrorType}
    \vspace{-1em}
\end{figure}

This evaluation was conducted primarily through manual review (2 human annotators), utilizing Gemini-2.5-pro as an assistive tool based on 6 manually defined error categories, including perceptual, spatial transformation, spatial memorization, instruction following, methodological, and calculation \& reasoning error (detailed in Appendix~\ref{Error Types}). 
To account for diversity in developers, model sizes, and open/closed-source paradigms, we selected 4 models for deeper analysis: Gemini-2.5-pro and o1 (the top-performing closed-source models), Qwen2.5-VL-72B (a leading open-source model), and its smaller counterpart, Qwen2.5-VL-7B. To ensure the reliability of our error taxonomy, two annotators independently annotated a subset of 100 errors. We calculated the Cohen's Kappa coefficient ($\kappa = 0.85$), indicating strong inter-annotator agreement. Disagreements were resolved through discussion with a third expert.

\textbf{Perceptual and Spatial Transformation Errors Dominate Failures} 
The dominance of Perceptual and Spatial Transformation errors, which collectively account for nearly 60\% of all failures, quantitatively supports our central hypothesis that the primary MLLM bottleneck stems from fundamental failures in visual perception and transformation, not from high-level reasoning deficits. In contrast, the low frequency of Calculation \& Reasoning and Instruction Following errors confirms the benchmark's effective isolation of spatial deficits. Methodological errors, the third-largest category at over 23\%, indicate that models often adopt suboptimal problem-solving strategies. This highlights a clear direction for future improvements: enhancing spatial visualization capabilities by augmenting the training data with more correct solutions.

\textbf{Model Scaling Fails to Resolve Core Spatial Deficits} 
A model's absolute error count correlates with its performance rank: Gemini-2.5-pro had the fewest errors (204), followed by o1 (236), Qwen2.5-VL-72B (272), and Qwen2.5-VL-7B (328). Although top models show similar error profiles, Gemini-2.5-pro's lower rate of Methodological errors partly explains its superior performance.
The limits of model scaling become evident when comparing Qwen2.5-VL-7B and Qwen2.5-VL-72B. Despite a tenfold parameter increase, their core error patterns remained strikingly similar, with Perceptual and Transformation errors still dominant. While the 72B model nearly eliminated Spatial Memorization and Calculation errors, it made only limited gains on these most critical error types. This reveals a crucial insight: scaling alone does not resolve fundamental spatial reasoning deficits. True progress will likely require innovations in training paradigms, such as~\citep{deepseekr1}, rather than merely increasing model size.

\subsubsection{Analysis of Test Cases} 
\label{5.3}

\begin{figure}[t]
    \centering
    \includegraphics[width=1\linewidth]{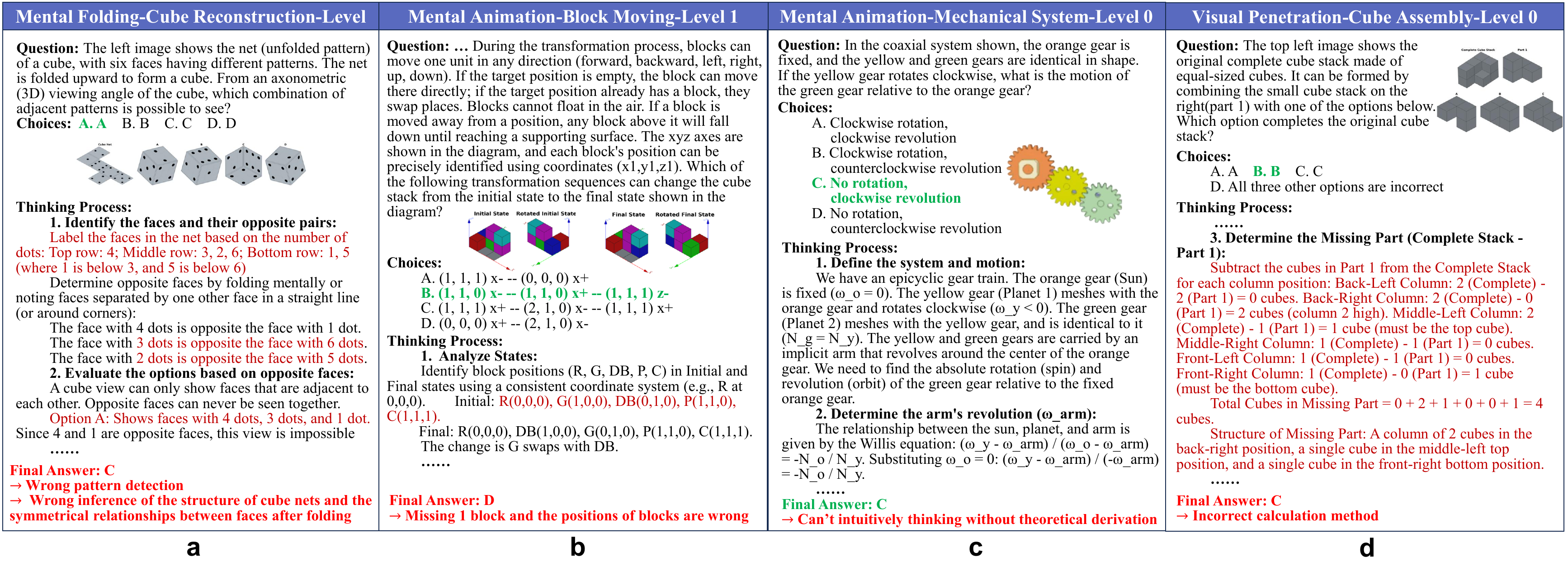}
    \caption{Case study of Gemini-2.5-pro's reasoning in different tasks.}
    \label{fig:TestCase}
    \vspace{-1em}
\end{figure}

To complement the statistical analysis, we conducted a qualitative case study of Gemini-2.5-pro's reasoning processes. The model exhibited strong reasoning, following logically coherent and complete processes, validating the effectiveness of our evaluation results. This analysis reveals a significant gap between its abstract reasoning capabilities and its visuospatial processing abilities, reinforcing that the primary bottleneck is not high-level logic but fundamental perception and visualization.

\textbf{Deficiencies Found in Both Perception and Visualization} A qualitative case study of Gemini-2.5-pro's reasoning reveals errors occur at two distinct stages: perceiving visible information and reasoning about unseen spatial relationships. In processing visible information, the model exhibited deficiencies in 2D tasks like color recognition and complex pattern identification (\autoref{fig:TestCase}.a). These perceptual failures were more pronounced in 3D space, where it struggled to accurately identify the quantity, position, and spatial relationships of stacked cubes (\autoref{fig:TestCase}.b). This difficulty is quantified by a stark performance drop, with accuracy plummeting from 95\% on the 2D Arrow Moving task to just 35\% on analogous 3D tasks. The model's primary struggles, however, emerged when reasoning about unseen information. It consistently failed tasks requiring mental manipulation, such as accurately inferring the structure of cube nets or the symmetrical relationships between faces after folding. 

\textbf{Pre-training Biases Drive Non-Simulative Problem Solving} The case study also uncovered strong pre-training biases that shape the model's problem-solving approach. For Mechanical System tasks, which were designed to be solvable via pure spatial visualization, Gemini-2.5-pro often defaulted to applying theoretical physics formulas instead of mentally simulating the motion (\autoref{fig:TestCase}.c). This behavior diverges sharply from human strategies and reveals a critical misalignment between the model's problem-solving approach and genuine spatial intelligence, suggesting its internal world model is more analytical than simulative. These qualitative examples directly illustrate the types of Methodological failures identified in our statistical analysis, forming a cohesive picture of current MLLM limitations.

\section{Conclusion}
We introduce \textbf{\textit{SpatialViz-Bench}}, a cognitive-science–inspired for testing spatial visualization in MLLMs, designed for continuous task expansion while ensuring fair evaluation by preventing data contamination via a dynamic test bank. It comprises \emph{12} tasks (\emph{1,180} problems) across \emph{4} core sub-abilities: mental rotation, mental folding, visual penetration, and mental animation. Its results show strong discriminative power, revealing the primary limitation in models is visuospatial acquisition over logical reasoning, guiding targeted optimizations in spatial skills.

\medskip

{
\small
\bibliographystyle{unsrtnat}
\bibliography{reference}
}

\newpage
\startcontents[appendix]
\section*{Appendix}
\renewcommand*\contentsname{Appendix}
\printcontents[appendix]{}{1}{\setcounter{tocdepth}{2}}

\newpage
\appendix
\section{Detailed Related Works}
\subsection{Current Landscape in Spatial Reasoning Benchmarks}
Spatial reasoning is foundational to embodied intelligence, supporting critical tasks like navigation, interaction, and scene understanding. The evaluation of this ability in MLLMs has historically focused on two primary areas: spatial perception and spatial memorization, both of which rely on interpreting directly observable, explicit visual information.

\textbf{Spatial Perception}, the ability to interpret spatial relationships from static visual input, is the most established area. Early benchmarks targeted perceptual-level understanding, such as monocular depth estimation and object localization. With the rise of MLLMs, this has shifted to visual question answering formats. For instance, datasets like VSR~\citep{vsr} and What’sUp~\citep{whatsup} benchmark models' comprehension of object-centric spatial relationships. Others, including SpatialVLM~\citep{spatialvlm}, Spatial-MM~\citep{spatialmm}, and MMRel~\citep{mmrel}, further expand this evaluation to include relative distances, camera-object perspectives, and object size comparisons. More advanced benchmarks like Blink~\citep{blink}, with its Multi-view Reasoning task, and SpatialRGPT-bench~\citep{spatialrgpt}, which incorporates world knowledge and multi-hop reasoning, have pushed the boundaries but remain centered on interpreting what is explicitly perceived.

\textbf{Spatial Memorization}, the ability to track objects and their relationships in dynamic scenes, has been increasingly addressed by video-based benchmarks. VCBench~\citep{vcbench} evaluates this through tasks like Flash Grid and 3D Navigator, which test a model’s capacity to retain 2D spatial positions and predict trajectories in 3D space. Similarly, VSI-bench~\citep{vsibench} focuses on skills essential for navigation, such as egocentric-to-allocentric transformation and perspective-shifting.

While these efforts have built a strong foundation, they predominantly assess reasoning based on explicit visual cues. They largely neglect the more advanced capability of spatial visualization—the mental manipulation of shapes and inference of implicit spatial information—leaving a significant gap in the current evaluation landscape.

\subsection{The Inadequate Evaluation of Spatial Visualization}
Despite its importance, the evaluation of spatial visualization is fraught with challenges, including obscured categorization in general benchmarks, high risk of data contamination, and a lack of diagnostic depth.

\textbf{Obscured Categorization} Spatial visualization is often not recognized as a distinct spatial skill. Instead, it is frequently subsumed under broader domains like mathematical or logical reasoning within general-purpose MLLM benchmarks. Examples are widespread: it appears as the 3D-Geometry category in MM-IQ~\citep{mmiq} and MARVEL~\citep{marvel}, the 3D Spatial Simulation category in EMMA~\citep{emma}, 3D Shapes in LogicVista~\citep{logicvista}, IQ-Test in Blink~\citep{blink}, and Descriptive/Transformation Geometry in Math-Vision~\citep{mathvision}. While VisualPuzzles~\citep{visualpuzzles} correctly situates it under spatial reasoning, this is an exception. This common miscategorization diverts focus from developing and evaluating spatial visualization as a core ability, treating it merely as a type of puzzle.

\textbf{Risk of Data Contamination} The difficulty of designing novel spatial visualization tasks means that existing benchmarks often source questions from public materials like IQ tests, administrative exams, and math contests. This practice creates a high risk of data contamination, as these materials are likely part of the massive web-scraped datasets used for pretraining MLLMs. For example, work by \citet{defining}  collects data entirely from online psychological tests. Consequently, a model's high performance on such benchmarks may not reflect true reasoning capabilities but rather memorization from the training data, compromising evaluation validity.

\textbf{Non-Diagnostic Evaluation} Current evaluations are often caught between two non-diagnostic extremes. On one hand, the heterogeneous, mixed-format questions in general benchmarks make it difficult to isolate and diagnose errors in spatial visualization specifically. On the other hand, specialized datasets are often too narrowly focused on a single sub-skill. For example, SPARE3D~\citep{spare3d} and CLEVR-MRT~\citep{clevrmrt} concentrate on mental rotation, while SRBench~\citep{mind} uses only paper folding tasks to assess the entire ability. This narrow scope fails to provide a comprehensive assessment of a model's overall spatial visualization proficiency.

In contrast to these prior works, our benchmark is designed to be systematic and diagnostic. It is structured around \emph{4} core sub-skills of spatial visualization identified in cognitive psychology, with curated tasks targeting each ability. By employing procedural generation for most tasks, our benchmark ensures greater reliability, reduces the risk of training-set overlap, and enables scalable data creation for both evaluation and future training. Furthermore, by summarizing the essential phases of spatial visualization, our framework allows for a more granular analysis to identify the root causes of reasoning errors.
\section{Data Curation Details}
\subsection{Task Construction}
\label{task_construction}
\textbf{1. Mental Rotation}

\textbf{2D Rotation Task.} A colored grid pattern with a red corner marker is rotated by $90^\circ$/$180^\circ$/$270^\circ$ to generate positive samples. Negative samples involve horizontal/vertical mirroring. We further replace symmetric color fills with non-centrally symmetric patterns. Negatives include mirror flips and internal rotations of pattern components, increasing spatial reasoning difficulty. As shown in~\autoref{alg_2DR}.

\textbf{3D Rotation Task.} A connected cube stack is rotated along x/y/z axis to form positives. Negatives are created by removing one cube or mirroring the isometric view, ensuring no simple rotation can reproduce them. Spatial complexity is increased by enlarging assembly dimensions, requiring enhanced 3D rotational reasoning. As shown in~\autoref{alg_create} and~\autoref{alg_3DR}.

\textbf{Three-View Projection Task.} This task has two categories. Firstly, given isometric, front, and top views of a connected cube stack with marked reference cubes, the task is to select the correct left view. Negatives involve altering reference cube positions or substituting the right view. We further introduce real engineering parts from the {DeepCAD dataset~\citep{deepcad}}, rendered into standard projections via FreeCAD. Negatives are crafted through random internal lines deletion, view flipping/rotation, or transformations on unseen views. As shown in~\autoref{alg_3Vcubes} and~\autoref{alg_3VCAD}.

\textbf{2. Mental Folding}

\textbf{Paper Folding Task.} A Python-based pipeline generates $m\times n$ grid patterns undergoing sequential folds (vertical/horizontal/diagonal), followed by hole-punching and unfolding. The task requires identifying the correct unfolded hole distribution. Negative samples are generated by mirroring, deleting, adding, or relocating holes to violate fold-induced symmetry. Task difficulty increases with more folds, larger grids, and denser hole placements. As shown in~\autoref{alg_paper} and~\autoref{alg_PF}.

\textbf{Cube Unfolding Task.} Given a cube with six uniquely colored faces and a view from a corner (three visible faces), the task is to select the correct 2D net (11 possibilities as shown in~\autoref{fig:CubeNets}). Positives can be crafted either by using different cube nets of the same cube or by fixing the mapping of visible faces while randomly shuffling the remaining faces. Negatives are crafted by swapping visible face colors or flipping visible-opposite face pairs. We further replace solid colors with non-centrally symmetric patterns. View angles prioritize faces with asymmetric patterns. Internal rotations of pattern components are introduced to further increase the reasoning difficulty. To push the difficulty even further, all six faces feature random colored-dot patterns on a 3×3 grid. As shown in~\autoref{alg_netr},~\autoref{alg_netu} and~\autoref{alg_CU}.

\textbf{Cube Reconstruction Task.} Cubes have six uniquely colored faces. Two task variants exist: (1) select the correct vertex view of a cube when given its net pattern, with negative samples created by mirroring the correct view; (2) identify the color of a face opposite to a given colored face. Difficulty progression follows the cube unfolding tasks. As shown in~\autoref{alg_netr} and~\autoref{alg_CR}.

\textbf{3. Visual Penetration}

\textbf{Cross-Section Task.} Nine basic geometric solids (e.g., triangular/rectangular/circular prisms/pyramids/frustums) are combined in pairs with conical shapes on top. Cross-sections are generated by slicing the composite shapes using planes parallel to the XY/YZ/XZ planes. Negative samples are constructed by adjusting the relative geometric proportions within the composite. Task complexity is increased by introducing composites with three solids, which often produce disconnected cross-sections that demand enhanced visual reasoning. Additional complexity is introduced by generating oblique cross-sections at $45^\circ$/$135^\circ$. As shown in~\autoref{alg_CS}.

\textbf{Cube Counting Task.} The task requires inferring the total cube count of a connected cube stack based on two orthogonal projection views. The minimum and maximum counts are mathematically derived to guide the construction of answer options. Constraints increase to three orthogonal projection views, reducing the number of possible solutions while increasing view integration complexity. Task difficulty further increases by expanding the spatial dimensions of the cubic assemblies. As shown in~\autoref{alg_create} and~\autoref{alg_CC}.

\textbf{Cube Assembly Task.} A pyramid-like cube stack is split into two connected parts. Tasks require identifying the complementary piece that fits the reference part. Negative samples are generated by modifying the correct piece through the addition or removal of cubic units. The difficulty is further increased by enlarging the spatial dimensions and dividing the structure into three parts instead of two. As shown in~\autoref{alg_split} and~\autoref{alg_CA}.

\textbf{4. Mental Animation}

\textbf{Arrow Moving Task.} For the easy version, an arrow with random initial position and orientation in a 3×3 grid operates by ego-centric rules: movement occurs in 4 directions (forward/backward/left/right), with "forward" always indicating the arrow's current orientation. The arrow reorients to the movement direction after each movement. Valid operation sequences are algorithmically generated; negative samples share the same initial state but yield incorrect endpoints. For the hard version, multiple colored arrows are introduced with extended rules: empty positions allow direct entry; occupied positions trigger object exchanges while maintaining Level 0 movement principles. Tasks include predicting final states from sequences, or inferring correct sequences from state pairs. As shown in~\autoref{alg_arrowpath},~\autoref{alg_arrowmap},~\autoref{alg_AM0} and~\autoref{alg_AM1}.

\textbf{Block Moving Task.} Colored cube stack combines directional movement with gravity simulation. Cubes move along six directions with unsupported cubes falling until reaching support and swapping positions as same as Arrow Moving Task. Increased spatial complexity and longer sequences elevate reasoning difficulty. As shown in~\autoref{alg_block} and~\autoref{alg_BM}.

\textbf{Mechanical System Task.} We use open-source mechanical system simulations, classifying complexity by module quantity and designing appropriate questions. These tasks assess advanced mental animation abilities, particularly to understand how the motion of one component affects others.

\subsection{Programmatic Data Generation Pipeline}
\begin{figure}[t]
    \centering
    \includegraphics[width=0.9\linewidth]{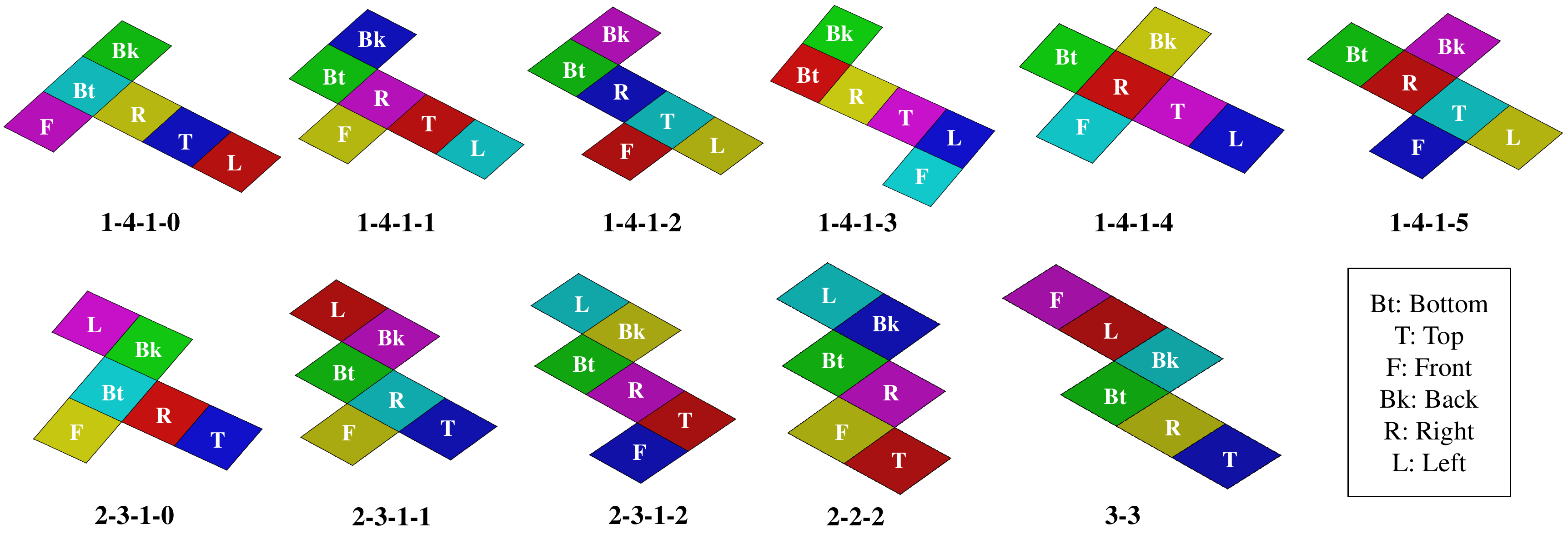}
    \caption{The eleven unfolded patterns of a cube with their corresponding numbered names. Assuming the square in row 1, position 0 represents the bottom face, and position 1 represents the right face, the corresponding arrangement of the remaining faces can be determined, facilitating the rotation of the cube.}
    \label{fig:CubeNets}
\end{figure}

FreeCAD, an open-source Computer-Aided Design (CAD) software, provides deep integration with Python programming language, enabling parametric model construction through programming. We leveraged the synergy between FreeCAD and Python to successfully automate the generation of 9 spatial visualization tasks: 2DRotation, 3DRotation, 3ViewProjection, CubeFolding, CubeReconstruction, CrossSection, CubeCounting, CubeAssembly, and BlockMoving. Additionally, two tasks—PaperFolding and ArrowMoving—were implemented solely using Python. For the MechanicalSystem task, due to its complexity and specific requirements, we employed precise manual design methods. To supplement the task overview presented in Section \ref{4.2}, the following sections provide detailed pseudocode for each programmatically generated task, offering more systematic and in-depth technical insights.

\textbf{Mental Rotation Tasks.} \autoref{alg_2DR} presents the pseudocode for the 2D Rotation Task. For the 3D Rotation Task, Three-View Projection Task, Cube Counting Task, and Block Moving Task, we need to construct connected cube stacks, with the core functions detailed in \autoref{alg_create}. \autoref{alg_3DR} demonstrates the complete implementation process of the 3D Rotation Task. The method for generating three-view projections of marked cube stacks is elaborated in \autoref{alg_3Vcubes}. \autoref{alg_3VCAD} describes the process of importing models from the DeepCAD dataset and generating their three-view projections.

\textbf{Mental Folding Tasks.} 
\autoref{alg_paper} implements a Paper class for simulating the dynamic processes of paper folding, holes punching, and unfolding. Based on this simulation framework, \autoref{alg_PF} constructs the data for the Paper Folding Task. \autoref{alg_netr} presents the core functions for transforming 11 standard cube nets (as shown in \autoref{fig:CubeNets}) into three-dimensional cubes. Utilizing these transformation functions, while \autoref{alg_netu} demonstrates how different unfolding patterns can produce the same cube.\autoref{alg_CU} and \autoref{alg_CR} provide the complete pseudocode implementations for the Cube Unfolding Task and Cube Reconstruction Task, respectively.

\textbf{Visual Penetration Tasks.} \autoref{alg_CS} details the implementation pseudocode for the Cross-Section Task. \autoref{alg_CC} comprehensively presents the data generation procedure as well as the mathematical calculation process to guide the construction of answer options in the Cube Counting Task. \autoref{alg_split} contains the core functions for decomposing a complete cube stack into multiple connected parts. Building upon these functions, \autoref{alg_CA} provides the complete construction pseudocode for the Cube Assembly Task.

\textbf{Mental Animation Tasks.} \autoref{alg_arrowpath} implements an ArrowPath class for simulating the movement process of an arrow centered on itself. \autoref{alg_arrowmap} implements an ArrowMap class that inherits from the ArrowPath class, designed to simulate movement and exchange operations in multi-arrow environments. Based on the ArrowPath class, \autoref{alg_AM0} details the data construction process for the single-arrow version of the Arrow Moving Task. Correspondingly, using the ArrowMap class, \autoref{alg_AM1} elucidates the data construction process for the multi-arrow version of the Arrow Moving Task. \autoref{alg_block} implements a Block class for simulating the movement and exchange processes of blocks that follow gravitational rules. Building upon this Block class, \autoref{alg_BM} presents the complete pseudocode implementation of the Block Moving Task.

\subsection{Manul Design for Mechanical System Task}
\label{manual}
To ensure the objectivity and quality of the Mechanical System task, we first collected simulation materials from open-source platforms. The question-answer pairs were designed by members of the author team, who strictly followed a standardized template based on the observable and deterministic animations (e.g., "If component A rotates clockwise, how does component B move?"). This structured process was designed to minimize subjectivity and focus the evaluation specifically on a model's ability to infer causal dynamics from visual input. To verify the accuracy of these question-answer pairs, we recruited two graduate student annotators from our research group, who received compensation for their contributions. They first performed independent reviews of each sample and then discussed their findings to resolve any discrepancies and reach a final consensus. This rigorous process ultimately produced 80 validated data samples.

\clearpage
\subsection{Pseudocode}
\label{pseudocode}
\begin{algorithm}[ht]
\caption{2D Rotation Task}
\label{alg_2DR}
\begin{algorithmic}[1]
\State \textbf{Input:} Color(Pattern) set $C$, grid size $(H, W)$, unit length $s$, marker length $s'$, task mode $m$
\State Initialize binary matrix $M \in \{0,1\}^{H \times W}$ with random values
\State Initialize empty lists $positive\_samples$, $negative\_samples$
\Function{DrawGridWithMarker}{$M,\ C,\ H,\ W,\ s,\ s',\ record=list()$}
    \For{$i \gets 0$ to $H{-}1$}
        \For{$j \gets 0$ to $W{-}1$}
            \State $pos \gets (j \cdot s,\ (H-1-i) \cdot s,\ 0)$
            \State $square\gets$\texttt{FreeCAD.makePlane}$(s,\ s,\ (pos,0^\circ))$
            \If{$M[i][j] = 1$}
                \If{record is empty}:
                    \State Randomly select $c \in C$ and assign $c$ to $square$ at $pos$
                    \State Append $c$ to $record$
                \Else 
                    \State Assign \texttt{rotate}$(\texttt{Pop}(record,0),\ 90^\circ)$ to $square$ at $pos$
                \EndIf
            \EndIf
        \EndFor
    \EndFor
    \State Randomly select $corner \in$ \{``top\_left'', ``top\_right'', ``bottom\_left'', ``bottom\_right''\}
    \State $pos_{\text{marker}} \gets$ \texttt{get\_marker\_pos}$(H,\ W,\ s,\ s',\ corner)$
    \State \texttt{FreeCAD.makePlane}$(s',\ s',\ (pos_{\text{marker}},0^\circ))$ with red color
    \State $img \gets$ \texttt{FreeCAD.saveImage}()
    \State \Return $img$, $record$
\EndFunction
\vspace{0.5em}
\State $ref\_img, record \gets$ \texttt{DrawGridWithMarker}$(M,C,H, W,s,s')$
\If{$m = \text{``pattern''}$}
    \State $transform\_image, record \gets$ \texttt{DrawGridWithMarker}$(M,C,H, W,s,s',record)$
    \State Append $transform\_img$ to $negative\_samples$
\EndIf
\For{$angle \in \{90^\circ, 180^\circ, 270^\circ\}$}
    \State $img \gets$ \texttt{rotate}$(ref\_img, angle)$
    \State Append $img$ to $positive\_samples$
\EndFor
\For{$flip\_dir \in \{$``horizontal'', ``vertical''\}}
    \State $img \gets$ \texttt{flip}$(ref\_img, flip\_dir)$
    \State Append $img$ to $negative\_samples$
\EndFor
\State $samples\gets(positive\_samples, negative\_samples)$ 
\State Shuffle $samples$ to assign $[A,B,C,D]$ and record $answer\_id$
\State $data \gets$ \texttt{create\_data}$(ref\_img, samples, question, answer\_id)$
\end{algorithmic}
\end{algorithm}

\clearpage
\begin{algorithm}[ht]
\caption{Fucntions for Creating Cubes with None-isolated Regions}
\label{alg_create}
\begin{algorithmic}[1]
\State \textbf{Input:} Spatial size $(X, Y, Z)$, cube size $s$
\State Initialize zero value 3D tensors $placement \in \{0\}^{Z \times Y \times X}$, empty list $cubes$
\Function{CreateCube}{$x,\ y,\ z$}
    \State $cube\gets$\texttt{FreeCAD.makebox}$(s, s, s, (x,y,z))$ and append $cube$ to $cubes$
    \State $placement[z][y][x] \gets 1$
\EndFunction

\vspace{0.5em}

\Function{CreateCubes}{$X,\ Y, \ Z$}
    \For{$z \gets 0$ to $Z{-}1$}
        \For{$y \gets 0$ to $Y{-}1$}
            \For{$x \gets 0$ to $X{-}1$}
                \If{$z = 0$ or $placement\_space[z{-}1][y][x] = 1$}
                    \State With $50\%$ probability \texttt{CreateCube}$(x,y,z)$
                \EndIf
            \EndFor
        \EndFor
    \EndFor
\EndFunction

\vspace{0.5em}

\Function{ConnectIsolatedCubes}{$X,\ Y$}
    \State $cubes_{xy} \gets \{(x,y) \mid placement[0][y][x] = 1\}$
    \State Initialize empty set $visited$, empty list $regions$
    \State $directions \gets$ [(-1,0),(1,0),(0,-1),(0,1),(-1,-1),(-1,1),(1,-1),(1,1)]
    \ForAll{$(x,y) \in cubes_{xy}$}
        \If{$(x,y) \notin visited$}
            \State Initialize empty list $region$, empty queue $queue$
            \State Add $(x,y)$ to $visited$, add $(x,y)$ to $queue$
            \While{$queue$ is not empty}
                \State $(cx, cy) \gets \texttt{popLeft}(queue)$
                \State Append $(cx, cy)$ to $region$
                \ForAll{$(dx, dy) \in directions$}
                    \State $(nx, ny) \gets (cx + dx, cy + dy)$
                    \If{$0 \le nx < X$ and $0 \le ny < Y$ and $(nx, ny) \notin visited$ \newline \hspace*{8em} and $placement[0][ny][nx] = 1$} 
                        \State Add $(nx, ny)$ to $visited$, add $(nx, ny)$ to $queue$
                    \EndIf
                \EndFor
            \EndWhile
            \State Append $region$ to $regions$
        \EndIf
    \EndFor
    \If{$|regions| > 1$}
        \For{$i \gets 0$ to $|regions|-2$}
            \State Find $(x_1, y_1), (x_2, y_2)$ with min $L_1$ distance between $regions[i]$ and $regions[i+1]$
            \State $x \gets x_1,\ y \gets y_1$
            \While{$(x \ne x_2)$ or $(y \ne y_2)$}
                \If{$x \ne x_2$ and $y \ne y_2$} \State $x \gets x \pm 1,\ y \gets y \pm 1$
                \ElsIf{$x \ne x_2$} \State $x \gets x \pm 1$
                \ElsIf{$y \ne y_2$} \State $y \gets y \pm 1$
                \EndIf
                \If{$placement\_space[0][y][x] = 0$} 
                    \State \texttt{CreateCube}$(placement,x, y,0)$
                \EndIf
            \EndWhile
        \EndFor
    \EndIf
\EndFunction   
\end{algorithmic}
\end{algorithm}

\clearpage
\begin{algorithm}[ht]
\caption{3D Rotation Task}
\label{alg_3DR}
\begin{algorithmic}[1]
\State \textbf{Input:} Spatial size $(X, Y, Z)$, cube size $s$
\State Initialize zero value 3D tensors $placement \in \{0\}^{Z \times Y \times X}$, empty list $cubes$
\State Initialize empty lists $positive\_samples$, $negative\_samples$

\State Update $placement, cubes$ with \texttt{CreateCubes}$(X,\ Y,\ Z)$
\State Update $placement, cubes$ with \texttt{ConnectIsolatedCubes}$(X,\ Y)$
\State $ref\_img \gets\texttt{FreeCAD.saveImage}(cubes)$

\vspace{0.5em}
\For{$i \gets 1$ to $4$}
    \State Randomly select $axis \in \{x, y, z\}$ and $angle \in \{90^\circ, 180^\circ, 270^\circ\}$
    \State $rotated\_cubes \gets$ \texttt{rotate}$(cubes,\ axis,\ angle)$
    \State $rotated\_img\gets \texttt{FreeCAD.saveImage}(rotated\_cubes)$
    \State Append $rotated\_img$ to $positive\_samples$
\EndFor
\vspace{0.5em}
\State $cubes'\gets$Randomly remove a cube from $cubes$ and rotate the left cubes as above
\State $rotated\_removed\_img\gets\texttt{FreeCAD.saveImage}(cubes')$
\State Append $rotated\_removed\_img$ to $negative\_samples$
\vspace{0.5em}
\For{$flip\_dir \in \{$``horizontal'', ``vertical''\}}
    \State Randomly choose $sample$ from $positive\_samples$ 
    \State $img \gets$ \texttt{flip}$(sample, flip\_dir)$
    \State Append $img$ to $negative\_samples$
\EndFor
\vspace{0.5em}
\State $samples\gets(positive\_samples, negative\_samples)$
\State Shuffle $samples$ to assign $[A,B,C,D]$ and record $answer\_id$
\State $data \gets$ \texttt{create\_data}$(ref\_img, samples, question, answer\_id)$
\end{algorithmic}
\end{algorithm}

\clearpage
\begin{algorithm}[ht]
\caption{Three-View Projection Task with Marked Cube Stack}
\label{alg_3Vcubes}
\begin{algorithmic}[1]
\State \textbf{Input:} Spatial size $(X, Y, Z)$, cube size $s$
\State Initialize zero value 3D tensors $placement \in \{0\}^{Z \times Y \times X}$, empty list $cubes$
\State Initialize empty lists $positive\_samples$, $negative\_samples$
\State Update $placement, cubes$ with \texttt{CreateCubes}$(X,\ Y,\ Z)$
\State Update $placement, cubes$ with \texttt{ConnectIsolatedCubes}$(X,\ Y)$

\vspace{0.5em}
\Function{ColorVisibleFaces}{$X,\ Y,\ Z,\ colored\_num$}
    \State $cubes\gets$ Find cubes that can be seen from front or top or left view
    \State Randomly color \texttt{min}$(colored\_num, |cubes|)$ cubes in red
\EndFunction
\vspace{0.5em}
\Function{SaveViews}{$cubes$}
    \State Initialize empty list $views$
    \ForAll{$view \in \{\text{``Isometric'', ``Top'', ``Front'', ``Left''}\}$}
        \State $img\gets$\texttt{FreeCAD.saveView}$(view)$ and append $img$ to $views$
    \EndFor
    \State \Return $views$
\EndFunction
\vspace{0.5em}
\State Update $cubes$ with \texttt{ColorVisibleFaces}$(X, Y, Z, colored\_num)$
\State $views\gets$ \texttt{SaveViews($cubes$)}
\State Select $left\_view$ from $views$ to $positive\_samples$ 
\State Select $right\_view$ from $views$ to $negative\_samples$
\vspace{0.5em}
\State Cleaer all colors and update $cubus$ with \texttt{ColorVisibleFaces}$(X, Y, Z, colored\_num)$ as above
\State $new\_views\gets$ \texttt{SaveViews($cubes$)}
\State Select $left\_view$ and $right\_view$ from $new\_views$ to $negative\_samples$
\vspace{0.5em}
\State $samples\gets(positive\_samples, negative\_samples)$
\State Shuffle $samples$ to assign $[A,B,C,D]$ and record $answer\_id$
\State $ref\_img\gets (isometric\_view,\ top\_view,\ front\_view)$
\State $data \gets$ \texttt{create\_data}$(ref\_img, samples, question, answer\_id)$
\end{algorithmic}
\end{algorithm}

\clearpage
\begin{algorithm}[ht]
\caption{Three-View Projection Task with Models from DeepCAD Datasets}
\label{alg_3VCAD}
\begin{algorithmic}[1]
\State \textbf{Input:} step file path $pth$
\State Initialize empty lists $positive\_samples$, $negative\_samples$
\State $shape\gets$\texttt{Open}$(pth)$
\State $views\gets$ \texttt{SaveViews($shape$)}
\Function{CreateIncorrectView}{$view,\ mode$}
    \If{$mode=0$}
        \State $img'\gets$Extract all internal lines and randomly delete 1 line
    \ElsIf{$mode=1$}
        \State $img'\gets$\texttt{rotate}$(view, 90^\circ)$
    \ElsIf{$mode=2$}
        \State $img'\gets\texttt{flip}(view, \text{``horizontal'' or ``vertical'})$
    \EndIf
    \State \Return img'
\EndFunction
\vspace{0.5em}
\State $ref\_view\gets$Choose view from $views$ with max area
\State $(questioned\_view, other\_view)\gets$ Randomly assign $views$ except for $ref\_view$ 
\State Append $questioned\_view$ to $positive\_samples$
\For{$mode\gets$ 0 to 2}
    \State $incorrect\_view\gets$\texttt{CreateIncorrectView}($questioned\_view$ or $other\_view$, $mode$)
    \State Append $incorrect\_view$ to $negative\_samples$
\EndFor
\vspace{0.5em}
\State $samples\gets(positive\_samples, negative\_samples)$
\State Shuffle $samples$ to assign $[A,B,C,D]$ and record $answer\_id$
\State $ref\_img\gets (isometric\_view,\ top\_view,\ front\_view)$
\State $data \gets$ \texttt{create\_data}$(ref\_img, samples, question, answer\_id)$
\end{algorithmic}
\end{algorithm}

\clearpage
\begin{algorithm}[ht]
\caption{Simulation for Paper Folding, Punching  and Unfolding}
\label{alg_paper}
\begin{algorithmic}[1]
\State \textbf{Class} Paper
\State \textbf{Attributes:} 
\State $\quad grid$, $complete\_grid$: 2D arrays representing current and complete paper states
\State $\quad original\_rows$, $original\_cols$: initial dimensions
\State $\quad current\_rows$, $current\_cols$: current dimensions after folding
\State $\quad folds$: list of fold operations

\Function{Fold}{$direction$, $line$ or $diagonal\_points$}
\If{$direction$ is horizontal}
    \State Calculate folded area
    \State Update $complete\_grid$ by marking folded area as -1
    \State Create new grid with updated dimensions
\ElsIf{$direction$ is vertical}
    \State Similar to horizontal but for columns
\ElsIf{$direction$ is diagonal}
    \State Calculate diagonal line equation
    \State Mark appropriate triangular area as -1
\EndIf
\State Record fold operation in $folds$
\EndFunction
\vspace{0.5em}
\Function{Punch}{$points$}
\For{each $(x,y)$ in $points$}
    \State Set $grid[x][y] \gets 1$
    \State Set corresponding $complete\_grid$ position to 1
\EndFor
\State Record punch operation in $folds$
\EndFunction
\vspace{0.5em}
\Function{Unfold}{}
\For{each $fold$ in reverse $folds$}
    \If{$fold$ is horizontal}
        \State Mirror grid about fold line
    \ElsIf{$fold$ is vertical}
        \State Mirror grid about fold line
    \ElsIf{$fold$ is diagonal}
        \State Mirror grid about diagonal line
    \EndIf
    \State Update current dimensions of paper
\EndFor
\State Clear $folds$ list
\EndFunction
\vspace{0.5em}

\Function{CreateIncorrectView}{$mode$}
    \State Create incorrect variant by:
    \If{$mode=\text{``row''}$}
        \State Either remove a row of holes, add extra row, or swap rows
    \ElsIf{$mode=\text{``col''}$}
        \State Either remove a column of holes, add extra column, or swap columns
    \Else
        \State Combine row and column errors
    \EndIf
    \State Update $paper$ with above changes
   
\EndFunction

\end{algorithmic}
\end{algorithm}

\clearpage
\begin{algorithm}[ht]
\caption{Paper Folding Task}
\label{alg_PF}
\begin{algorithmic}[1]
\State \textbf{Input:} Dimensions of paper $(rows, cols)$, number of folds $steps$, number of holes $punches$
\State Initialize $paper$ with dimensions $rows \times cols$
\State Initialize empty lists $ref\_imgs$, $positive\_samples$, $negative\_samples$
\For{$step \gets1$ to $steps$}
    \If{$step = steps$}
        \State $direction \gets \text{``diagonal''}$
    \Else
        \State $direction \gets \text{Randomly select $direction\in$ [``horizontal'',``vertical'']}$
    \EndIf
    \If{$direction = \text{``horizontal''}$}
        \State $line \gets \texttt{randomInt}(1, paper.current\_rows-1)$
        \State $paper.\texttt{Fold}(direction, line)$
    \ElsIf{$direction = \text{``vertical''}$}
        \State $line \gets \texttt{randomInt}(1, paper.current\_cols-1)$
        \State $paper.\texttt{Fold}(direction, line)$
    \ElsIf{$direction = \text{``diagonal''}$}
        \State $diagonal\_points \gets \text{Randomly select one set of 45-degree line endpoints}$
        \State $paper.\texttt{Fold}(direction, diagonal\_points)$
    \EndIf
    \State $img\gets$\texttt{draw\_paper}$(paper)$ and append $img$ to $ref\_imgs$
\EndFor
\vspace{0.5em}
\State $points \gets \text{Randomly select } punches \text{ zero positions}$
\State $paper.\texttt{Punch}(points)$
\State $img\gets$\texttt{draw\_paper}$(paper)$ and append $img$ to $ref\_imgs$
\vspace{0.5em}
\State $paper.\texttt{Unfold}()$
\State $img\gets$\texttt{draw\_paper}$(paper)$ and append $img$ to $positive\_samples$
\vspace{0.5em}
\State Initialize $paper'$ with same dimensions as $paper$
\State $paper'.grid\gets paper.grid$ to copy the state of unfolded paper
\State Determine the incorrect view $mode$
\For{$i\gets$1 to 3}
    \State Update $paper'$ with $paper'.\texttt{CreateIncorrectView}(mode)$
    \State $img\gets$\texttt{draw\_paper}$(paper')$ and append $img$ to $negative\_samples$
\EndFor
\vspace{0.5em}
\State $samples\gets(positive\_samples, negative\_samples)$
\State Shuffle $samples$ to assign $[A,B,C,D]$ and record $answer\_id$
\State $data \gets$ \texttt{create\_data}$(ref\_imgs, samples, question, answer\_id)$
\end{algorithmic}
\end{algorithm}

\clearpage
\begin{algorithm}[ht]
\caption{Functions for Reconstruting Cube from 11 Kinds of Cube Nets}
\label{alg_netr}
\begin{algorithmic}[1]
\State \textbf{Input: } cube size $s$
\State Define rotation operators:
\State $\quad R_x(\theta)$: Rotation about X-axis by $\theta$ degrees
\State $\quad R_y(\theta)$: Rotation about Y-axis by $\theta$ degrees
\State $\quad R_z(\theta)$: Rotation about Z-axis by $\theta$ degrees

\Function{Net2Cube}{$plane\_name,\ map,\ view,\ rot$}
    \State Initialize placement dictionary $planes$
    
    \State $planes[\text{``Top''}] \gets ((s/2, s/2, s),\ R_y(180^\circ))$
    \State $planes[\text{``Bottom''}] \gets ((s/2, s/2, 0),\ Rx(0)$
    \State $planes[\text{``Right''}] \gets ((s, s/2, s/2),\ R_y(-90^\circ))$
    \State $planes[\text{``Left''}] \gets ((0, s/2, s/2),\ R_y(90^\circ) \circ R_z(90^\circ))$
    \State $planes[\text{``Back''}] \gets ((s/2, s, s/2),\ R_x(90^\circ))$
    
    \If{$plane\_name$ is \text{``2-2-2''}}
        \State $planes[\text{``Top''}] \gets (s/2, s/2, s),\ R_x(180^\circ) \circ R_z(-90^\circ)$
    \ElsIf{$plane\_name$ is $\text{``1-4-1''}$}
        \State $planes[\text{``Left''}] \gets (0, s/2, s/2),\ R_y(90^\circ) \circ$
    \EndIf
    
    \If{$plane\_name\in [\text{``1-4-1-0'', ``2-3-1-0''}]$}
        \State $planes[\text{``Front''}] \gets ((s/2, 0, s/2),\ R_x(-90^\circ))$
    \ElsIf{$plane\_name\in [\text{``1-4-1-1'', ``1-4-1-4'', ``2-3-1-1'', ``2-2-2''}]$}
        \State $planes[\text{``Front''}] \gets ((s/2, 0, s/2),\ R_x(-90^\circ)\circ R_z(-90^\circ))$
    \ElsIf{$plane\_name\in [\text{``1-4-1-2'', ``1-4-1-5'', ``2-3-1-2'', ``3-3''}]$}
        \State $planes[\text{``Front''}] \gets ((s/2, 0, s/2),\ R_x(-90^\circ)\circ R_z(180^\circ))$
    \ElsIf{$plane\_name \text{ is ``1-4-1-3''}$}
        \State $planes[\text{``Front''}] \gets ((s/2, 0, s/2),\ R_x(-90^\circ)\circ R_z(90^\circ))$
    \EndIf

    \If {$plane\_name\in[\text{``1-4-1-4'', ``1-4-1-5''}]$}
        \State $planes[\text{``Back''}] \gets ((s/2, s, s/2),\ R_x(90^\circ)\circ R_z(90^\circ))$
    \EndIf
    \vspace{0.5em}
    \State \textbf{Form a cube by:}
    \ForAll{$face\_name\in planes$}
        \State $placement\gets planes[face\_name]$
        \State $square\gets$\texttt{FreeCAD.makePlane}$(s,\ s,\ placement)$
        \State $c\gets map[face\_name]$
        \If{$rot$ is true}
            \State Assign \texttt{rotate}$(c,\ 90^\circ)$ to $square$ at $placement$
        \Else
            \State Assign $c$ to $square$ at $placement$
        \EndIf
    \EndFor
    \State $img \gets$ \texttt{FreeCAD.saveView}$(view)$
    \State \Return $img$
\EndFunction
\vspace{0.5em}
\Function{DrawNet}{$net,\ map,\ s,\ rot$}
    \For{$face\_name \in net$}
        \State $i,\ j\gets net[face\_name]$
        \State $pos \gets (j \cdot s,\ (H-1-i) \cdot s,\ 0)$
        \State $square\gets$\texttt{FreeCAD.makePlane}$(s,\ s,\ (pos,0^\circ))$
        \State $c\gets map[face\_name]$
        \If{$rot$ is true}
            \State Assign \texttt{rotate}$(c,\ 90^\circ)$ to $square$ at $pos$
        \Else
            \State Assign $c$ to $square$ at $pos$
        \EndIf
    \EndFor
    \State $img \gets \texttt{FreeCAD.saveImage}()$
    \State \Return $img$
\EndFunction
\end{algorithmic}
\end{algorithm}

\clearpage
\begin{algorithm}[ht]
\caption{Functions for Unfolding Cube to 11 kinds of Cube Nets}
\label{alg_netu}
\begin{algorithmic}[1]
\State Using the same parameter definitions as those in \autoref{alg_netr}

\Function{DrawNetWiPivot}{$plane\_name, net, map, s, rot$}
    \State $pivot\_plane\_name\gets \text{``1-4-1-0''}$
    \State Initialize rotation dictionary $planes$
    \If{$plane\_name\in[\text{``1-4-1-1'', ``1-4-1-4'', ``2-3-1-1'', ``2-2-2''}]$}
        \State $planes[\text{``Front''}] \gets R_z(90^\circ))$
    \ElsIf{$plane\_name\in[\text{``1-4-1-2'', ``1-4-1-5'', ``2-3-1-2'', ``3-3''}]$}
        \State $planes[\text{``Front''}] \gets  R_z(-180^\circ))$
    \ElsIf{$plane\_name$ \text{ is ``1-4-1-3''}}
        \State $planes[\text{``Front''}] \gets R_z(-90^\circ))$
    \EndIf
    \If{$plane\_name\in [\text{``1-4-1-4'', ``1-4-1-5''}]$}
        \State $planes[\text{``Back''}] \gets R_z(-90^\circ))$
    \EndIf
    \If{$plane\_name\in[\text{``2-3-1-0'', ``2-3-1-1'', ``2-3-1-2'', ``3-3'', ``2-2-2''}]$}
        \State $planes[\text{``Left''}] \gets R_z(-90^\circ))$
    \EndIf
    \If{$plane\_name$ \text{ is ``2-2-2''}}
        \State $planes[\text{``Top''}] \gets R_z(-90^\circ))$
    \EndIf
    \vspace{0.2em}
    \State \textbf{Create a net which can form the same cube with pivot plane:}
    \For{$face\_name \in net$}
        \State $i,\ j\gets net[face\_name]$
        \State $pos \gets (j \cdot s,\ (H-1-i) \cdot s,\ 0)$
        \State $square\gets$\texttt{FreeCAD.makePlane}$(s,\ s,\ (pos,0^\circ))$
         \If{$rot$ is true}
            \State Assign \texttt{rotate}$(c,\ 90^\circ)$ to square at $pos$
        \Else
            \State Assign $c$ to $square$ at $pos$
        \EndIf
        \If{$plane\_name \ne \text{``1-4-1-0''}$}
            \If{$face\_name\in planes$}
                \State $rotation \gets planes[face\_name]$
                \State $square.Placement.Rotation\gets rotation$
            \EndIf
        \EndIf
    \EndFor
    \State $img \gets \texttt{FreeCAD.saveImage}()$
\EndFunction
\end{algorithmic}
\end{algorithm}

\clearpage
\begin{algorithm}[t]
\caption{Cube Unfolding Task}
\label{alg_CU}
\begin{algorithmic}[1]
\State \textbf{Input:} Color(Pattern) set $C$, unit length $s$, task mode $m$
\State Initialize 11 cube nets \newline $nets:\{face\_name: (i,j)|face\_name \in \{\text{``Top'',\ ``Bottom'',\ ``Right'',\ ``Left'',\ ``Back'',\ ``Front''}\}\}$
\State Initialize empty lists $positive\_samples$, $negative\_samples$
\State $map:\{face\_name:c |c\in C\}\gets$Randomly shuffle set $C$ and assign it to six faces
\State Randomly select a $view\in$8 corner views of a cube
\State $pivot\_net\_name\gets \text{``1-4-1-0''}$

\State $ref\_img\gets$\texttt{Net2Cube}$(pivot\_net\_name, map, view, rot=\text{false})$

\For{$i\gets1$ to 2}
    \State $plane\_name, net\gets$ Randomly select net from $nets$ 
    \State $img\gets$\texttt{DrawNetWiPivot}$(plane\_name, net, map, s, rot=\text{false})$
    \State Append $img$ to $positive\_samples$
    \If{$m=\text{``pattern''}$}
        \State $img'\gets$\texttt{DrawNetWiPivot}$(plane\_name, net, map, s, rot=\text{true})$
        \State Append $img'$ to $negative\_samples$
    \EndIf
\EndFor
\vspace{0.3em}
\State $map'\gets$Fix the mapping of $face\_name\in view$, and random shuffle the others
\For{$i\gets1$ to 2}
    \State $plane\_name,net\gets$ Randomly select net from $nets$ 
    \State $img\gets$\texttt{DrawNetWiPivot}$(plane\_name,net, map, s, rot=\text{false})$
    \State Append $img$ to $positive\_samples$
\EndFor
\vspace{0.3em}
\State $map'\gets$Swap the colors(patterns) of a randomly selected $face\in view$ with its opposite face
\State $plane\_name, net\gets$ Randomly select net from $nets$ 
\State $img\gets$\texttt{DrawNetWiPivot}$(plane\_name, net, map', s, rot=\text{false})$ 
\State Append $img$ to $negative\_samples$
\vspace{0.5em}
\State $samples\gets(positive\_samples, negative\_samples)$
\State Shuffle $samples$ to assign $[A,B,C,D]$ and record $answer\_id$
\State $data \gets$ \texttt{create\_data}$(ref\_img, samples, question, answer\_id)$
\end{algorithmic}
\end{algorithm}

\begin{algorithm}[ht]
\caption{Cube Reconstruction Task}
\label{alg_CR}
\begin{algorithmic}[1]
\State \textbf{Input:} Color(Pattern) set $C$, unit length $s$, task mode $m$
\State Initialize 11 cube nets \newline $nets:\{face\_name: (i,j)|face\_name \in \{\text{``Top'',\ ``Bottom'',\ ``Right'',\ ``Left'',\ ``Back'',\ ``Front''}\}\}$
\State Initialize empty lists $positive\_samples$, $negative\_samples$
\State $map:\{face\_name:c |c\in C\}\gets$Randomly shuffle set $C$ and assign it to six faces
\State $net\in\{0,1\}^{3\times 5}\gets$ Randomly select net from $nets$ 
\State $ref\_img\gets$\texttt{DrawNet}$(net, map, s, rot=\text{false})$ and append $img$ to $positive\_samples$
\For{$i\gets1$ to 3}
    \State $view\gets$ Randomly select a view from 8 corner views of a cube
    \State $img\gets$\texttt{Net2Cube}$(net, map, view, rot=\text{false})$ 
    \State Append $img$ to $positive\_samples$
\EndFor
\For{$flip\_dir \in \{$``horizontal'', ``vertical''\}}
    \State Randomly choose $sample$ from $positive\_samples$ 
    \State $img \gets$ \texttt{flip}$(sample, flip\_dir)$
    \State Append $img$ to $negative\_samples$
\EndFor
\vspace{0.5em}
\State $samples\gets(positive\_samples, negative\_samples)$
\State Shuffle $samples$ to assign $[A,B,C,D]$ and record $answer\_id$
\State $data \gets$ \texttt{create\_data}$(ref\_img, samples, question, answer\_id)$
\end{algorithmic}
\end{algorithm}

\clearpage
\begin{algorithm}[ht]
\caption{Cross-Section Task}
\label{alg_CS}
\begin{algorithmic}[1]
\State \textbf{Input:} Number of objects $num$, number of sections per mode $k$, whether rotate the slicing plane $rot$
\State Initialize candidate objects list $objects$, empty list $selected\_objects$
\State Initialize empty lists $positive\_samples$, $negative\_samples$
\Function{GetSections}{$compound,\ k,\ plane$}
    \State Initialize empty list $imgs$
    \State Determine $coord_{min}$ and $coord_{max}$ from bounding box
    \State $step \gets (coord_{max} - coord_{min})/(k+1)$
    
    \For{$i \gets 1$ to $k$}
        \State $offset \gets coord_{min} + i \times step$
        \State $normal\_vector \gets$ unit vector normal to $plane$
        \State $section \gets$ \texttt{FreeCAD.slice}($compound,\ normal\_vector,\ offset$)
        \State Rotate $section$ for better visualization
        \State $img\gets\texttt{FreeCAD.savaImage}(section)$ and append $img$ to $imgs$
    \EndFor
    \State\Return $imgs$
\EndFunction
\vspace{0.5em}

\Function{GetRotatedSections}{$compound,\ axis,\ center$}
    \State $axis\_vector \gets$ Corresponding unit vector of $axis$
    \State $plane\gets$ Parallel to $axis$
    
    \For{$angle \in \{45^\circ, 135^\circ\}$}
        \State $axix\_vector' \gets$ \texttt{rotate}$(axis\_vector,angle,plane)$
        \State $offset \gets axix\_vector \cdot center$
        \State $section \gets$ \texttt{FreeCAD.slice}($compound,\ axis\_vector,\ offset$)
        \State Rotate $section$ for better visualization
        \State $img\gets\texttt{FreeCAD.savaImage}(section)$ and append $img$ to $imgs$
    \EndFor
    \State\Return $imgs$
\EndFunction
\vspace{0.5em}

\State $selected\_objects\gets $Randomly select $num$ objects from $objects$
\State Randomly assign sizes to objects in $selected\_objects$
\State $compound\gets$ Create objects in FreeCAD and compound objects
\State $center\gets$Obtain the center of compound object
\For{$plane\in\{\text{``XY'',``XZ'', ``YZ''}\}$}
    \State $imgs\gets$ \texttt{GetSections}$(compound,\ k,\ plane)$
    \State Append $imgs$ to $positive\_samples$
\EndFor
\If{$rot$ is true}
    \For{$axis\in \{\text{``x'',\ ``y'',\ ``z''}\}$}
        \For{$angle\in\{45^\circ,\ 135^\circ\}$}
            \State $imgs\gets$ \texttt{GetRotatedSections}$(compound,\ axis,\ center)$
            \State Append $imgs$ to $positive\_samples$
        \EndFor
    \EndFor
\EndIf
\State $compound'\gets$Randomly alter the relative ratios of objects in $compound$
\State $imgs\gets$ Use any of the above approaches to obtain cross-sections of $compound'$
\State \State Append $imgs$ to $negative\_samples$

\vspace{0.5em}
\State $samples\gets(positive\_samples, negative\_samples)$
\State Shuffle $samples$ to assign $[A,B,C,D]$ and record $answer\_id$
\State $data \gets$ \texttt{create\_data}$(ref\_img, samples, question, answer\_id)$    
\end{algorithmic}
\end{algorithm}

\clearpage
\begin{algorithm}[ht]
\caption{Cube Counting Task}
\label{alg_CC}
\begin{algorithmic}[1]
\State \textbf{Input:} Spatial size $(X, Y, Z)$, cube size $s$, number of constraint views $num$
\State Initialize zero value 3D tensors $placement \in \{0\}^{Z \times Y \times X}$, empty list $cubes$
\State Initialize empty list $samples$

\Function{DetectGrid}{$view,\ row\_num\, col\_num$}
    \State $contours\gets$Find contours in $view$
    \State Initialize $grid$ matrix of size $row\_num \times col\_num$
    \For{$contour\in contours$}
        \State $(x,y,w,h) \gets$ Bounding rectangle of $contour$
        \State $row \gets y / h$, $col \gets x / w$
        \If{$row$ and $col$ within bounds}
            \State $grid[row][col] \gets 1$
        \EndIf
    \EndFor
    \State \Return $grid$
\EndFunction
\vspace{0.5em}

\Function{GetCubeAnswer}{$front,\ top,\ left,\ num$}
    \State $sum\_front\_col \gets$ Column sums of $front$
    \State $sum\_top\_col \gets$ Column sums of $top$
    \State $max\_2view \gets sum\_front\_col \cdot sum\_top\_col$
    \State $min\_2view \gets \texttt{sum}(sum\_top\_col - 1 + sum\_front\_col)$
    
    \If{$num = 2$}
        \State \Return $(max\_2view, min\_2view)$
    \EndIf
    
    \vspace{0.5em}
    \State $sum\_left\_col \gets$ Column sums of $left$
    \State Initialize answer matrix with the same dimension as $top\in\{0\}^{H\times W}$
    
    \For{$row \gets 0$ to $H-1$}
        \For{$col \gets 0$ to $W-1$}
            \If{$top[row][col] = 1$}
                \State $ans[row][col] \gets \texttt{min}(sum\_front\_col[col], sum\_left\_col[row])$
            \EndIf
        \EndFor
    \EndFor
    
    \State $max\_3view \gets \texttt{sum}(ans)$
    \State $sum\_top\_row \gets$ Row sums of $top$
    \State $min\_3view \gets \texttt{max}(\texttt{sum}(sum\_top\_row - 1 + sum\_left\_col) , min\_2view)$
    \State \Return $(max\_3view, min\_3view)$
\EndFunction
\vspace{0.5em}

\State Update $placement, cubes$ with \texttt{CreateCubes}$(X,\ Y,\ Z)$
\State Update $placement, cubes$ with \texttt{ConnectIsolatedCubes}$(X,\ Y)$
\State $(front\_view,\ top\_view,\ left\_view)\gets$ \texttt{SaveViews($cubes$)}
\State $front\_mat,\ top\_mat,\ left\_mat\gets \newline \texttt{DetectGrid}(front\_view),\ \texttt{DetectGrid}(top\_view),\ \texttt{DetectGrid}(left\_view)$
\If{$num=2$}
    \State $ref\_img\gets(top\_view,\ front\_view)$
    \State $(max\_view, min\_view)\gets\texttt{GetCubeAnswer}(front\_mat,\ top\_mat,\ left\_mat,\ 2)$
\ElsIf{$num=3$}
    \State $ref\_img\gets(top\_view,\ front\_view,\ left_view)$
    \State $(max\_view, min\_view)\gets\texttt{GetCubeAnswer}(front\_mat,\ top\_mat,\ left\_mat,\ 3)$
\EndIf
\vspace{0.5em}
\State $samples\gets$Generate correct and incorrect nums based on the $min\_view$ to $max\_view$ range
\State Shuffle $samples$ to assign $[A,B,C,D]$ and record $answer\_id$
\State $data \gets$ \texttt{create\_data}$(ref\_img, samples, question, answer\_id)$ 
\end{algorithmic}
\end{algorithm}

\clearpage
\begin{algorithm}[ht]
\caption{Functions for Splitting Cube Stack into Several Connected Parts}
\label{alg_split}
\begin{algorithmic}[1]
\Function{GetNeighbors}{$cube\_pos,\ cubes$}
    \State $(x,y,z) \gets cube\_pos$
    \State Initialize empty list $neighbours$
    \For{$dx \in \{-1,0,1\}$}
        \For{$dy \in \{-1,0,1\}$}
            \For{$dz \in \{-1,0,1\}$}
                \If{$|dx| + |dy| + |dz| = 1$} \Comment{6-connected neighborhood}
                    \State $neighbor\_pos \gets (x+dx, y+dy, z+dz)$
                    \If{$neighbor\_pos \in cubes$}
                        \State Append $neighbor\_pos$ to $neighbours$
                    \EndIf
                \EndIf
            \EndFor
        \EndFor
    \EndFor
    \State \Return $neighbors$
\EndFunction
\vspace{0.5em}

\Function{RegionGrowing}{$cubes,\ max\_cubes$}
    \State Initialize empty set $part$, empty list $queue$
    \State $start\_pos \gets$ Randomly select a position from $cubes$ and append $start\_pos$ to $queue$
    
    \While{$queue$ not empty and $|part| < max\_cubes$}
        \State $current\_pos \gets \texttt{pop}(queue, 0)$
        \If{$current\_pos \notin part$}
            \State Add $current\_pos$ to $part$
            \State $neighbors \gets$ \texttt{GetNeighbors}$(current\_pos,\ cubes)$
            \State Extend $[n \in neighbors \mid n \notin part]$ to $queue$
        \EndIf
    \EndWhile
    \State \Return $part$
\EndFunction
\vspace{0.5em}

\Function{IsContinuous}{$part$}
    \State Initialize empty set $part$, empty list $queue$ 
    \State $start\_pos \gets part[0]$ and append $start\_pos$ to $queue$
    
    \While{$queue$ not empty}
        \State $current\_pos \gets \texttt{pop}(queue, 0)$
        \If{$current\_pos \notin visited$}
            \State Add $current\_pos$ to $visited$
            \State $neighbors \gets$ \texttt{GetNeighbors}$(current\_pos,\ part)$
            \State Extend $[n \in neighbors \mid n \in part\,\, and \,\, n \notin visited]$ to $queue$
        \EndIf
    \EndWhile
    \State \Return Whether $|visited| = |part|$
\EndFunction
\vspace{0.5em}

\Function{SplitCubes}{$cubes,\ max\_cubes,\ num\_parts$}
    \State $part1 \gets$ \texttt{RegionGrowing}$(cubes, max\_cubes)$
    \If{\texttt{IsContinuous}$(part1)$}
        \State $remaining \gets$ Remove $part1$ from $cubes$
    \EndIf
    
    \If{\texttt{IsContinuous}$(remaining)$}
        \If{$num\_parts=2$}
            \State\Return \texttt{sort}$([part1, remaining])$ by size
        \ElsIf{$num\_parts=3$}
            \State Similarly find $part2$ from remaining cubes as above
            \State $part3 \gets$ Remove $part2$ from $remaining$
            \State \Return \texttt{sort}$([part1, part2, part3])$ by size
        \EndIf
    \EndIf
\EndFunction
\end{algorithmic}
\end{algorithm}

\clearpage
\begin{algorithm}[ht]
\caption{Cube Assembly Task}
\label{alg_CA}
\begin{algorithmic}[1]
\State \textbf{Input:} Spatial size $(X, Y, Z)$, cube size $s$, number of splitting parts $k$
\State Initialize zero value 3D tensors $placement \in \{0\}^{Z \times Y \times X}$, empty list $cubes$
\State Initialize empty lists $ref\_imgs$, $positive\_samples$, $negative\_samples$
\Function{CreateCubesPyramid}{$X,\ Y,\ Z$}
    \State Initialize $num=1$
    \For{$y\gets 0$ to $Y-1$}
        \State $num=\texttt{randomInt}(num,\ \texttt{min}(y+2,\ X))$
        \For{$x \gets 0$ to $num-1$}
            \State \texttt{CreateCube}$(x,\ y,\ 0)$
        \EndFor
    \EndFor
    \For{$z\gets 1$ to $Z-2$}
        \State Initialize $num=0$
        \For{$y\gets 0$ to $Y-1$}
            \State $num=\texttt{randomInt}(num,\ \texttt{max}(num, \texttt{sum}(placement[z-1][y])))$
            \For{$x \gets 0$ to $num-1$}
                \State \texttt{CreateCube}$(x,\ y,\ z)$
            \EndFor
        \EndFor
    \EndFor
    \For{$y\gets 0$ to $Y-1$}
        \For{$x \gets 0$ to $X-1$}
            \State With $50\%$ probability \texttt{CreateCube}$(x,\ y,\ Z-1)$
        \EndFor
    \EndFor
\EndFunction
\vspace{0.5em}

\State Update $placement, cubes$ with \texttt{CreateCubesPyramid}$(X,\ Y,\ Z)$
\State $cubes\_img\gets\texttt{FreeCAD.saveImage}(cubes)$ and append $cubes\_img$ to $ref\_imgs$
\State $parts\gets \texttt{SplitCubes}(cubes, max\_cubes,\ num\_parts)$
\For{$part\in parts[:-1]$}
    \State $part\_img\gets\texttt{FreeCAD.saveImage}(part)$ and append $part\_img$ to $ref\_imgs$
\EndFor
\State $part\_img\gets\texttt{FreeCAD.saveImage}(parts[-1])$ and append $part\_img$ to $positive\_samples$
\For{$i\gets$ 1 to 2}
    \State $part'\gets$Randomly remove 1 cube from part[-1]
    \State $part'\_img\gets\texttt{FreeCAD.saveImage}(part')$ and append $part\_img$ to $negative\_samples$
\EndFor
\vspace{0.5em}
\State $samples\gets(positive\_samples, negative\_samples)$
\State Shuffle $samples$ to assign $[A,B,C,D]$ and record $answer\_id$
\State $data \gets$ \texttt{create\_data}$(ref\_img, samples, question, answer\_id)$    
\end{algorithmic}
\end{algorithm}

\clearpage
\begin{algorithm}[ht]
\caption{Simulation for Arrow Moving}
\label{alg_arrowpath}
\begin{algorithmic}[1]
\State \textbf{Class} ArrowPath
\State \textbf{Attributes:}
\State \quad $W, H, k$: Map width, height, and step count
\State \quad $max\_step \gets \min(x, y)$
\State \quad$directions\gets \text{\{(0,1),(1,0),(0,-1),(-1,0)\}}$ \Comment{up, right, down, left}
\State \quad$path$: Initialize with empty list to record relative moving direction and steps
\State \quad$states$: Initialize with empty list to record pos and orientation during transformation
\Function{InitializeState}{}
    \State Reset $path,\ states$
    \State $orient\_id \gets\texttt{randomInt}(0,3)$ 
    \State $pos\in\{(x,y)\} \gets$ Randomly select a position in the map
    \State Append $(orient\_id,\ pos)$ to $states$
\EndFunction
\vspace{0.5em}

\Function{GetRelativeDirection}{$orient\_id$}
    \State $forward \gets directions[orient\_id]$
    \State $backward \gets (-forward[0], -forward[1])$
    \State $left \gets directions[(orient\_id - 1) \mod 4]$
    \State $right \gets directions[(orient\_id + 1) \mod 4]$
    \State \Return $\{\text{``forward'':forward, ``backward'':backward, ``left'':left, ``right'':right}\}$
\EndFunction
\vspace{0.5em}

\Function{UpdateOrientId}{$rel\_dir,\ orient\_id$}
    \If{$rel\_dir$ is \text{``backward''}} 
        \State $orient\_id \gets (orient\_id + 2) \mod 4$
    \ElsIf{$rel\_dir$ is \text{``left''}} 
        \State $orient\_id \gets (orient\_id - 1) \mod 4$
    \ElsIf{$rel\_dir$ is \text{``right''}} 
        \State $orient\_id \gets (orient\_id + 1) \mod 4$
    \EndIf
    \State\Return $orient\_id$
\EndFunction
\vspace{0.5em}

\Function{Move}{$state,\ rel\_dir,\ steps$}
    \State $pos,\ orient\_id\gets state$
    \State $move\_dir \gets \texttt{GetRelativeDirection}(orient\_id)[rel\_dir]$
    \State $new\_pos \gets [pos[0]+move\_dir[0]\times steps, pos[1]+move\_dir[1]\times steps]$
    \If{$new\_pos$ is invalid}
        \State\Return false
    \EndIf
    \State Append $(rel\_dir, steps)$ to $path$
    \State Append $(\texttt{UpdateOrientId}(rel\_dir,\ orient\_id),\ new\_pos)$ to $states$
    \State\Return true
\EndFunction
\vspace{0.5em}

\Function{GeneratePath}{$k,\ end\_state$=None}
    \For{$i\gets 1$ to $k$}
        \Repeat
            \State Randomly select $rel\_dir\in\{\text{``forward'',\ ``backward'',\ ``left'',\ ``right''}\}$
            \State $steps \gets \texttt{randomInt}(1, max\_step)$
            \State $valid\_flag\gets\texttt{Move}(states[-1],\ rel\_dir,\ steps)$
            \If{$end\_state$ is not None and $i=k$}
                \State $valid\_flag\gets valid\_flag \And state[-1]\ne end\_state$
            \EndIf
        \Until{$valid\_flag$ is true}
    \EndFor
\EndFunction
\end{algorithmic}
\end{algorithm}

\clearpage
\begin{algorithm}[ht]
\caption{Simulation for Arrows Moving}
\label{alg_arrowmap}
\begin{algorithmic}[1]
\State \textbf{Class} ArrowMap(Inherit from \textbf{Class} ArrowPath)
\State \textbf{Attributes:} 
\State \quad$colors$: Color set 
\State \quad$path$: Initialize with empty list to record arrow position, relative moving direction and steps
\State \quad $states$: Initialize with empty list to record map during transformation
\Function{InitializeState}{}
    \State Initialize empty matrix $state$
    \For{$y\gets$1 to $H$}
        \For{$x\gets$ 1 to $W$}
            \State With $50\%$ probability: 
            \State Randomly select $color\in colors$ 
            \State Randomly get $orient\_id\gets\texttt{randomInt}(0,3)$
            \State $state[pos]\gets$ Record $color$ and $orient\_id$ at $pos(x,y)$
        \EndFor
    \EndFor
    \State Append $state$ to $states$
\EndFunction
\vspace{0.5em}

\Function{Move}{$state, arrow\_pos,\ rel\_dir,\ steps$}
    \State $curr\_pos\gets arrow\_pos$
    \State $curr\_orient\_id,\ curr\_color\gets state[x][y]$
    \State $move\_dir \gets \texttt{GetRelativeDirection}(curr\_orient\_id)[rel\_dir]$
    \State $new\_pos \gets [pos[0]+move\_dir[0]\times steps, pos[1]+move\_dir[1]\times steps]$
    \If{$new\_pos$ is invalid}
        \State\Return false
    \EndIf
    \State $new\_orient\_id\gets \texttt{UpdateOrientId}(rel\_dir,\ orient\_id)$
    \If{$new\_pos=curr\_pos$ and $new\_orient\_id=curr\_orient\_id$}
        \State\Return false 
    \EndIf
    \State Append $arrow\_pos,\ rel\_dir,\ steps$ to $path$

    \If{$state[new\_pos]$ is None}
        \State $state[curr\_pos]\gets$ None
    \Else
        \State $target\_color,\ target\_orient\_id\gets state[new\_pos]$
        \State $target\_move\_dir\gets -move\_dir$
        \State $target\_rel\_directions\gets\texttt{GetRelativeDirection}(target\_orient\_id)$
        \State $taget\_rel\_dir\gets$ Find $\{key\in target\_rel\_directions \mid value=target\_move\_dir\}$
        \State $new\_target\_orient\_id\gets \texttt{UpdateOrientId}(taget\_rel\_dir,\ target\_orient\_id)$
        \State $state[curr\_pos]\gets$ $target\_color$ and $new\_target\_orient\_id$
    \EndIf
    \State $state[new\_pos]\gets$ $curr\_color$ and $curr\_orient\_id$
    \State\Return true
\EndFunction
\vspace{0.5em}

\Function{GeneratePath}{$k,\ end\_state$=None}
    \For{$i\gets 1$ to $k$}
        \Repeat
            \State Randomly select $arrow\_pos\in \{pos\mid state[pos]\text{ is not None}\}$
            \State Randomly select $rel\_dir\in\{\text{``forward'',\ ``backward'',\ ``left'',\ ``right''}\}$
            \State $steps \gets \texttt{randomInt}(1, max\_step)$
            \State $valid\_flag\gets\texttt{Move}(state, arrow\_pos, rel\_dir,\ steps)$
            \If{$end\_state$ is not None and $i=k$}
                \State $valid\_flag\gets valid\_flag \And state[-1]\ne end\_state$
            \EndIf
        \Until{$valid\_flag$ is true}
    \EndFor
\EndFunction
\end{algorithmic}
\end{algorithm}

\clearpage
\begin{algorithm}[ht]
\caption{Arrow Moving Task in Easy Version}
\label{alg_AM0}
\begin{algorithmic}[1]
\State \textbf{Input:} Dimension of map $(W, H)$, step count $k$
\State Initialize empty lists $positive\_samples$, $negative\_samples$
\State Initialize $arrow\_path$ with dimension ${W\times H}$
\State Initialize state with $arrow\_path.\texttt{InitializeState}()$ and record as $initial\_state$ 
\State Update $path,\ states$ with $arrow\_path.\texttt{GeneratePath}(k)$ 
\State Append $path$ to $positive\_samples$
\State $ref\_img\gets\texttt{draw\_map}(states[0], states[-1])$
\State Record $end\_state\gets states[-1]$ 
\vspace{0.5em}
\State From the same $initial\_state$ 
\For{$i\gets 1$ to 3}
    \State Update $path'$ with $arrow\_path.\texttt{GeneratePath}(k,\ end\_state)$
    \State Append $path'$ to $negative\_samples$
\EndFor
\vspace{0.5em}
\State $samples\gets(positive\_samples, negative\_samples)$
\State Shuffle $samples$ to assign $[A,B,C,D]$ and record $answer\_id$
\State $data \gets$ \texttt{create\_data}$(ref\_img, samples, question, answer\_id)$    
\end{algorithmic}
\end{algorithm}

\begin{algorithm}[ht]
\caption{Arrow Moving Task in Hard Version}
\label{alg_AM1}
\begin{algorithmic}[1]
\State \textbf{Input:} Dimension of map $(W, H)$, step count $k$, task mode $m$
\State Initialize empty lists $positive\_samples$, $negative\_samples$
\State Initialize $arrow\_map$ with dimension ${W\times H}$
\State Initialize state with $arrow\_map.\texttt{InitializeState}()$ and record as $initial\_state$ 
\State Update $path,\ states$ with $arrow\_map.\texttt{GeneratePath}(k)$ 
\State Append $path$ to $positive\_samples$
\If{$m=\text{``state''}$}
    \State $ref\_img\gets\texttt{draw\_map}(states[0])$
    \State Append $states[-1]$ to $positive\_samples$
\ElsIf{$m=\text{``path''}$}
    \State $ref\_img\gets\texttt{draw\_map}(states[0], state[-1])$
    \State Append $path$ to $positive\_samples$
\EndIf
\State Record $end\_state\gets states[-1]$ 
\vspace{0.5em}
\State From the same $initial\_state$ 
\For{$i\gets 1$ to 3}
    \State Update $path',\ states'$ with $arrow\_map.\texttt{GeneratePath}(k,\ end\_state)$
    \If{$m=\text{``state''}$}
        \State Append $states'[-1]$ to $negative\_samples$
    \ElsIf{$m=\text{``path''}$}
        \State Append $path'$ to $negative\_samples$ 
    \EndIf
\EndFor
\vspace{0.5em}
\State $samples\gets(positive\_samples, negative\_samples)$
\State Shuffle $samples$ to assign $[A,B,C,D]$ and record $answer\_id$
\State $data \gets$ \texttt{create\_data}$(ref\_img, samples, question, answer\_id)$    
\end{algorithmic}
\end{algorithm}

\clearpage
\begin{algorithm}[ht]
\caption{Simulation for Block Moving }
\label{alg_block}
\begin{algorithmic}[1]
\State \textbf{Class} Block
\State \textbf{Attributes:} 
\State \quad$X, Y, Z, k$: Spatial size and step count
\State \quad$directions$: 6 directions 
\State \quad$colors$: Color set
\State \quad$cubes\_info$: Initialize with empty list to record positions and colors of cube objects
\State \quad$transformation$: Initialize with empty list to record transformations
\Function{InitializeState}{}
    \State Update $cubes$ with $\texttt{CreateCubes}(X, Y, Z)$
    \State Assign randomly selected colors to $cubes$ and record their colors and positions in $cubes\_info$
\EndFunction
\vspace{0.5em}

\Function{HasSupport}{$x,\ y,\ z$}
    \If{$z=0$ or there is cube at $(x,y,z-1)$}
        \State \Return true
    \EndIf
    \State \Return flase
\EndFunction
\vspace{0.5em}

\Function{DropCubes}{}
    \State Sort $cubes\_info$ by $z$ of $pos$ in ascending order
    \For{$cube\in cubes\_info$}
        \State $(x,y,z)\gets$ Acquire position of $cube$ from $cubes\_info$
        \While{\texttt{HasSupport}$(x,y,z)$ is flase}
            \State Change the position of $cube$ to $(x,y,z-1)$ and update $z\gets z-1$
        \EndWhile
    \EndFor
\EndFunction
\vspace{0.5em}

\Function{CheckMove}{$from\_pos,\ to\_pos$}
    \If{($to\_pos$ is invalid) or ($\texttt{HasSupport}(to\_pos)$ is false) or (there is no cube at $from\_pos$)\newline\hspace*{2em}or (there is no cube at $to\_pos$ and $to\_pos$ is on top of $from\_pos$)}
        \State\Return false
    \EndIf
    \State\Return true
\EndFunction
\vspace{0.5em}

\Function{MoveCube}{$from\_pos,\ to\_pos$}
    \If{there is no cube at $to\_pos$}
        \State Update $cubes\_info$ with changing the position of $cube$ at $from\_pos$ to $to\_pos$
    \Else
        \State Update $cubes\_info$ with swapping the cube at $from\_pos$ and $to\_pos$
    \EndIf 
    \State $\texttt{DropCubes}()$
    \State Append $(from\_pos,\ to\_pos-from\_pos)$ to $transformation$
\EndFunction
\vspace{0.5em}

\Function{GenerateTransformation}{$k$}
    \For{$i\gets 1$ to $k$}
        \State Initialize empty list $possible\_moves$
        \ForAll{$cube\in cubes\_info$}
            \ForAll{$direction\in directions$}
                \State $to\_pos\gets $ The position of cube $from\_pos+direction$
                \If{\texttt{CheckMove}($from\_pos,\ to\_pos$) is true}
                    \State Append $(from\_pos,\ direction,\ to\_pos)$ to $possible\_moves$
                \EndIf
            \EndFor
        \EndFor
        \State Randomly select $(from\_pos,\ direction,\ to\_pos)\in possible\_moves$
        \State \texttt{MoveCube}($from\_pos,\ to\_pos$) 
    \EndFor
\EndFunction
\end{algorithmic}
\end{algorithm}

\clearpage
\begin{algorithm}[ht]
\caption{Block Moving Task}
\label{alg_BM}
\begin{algorithmic}[1]
\State \textbf{Input:} Spatial size $(X,\ Y,\ Z)$, step count $k$
\State Initialize empty lists $ref\_imgs$, $positive\_samples$, $negative\_samples$
\State Initialize $block$ with size $(X, Y, Z)$
\State Initialize with $block.\texttt{InitializeState}()$ and record as $initial\_cubes\_info$
\State $img\gets\texttt{FreeCAD.saveImage}(initial\_cubes)$ and append $img$ to $ref\_imgs$
\vspace{0.5em}
\State Update $transformation,\ cubes\_info$ with $block.\texttt{GenerateTransformation}(k)$ 
\State Append $transformation$ to $positive\_samples$
\State Record $final\_cubes\_info$ after transformation
\State $img\gets \texttt{FreeCAD.saveImage}(final\_cubes)$ and append $img$ to $ref\_imgs$
\vspace{0.5em}
\State From the same $initial\_cubes\_info$
\For{$i\gets$ 1 to 3}
    \Repeat
        \State Update $transformation',\ cubes\_info'$ with $block.\texttt{GenerateTransformation}(k)$
    \Until{$cubes\_info\ne final\_cubes\_info$}
    \State Append $transformation$ to $negative\_samples$
\EndFor
\vspace{0.5em}
\State $samples\gets(positive\_samples, negative\_samples)$
\State Shuffle $samples$ to assign $[A,B,C,D]$ and record $answer\_id$
\State $data \gets$ \texttt{create\_data}$(ref\_imgs, samples, question, answer\_id)$    
\end{algorithmic}
\end{algorithm}

\section{Dataset Characteristic}
\label{characteristic}
\textbf{Option Modality \& Format}
A significant majority of questions (818) feature image-based options to emphasize visual reasoning. The choice formats are intentionally varied, including standard A/B/C/D choices (508 questions), options with A/B/C/‘All three other options are incorrect’ (310 questions), and unique text (242 questions) or numeric (120 questions) answers to prevent models from overfitting to a single question style. For the numeric answers, we additionally provide direct numerical responses, and in \ref{ComparisonFormat} we present a comparative analysis of model performance across different question format.

\textbf{Answer Distribution}
The answer distribution is well-balanced across options A (26.5\%), B (27.5\%), and C (28.5\%). The lower frequency of option D (17.5\%) is a deliberate design choice to enhance the rigor of the evaluation. For many complex tasks, option D serves the distinct role of "All three other options are incorrect". This asymmetrical design is critical for two reasons. First, it acknowledges the difficulty of generating multiple high-quality distractors for complex 3D tasks, ensuring all visual options remain challenging. Second, it compels models to move beyond simple heuristics like "pick the most similar". Instead, this approach demands eliminative reasoning, requiring the model to rule out every other option to prove a genuine understanding of the spatial rules being tested.
\clearpage
\section{Data Examples}
\label{examples}
We present exemplars of varying difficulty levels for all tasks, with each sample containing an image, question, options, answer, and explanation. 

\begin{figure}[htbp]
    \centering
    \includegraphics[width=0.8\linewidth]{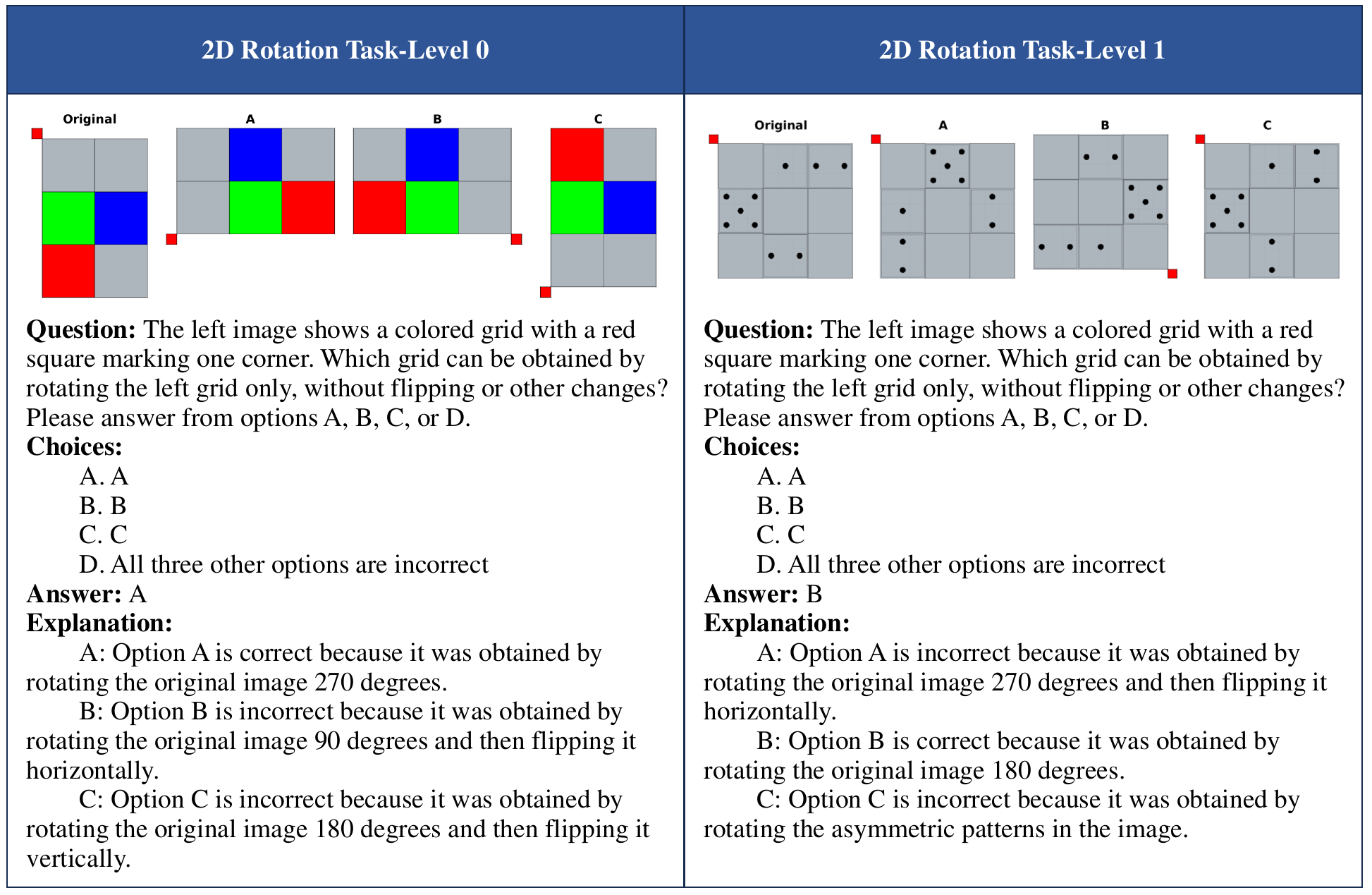}
    \caption{2D Rotation Task.}
    \label{fig:2DRotation}
\end{figure}

\begin{figure}[htbp]
    \centering
    \includegraphics[width=0.8\linewidth]{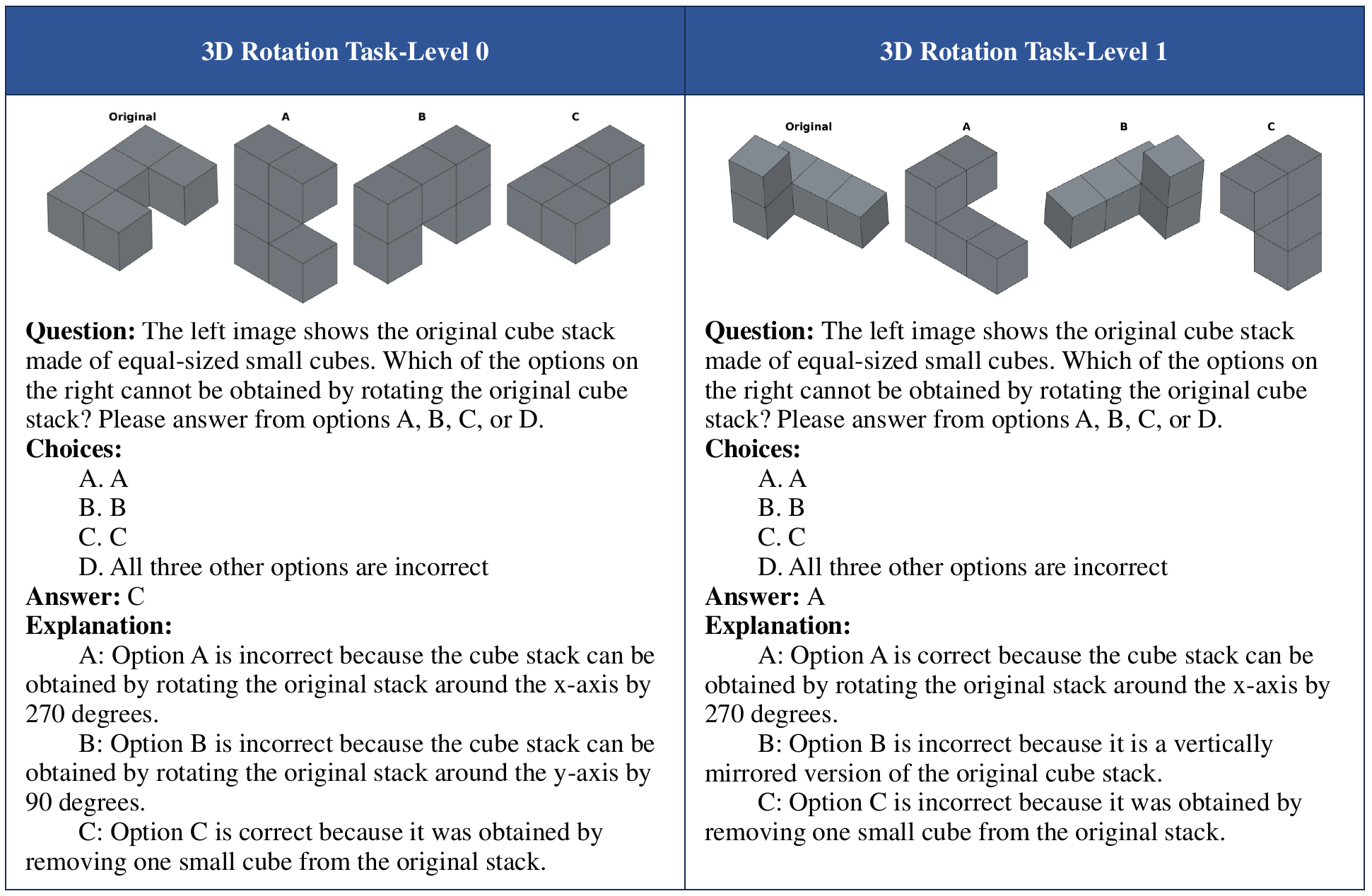}
    \caption{3D Rotation Task.}
    \label{fig:3DRotation}
\end{figure}

\begin{figure}[htbp]
    \centering
    \includegraphics[width=0.8\linewidth]{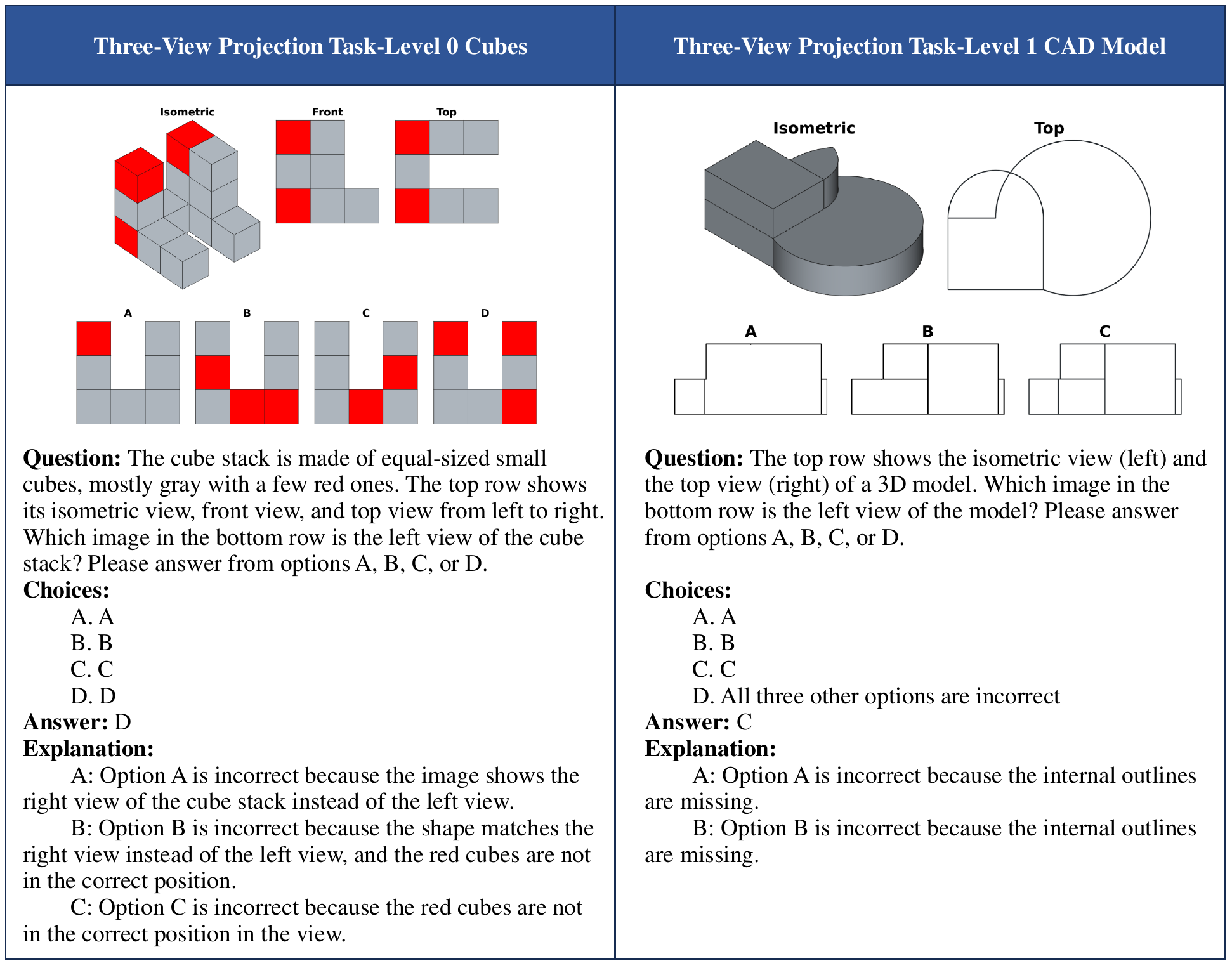}
    \caption{Three-view Projection Task.}
    \label{fig:3Views}
\end{figure}

\textbf{Mental Rotation} 2DRotation: \autoref{fig:2DRotation}, 3DRotation: \autoref{fig:3DRotation}, 3ViewProjection: \autoref{fig:3Views};

\begin{figure}[htbp]
    \centering
    \includegraphics[width=1\linewidth]{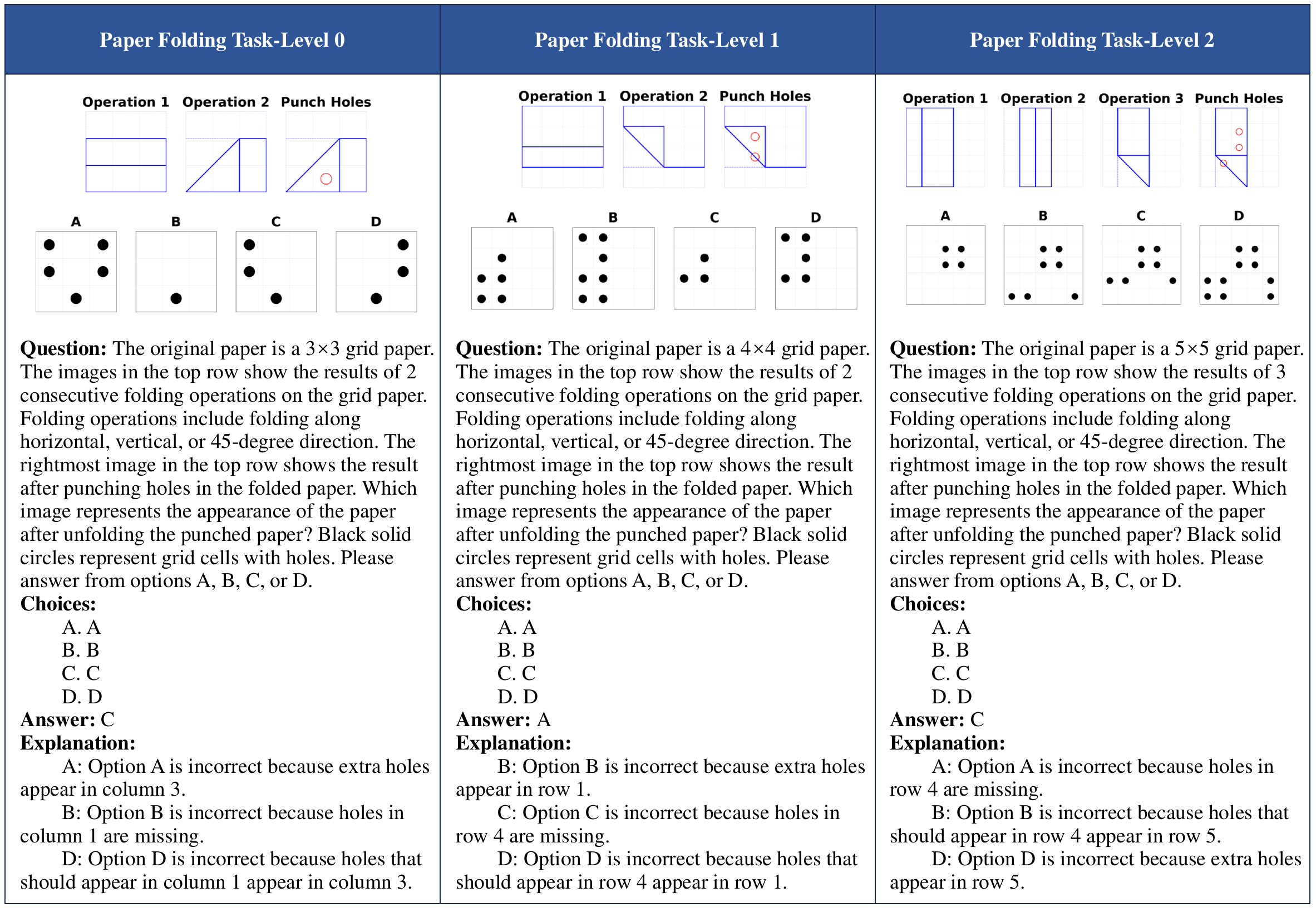}
    \caption{Paper Folding Task.}
    \label{fig:PaperFolding}
\end{figure}

\begin{figure}[htbp]
    \centering
    \includegraphics[width=1\linewidth]{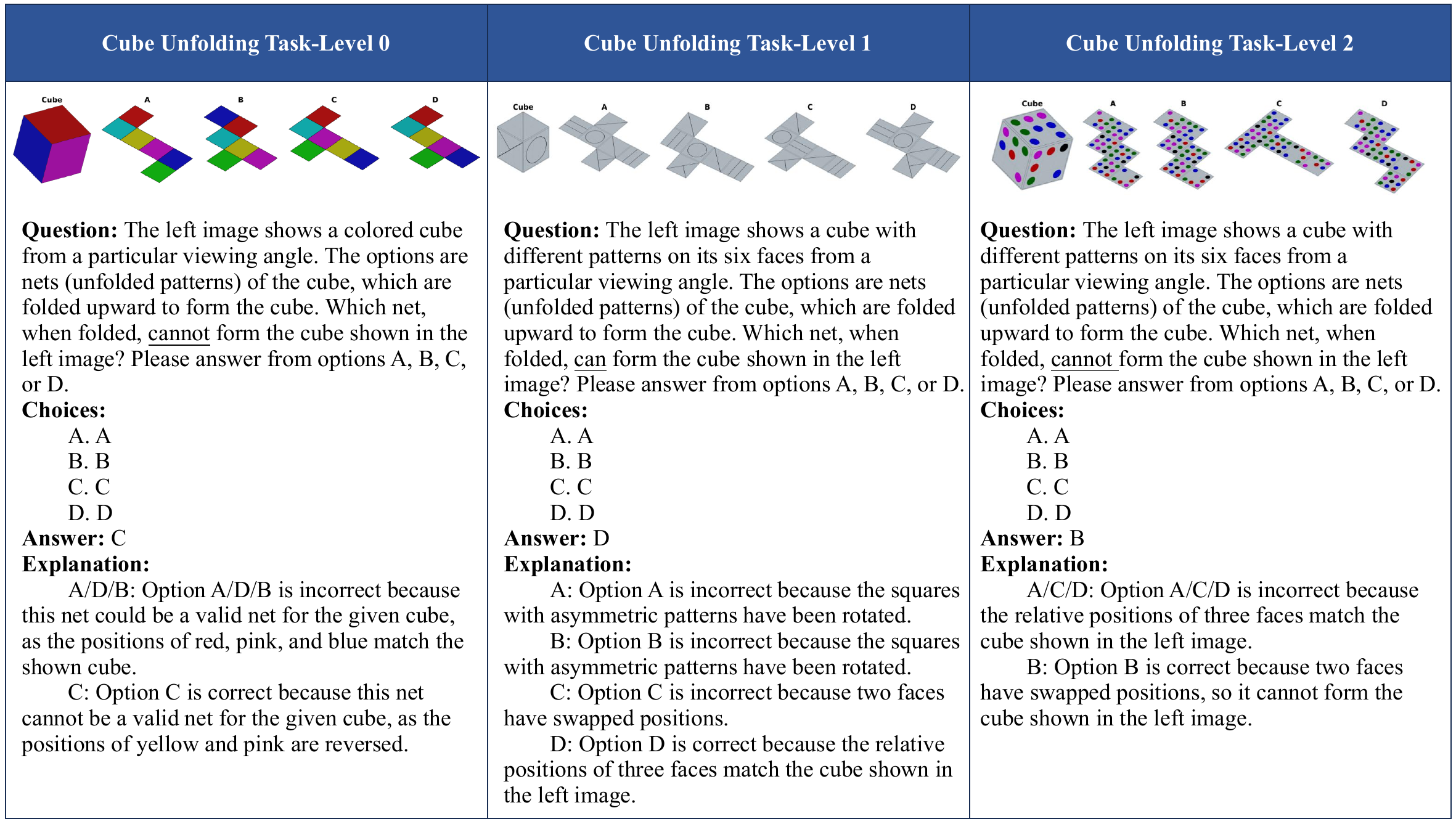}
    \caption{Cube Unfolding Task.}
    \label{fig:CubeUnfolding}
\end{figure}

\begin{figure}[htbp]
    \centering
    \includegraphics[width=1\linewidth]{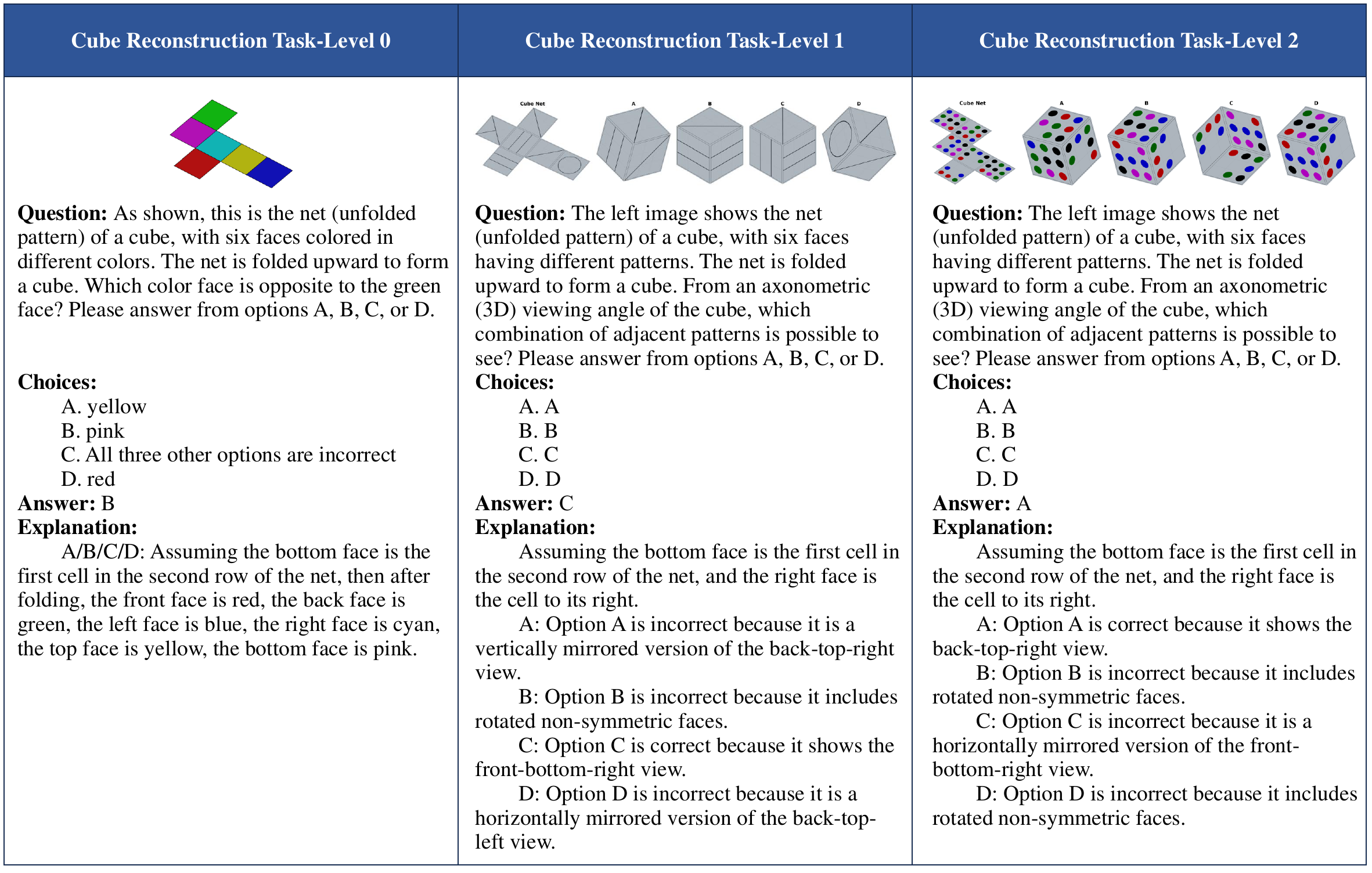}
    \caption{Cube Reconstruction Task.}
    \label{fig:CubeReconstruction}
\end{figure}

\textbf{Mental Folding} PaperFolding: \autoref{fig:PaperFolding}, CubeUnfolding: \autoref{fig:CubeUnfolding}, CubeReconstruction: \autoref{fig:CubeReconstruction};

\begin{figure}[htbp]
    \centering
    \includegraphics[width=1\linewidth]{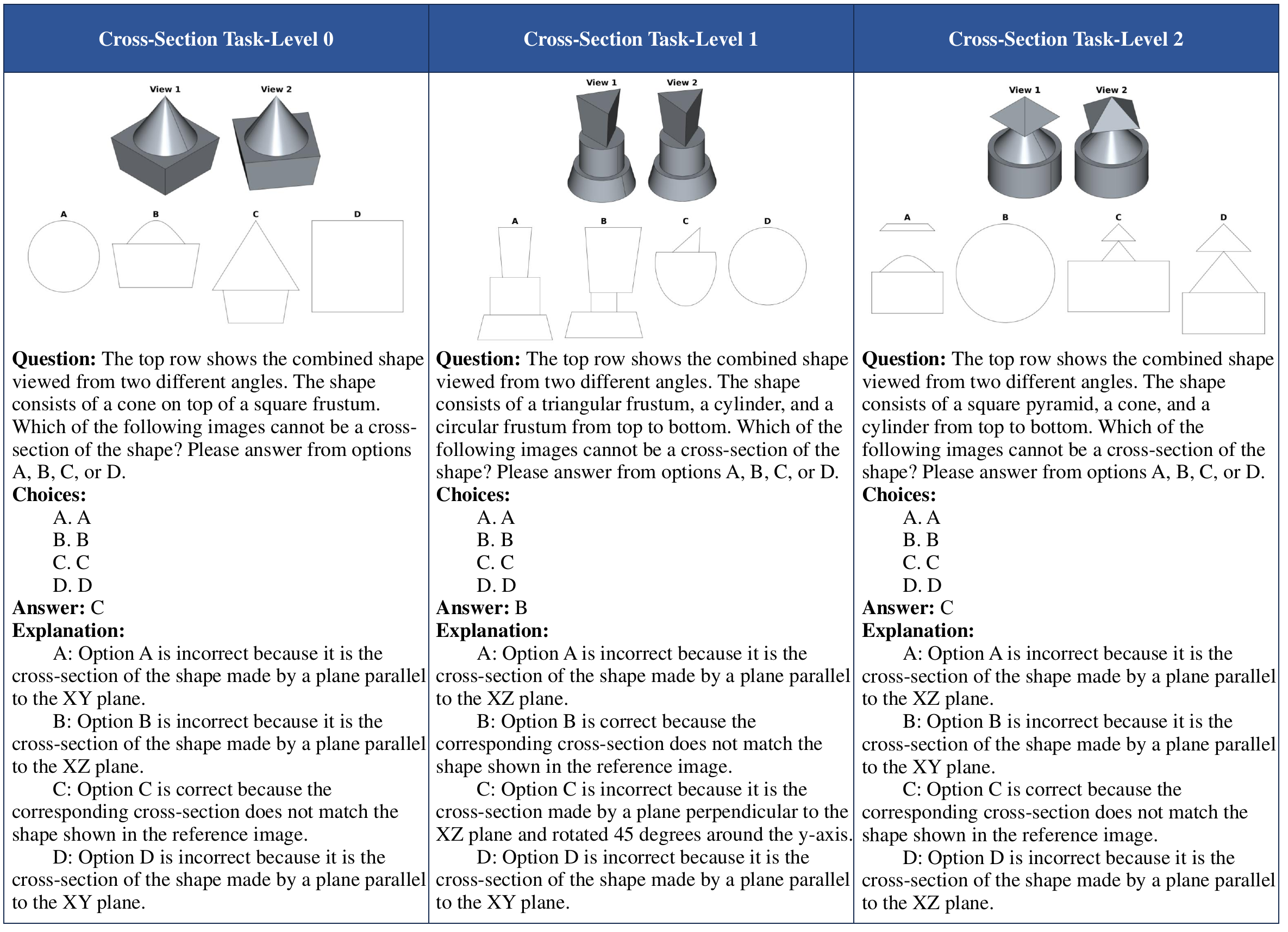}
    \caption{Cross-sectionn Task.}
    \label{fig:CrossSection}
\end{figure}

\begin{figure}[htbp]
    \centering
    \includegraphics[width=1\linewidth]{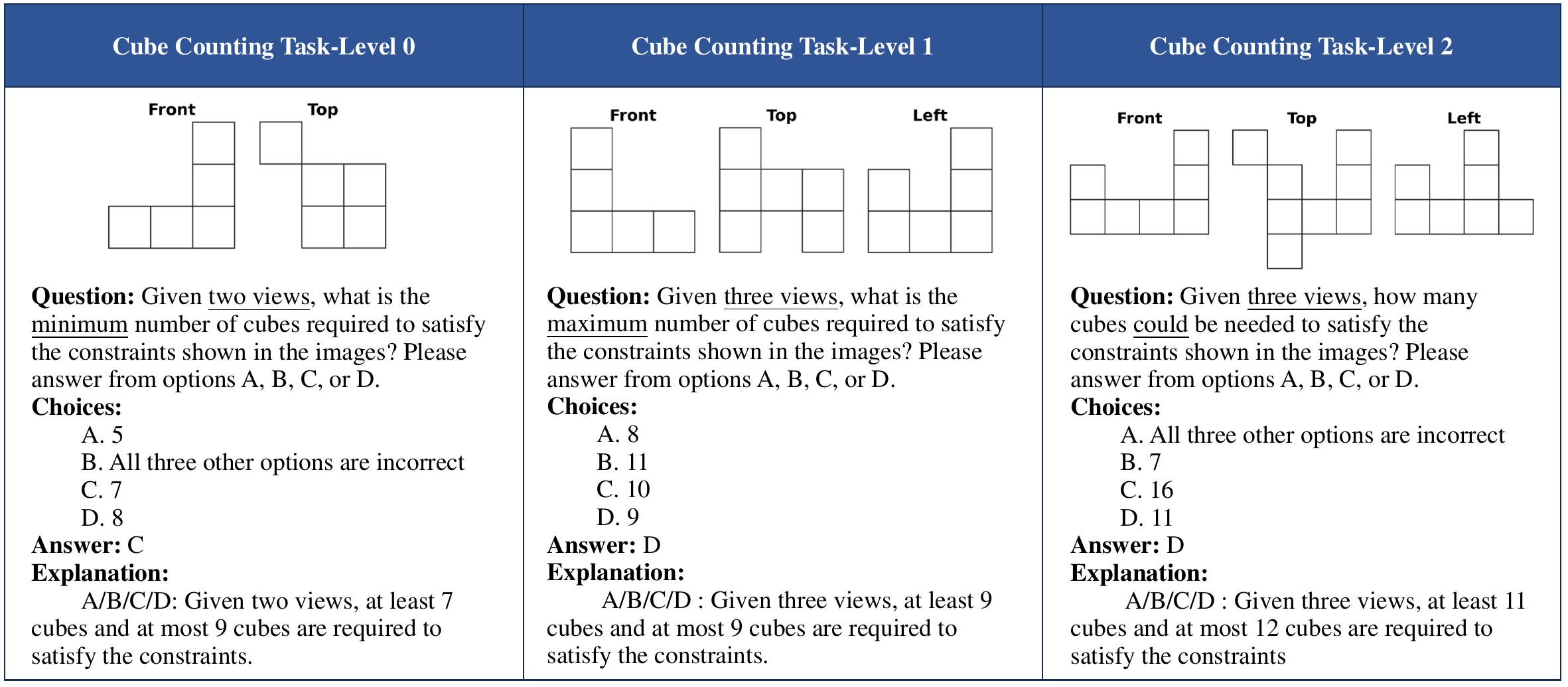}
    \caption{Cube Counting Task.}
    \label{fig:CubeCounting}
\end{figure}

\begin{figure}[htbp]
    \centering
    \includegraphics[width=0.8\linewidth]{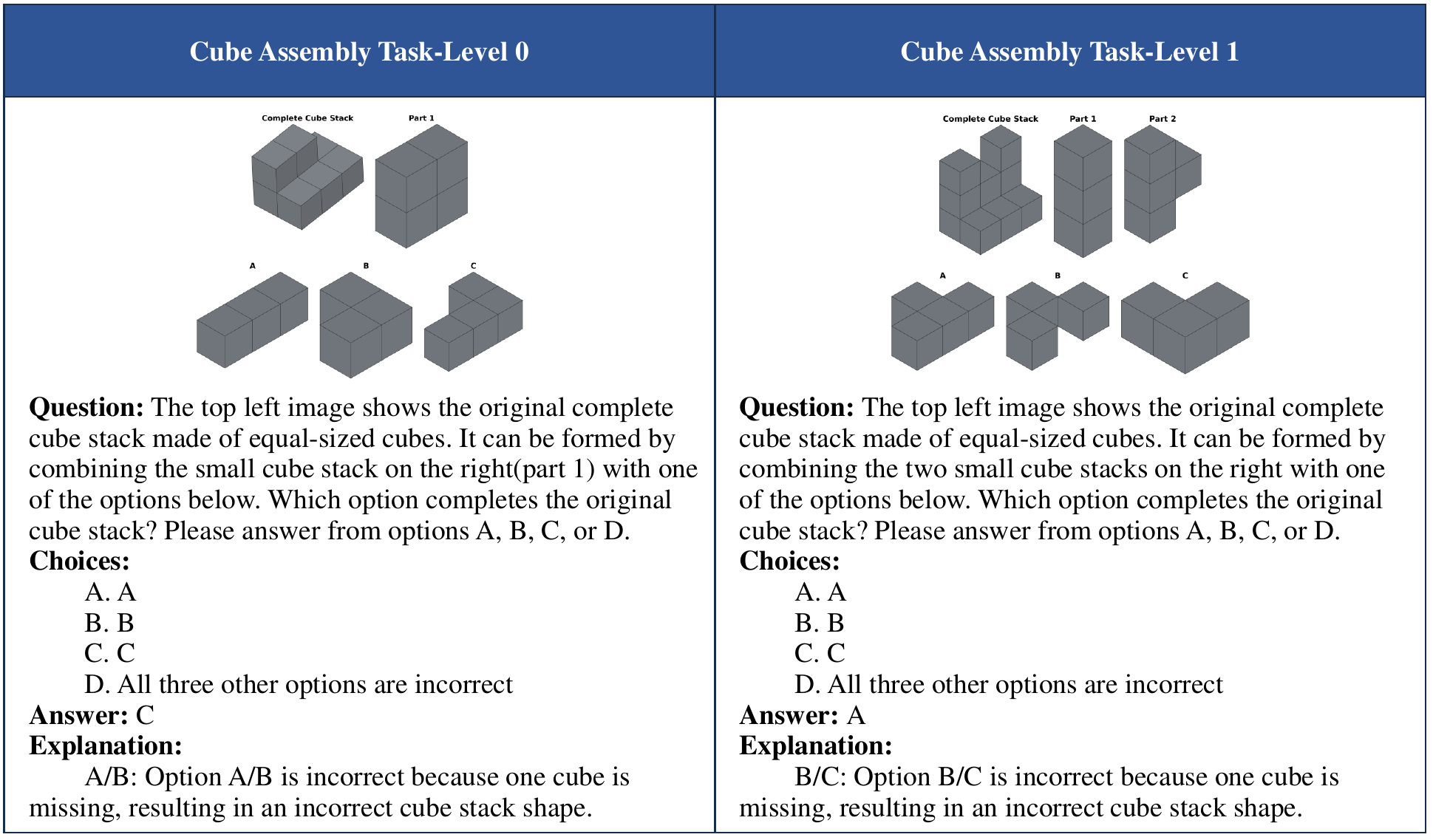}
    \caption{Cube Assembly Task.}
    \label{fig:CubeAssembly}
\end{figure}

\textbf{Visual Penetration} CrossSection: \autoref{fig:CrossSection}, CubeCounting: \autoref{fig:CubeCounting}, CubeAssembly: \autoref{fig:CubeAssembly};

\begin{figure}[htbp]
    \centering
    \includegraphics[width=1\linewidth]{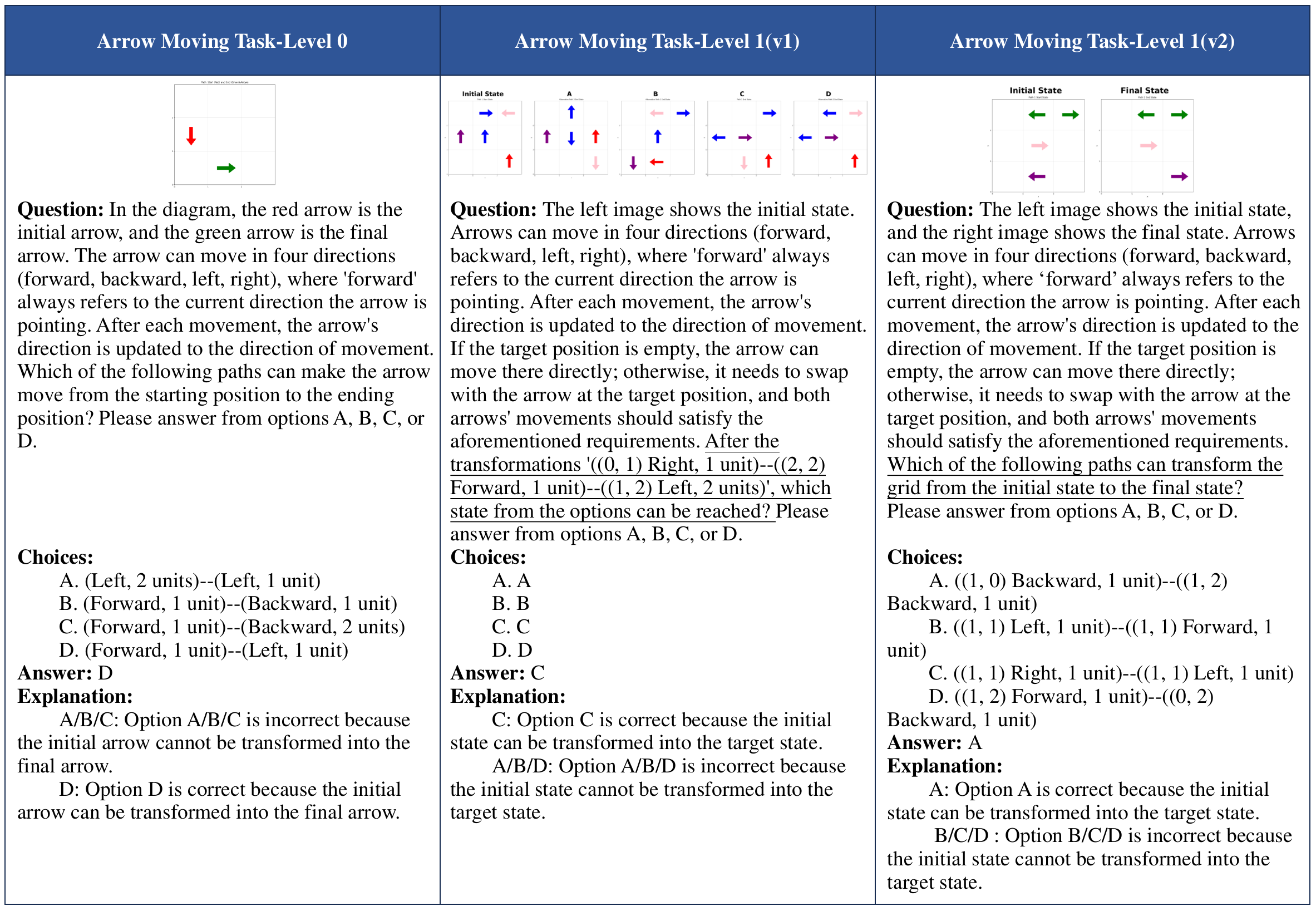}
    \caption{Arrow Moving Task.}
    \label{fig:ArrowMoving}
\end{figure}

\begin{figure}[htbp]
    \centering
    \includegraphics[width=0.8\linewidth]{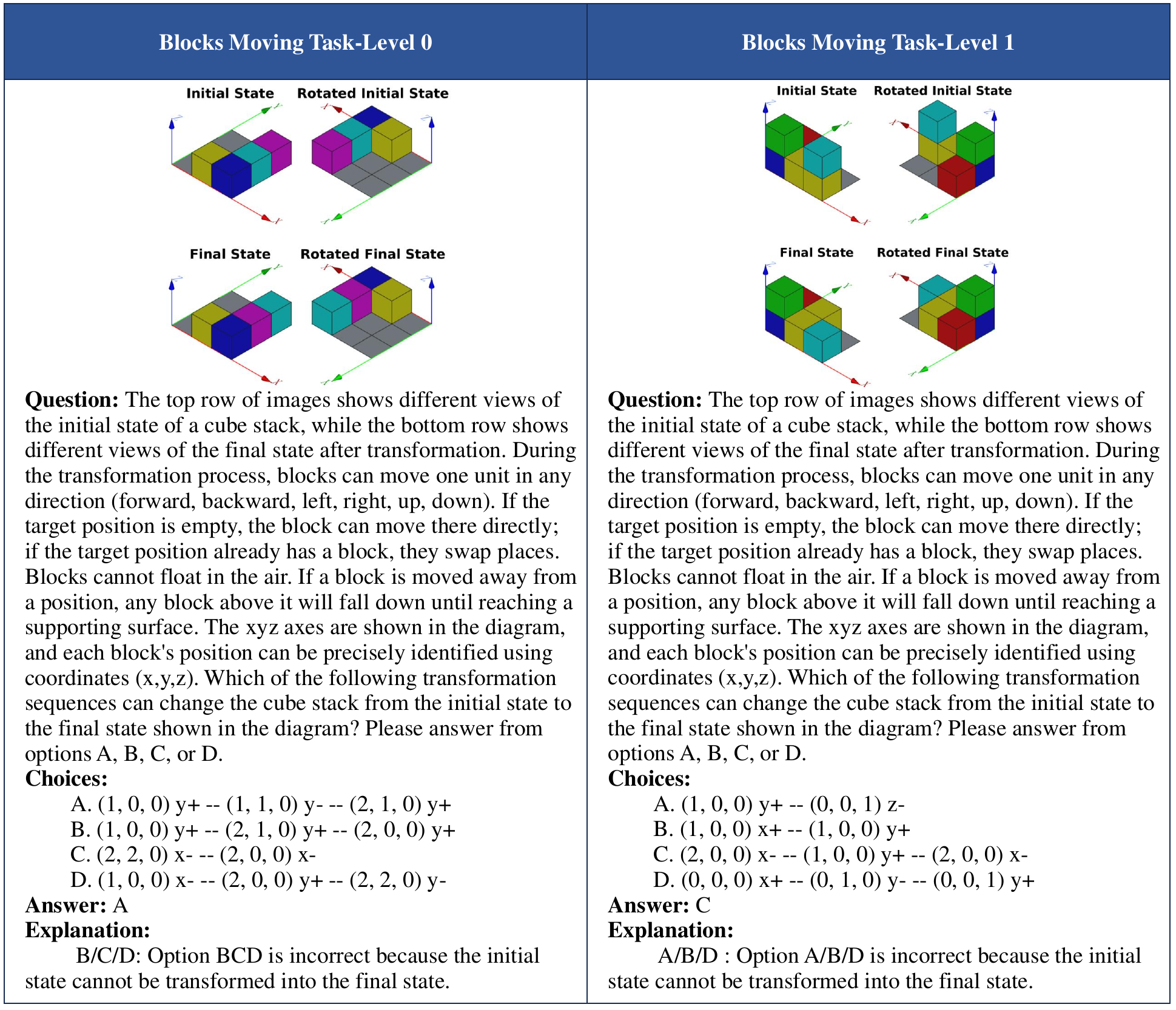}
    \caption{Block Moving Task.}
    \label{fig:BlockMoving}
\end{figure}

\begin{figure}[htbp]
    \centering
    \includegraphics[width=0.8\linewidth]{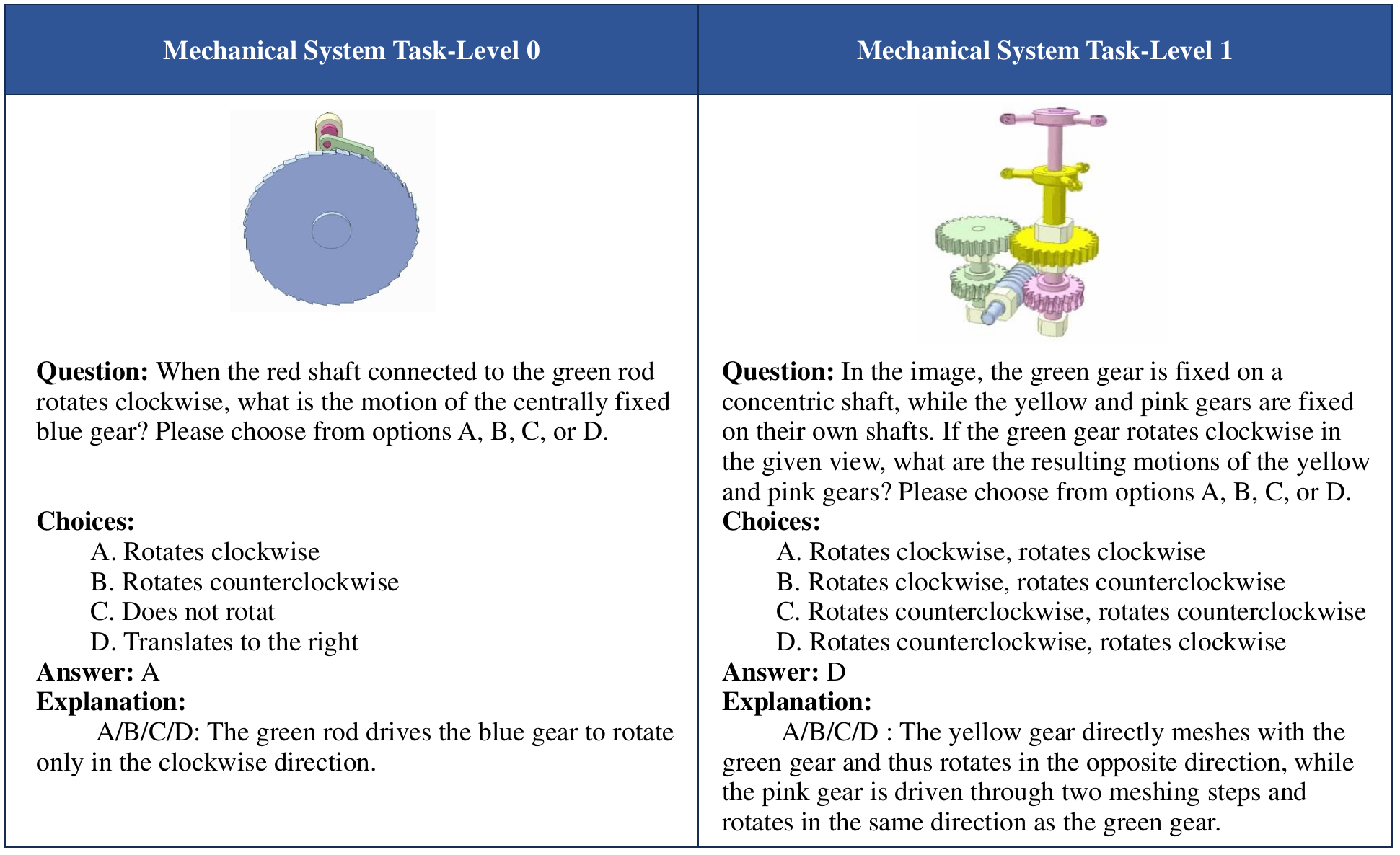}
    \caption{Mechanical System Task.}
    \label{fig:MechanicalSystem}
\end{figure}

\textbf{Mental Animation} ArrowMoving: \autoref{fig:ArrowMoving}, BlockMoving: \autoref{fig:BlockMoving}, MechanicalSystem: \autoref{fig:MechanicalSystem}.

\clearpage
\section{Evaluation Details}
\label{D}

\subsection{Models}
For the DeepseekVL2 series, InternVL2.5 series, InternVL3 series and SAIL-VL series, we deployed these models on H100 servers and used the officially provided code to load the pre-trained models for inference. For all other models, we employed API calls through OpenAI's client service for inference. All closed-source models accessed via API in this study were used with specific, identifiable versions to ensure consistency and reproducibility. Specifically, we used the following model versions:

\begin{itemize}[leftmargin=1.2em]
\vspace{-0.25em}
    \item \texttt{gpt-4o-2024-08-06} for GPT-4o
    \item \texttt{o1-2024-12-17} for o1
    \item \texttt{claude-3-5-sonnet-20240620} for Claude-3.5-Sonnet
    \item \texttt{claude-3-7-sonnet-20250219} for Claude-3.7-Sonnet
    \item \texttt{Gemini-2.5-flash-preview-04-17} for Gemini-2.5-flash
    \item \texttt{Gemini-2.5-pro-preview-03-25} for Gemini-2.5-pro
    \item \texttt{Doubao-1-5-vision-pro-32k-250115} for Doubao-1-5-vision-pro
    \item \texttt{qwen-vl-max-0408} for Qwen-VL-max
\end{itemize}

\subsection{Prompts for Response Generation}
\label{D.1}
We use the prompt template as follows: 

1) \textbf{Original CoT Prompt A} from DeepSeek-R1\citep{deepseekr1}:
"You should first provide a reasoning process, then provide a single option (A, B, C or D) as the final answer. The reasoning process and the answer are enclosed within <think></think> and <answer></answer> tags, respectively, i.e., <think>reasoning process</think>, <answer>answer</answer>.\verb|\n|Question: \textbf{<question here>}\verb|\n|A.\textbf{<option A here>}\verb|\n|B.\textbf{<option B here>}\verb|\n|C.\textbf{<option C here>}\verb|\n|D.\textbf{<option D here>}\verb|\n|"

2) \textbf{Variant CoT Prompt B} from EMMA\citep{emma}:
"Answer with the option’s letter from the given choices and put the letter in one '\verb|\|boxed{}'. Please solve the problem step by step.\verb|\n|Question: <question here>\verb|\n|A.<option A here>\verb|\n|B.<option B here>\verb|\n|C.<option C here>\verb|\n|D.<option D here>\verb|\n|"

3) \textbf{Non-CoT Prompt: }
"Answer with a single option letter (A, B, C, or D), enclosed within the <answer></answer> tag. For example: <answer>A</answer>. Ensure that your output contains only the final answer, without any intermediate reasoning or additional content.\verb|\n|Question: \textbf{<question here>}\verb|\n|A.\textbf{<option A here>}\verb|\n|B.\textbf{<option B here>}\verb|\n|C.\textbf{<option C here>}\verb|\n|D.\textbf{<option D here>}\verb|\n|"

\subsection{Zero-shot Setting}
Our decision to focus exclusively on the zero-shot evaluation setting is grounded in both methodological precedent and practical considerations. This approach aligns with the standards set by many recent, high-impact benchmark papers, such as Math-Vision~\citep{mathvision}, MM-IQ~\citep{mmiq}, and EMMA~\citep{emma}, all of which centered their evaluations on the zero-shot setting to assess novel reasoning capabilities.
While we considered few-shot prompting, we concluded its utility is limited in our context of complex spatial reasoning. For these intricate visualization tasks, providing examples with only the final answer offers minimal effective guidance. On the other hand, creating effective chain-of-thought examples that include complete, multi-step reasoning would be prohibitively expensive for comprehensive benchmarking.

\subsection{Methods for Answer Extraction}
\label{extraction rule}
To ensure robust evaluation and minimize parsing errors, we employ a hierarchical, two-stage rule-based approach for answer extraction.

\noindent\textbf{Stage 1: Coarse Extraction with Boundary Enforcement.} \\
Adopting the strategy from MME-CoT~\citep{mmecot}, we first attempt to locate the answer segment by scanning for a comprehensive set of standard identifiers, including XML-style tags (e.g., \texttt{<answer></answer>}) and natural language markers (e.g., "<answer>", "Answer:", "Final answer", "final answer", "Final Answer", "the answer is", "The answer is", "correct answer", "Correct answer", "Correct Answer", and "correct path"). The text following these markers is isolated and truncated at the first subsequent period delimiter. Critically, to prevent false positives where common words starting with option letters (e.g., ``\underline{A}ll'', ``\underline{B}ackward'') are mistakenly identified as answers, we enforce strict word boundary constraints. We utilize the regular expression \texttt{\textbackslash b([A-D])\textbackslash b} to accept only standalone option letters.

\noindent\textbf{Stage 2: Prioritized Fine-Grained Matching.} \\
In instances where the coarse extraction fails to yield a valid option, we trigger a secondary, high-precision extraction routine. This process iterates through a prioritized list of compiled regular expression patterns designed to handle specific formatting variations (e.g., tagged encapsulated outputs, boxed answers) and semantic fallback structures. The patterns are applied in the following order:

\begin{itemize}[leftmargin=1.1em]
    \small
    \vspace{-0.25em}
    \item \textbf{CoT Prompt A with tags:} \\
    \texttt{r"<answer>\textbackslash s*(?P<value>.*?)\textbackslash s*</answer>"}
    
    \item \textbf{CoT Prompt B with boxes:}\\
    \texttt{r"\textbackslash\textbackslash \{1,2\}boxed\{(?:(?:\textbackslash\textbackslash text|rm)\{)?(?P<value>[A-D])"}
    
    \item \textbf{Other common answer formats:} \\
    \texttt{r"<answer>\textbackslash s*option\textbackslash s+(?P<value>[A-D])(?=</answer>")}\\
    \texttt{r"(?:final|correct)\textbackslash s+answer\textbackslash s*(?:is:)\textbackslash s*(?:option\textbackslash s*)?(?P<value>[A-D])\textbackslash b"}\\
    \texttt{r"option\textbackslash s+(?P<value>[A-D])\textbackslash b"}\\
    \texttt{r"choose\textbackslash s+(?P<value>[A-D])\textbackslash b"}
    \vspace{-0.25em}
\end{itemize}

This dual-layer approach ensures high recall for compliant responses while maintaining precision against hallucinated or verbose outputs. Even with these rules, 100\% parsing success isn't guaranteed, as models can still flexibly produce outputs in non-standard formats. For the purpose of our comparative analysis, we designate the baseline coarse extraction method (excluding strict boundary enforcement) as \textbf{Extract Rule A}, and the comprehensive dual-stage strategy described herein as \textbf{Extract Rule B}.

For multiple-choice questions, a response is considered correct if and only if the extracted result contains exactly one uppercase option letter (A, B, C, or D) matching the standard answer. For non-choice questions, we perform direct string matching between the extracted result and the reference answer. This hybrid rule-based evaluation ensures consistent and fair judgment across both option-based and open-form tasks.

\subsection{Human Performance}
\label{human}
To establish a robust human baseline analogous to the tested MLLMs, we recruited 8 graduate students (4 Ph.D., 4 M.S.; aged 22-27) from mechanical engineering and computer science. All participants possessed strong backgrounds in geometry and physics, confirmed through their academic curriculum, and reported familiarity with spatial reasoning tasks. This selection criterion was chosen because it mirrors the specialized knowledge domains inherent in the models' training data. Participants were compensated at the standard rate for graduate research assistants.

To ensure data quality and minimize the impact of cognitive fatigue and time constraints, we curated a representative subset of the benchmark for the evaluation. Specifically, we randomly sampled 6 problems from each of the 12 task categories, resulting in a total of 72 problems per participant. Before commencing each task type, participants were briefed on the rules and completed several practice trials for familiarization. The evaluation protocol required participants to solve problems without the use of external aids (e.g., scratch paper, calculators), and they were allowed unlimited time per question. This approach was designed to emphasize and assess their intrinsic spatial visualization and mental manipulation capabilities, creating an evaluation condition comparable to assessing a model's internal reasoning processes without external memory aids. The reported human performance is the mean accuracy across all participants.

\subsection{Error Analysis}
\subsubsection{Model Selection for Direct Answer (Non-CoT) Evaluation}
Our Direct Answer evaluation tests model accuracy without induced reasoning chains. We excluded specific models based on 2 criteria:

\begin{enumerate}[leftmargin=1.2em]
    \vspace{-0.5em}
    \item \textbf{Reasoning-Centric Architectures:} Models explicitly designed for extended reasoning (e.g., o1, Gemini-2.5, Kimi-thinking, Llama-4 series) were excluded, as inhibiting CoT contradicts their core design principles.
    \item \textbf{Instruction Adherence:} Models unable to suppress reasoning traces despite strict formatting prompts (specifically InternVL3-2B) were excluded. This failure reflects a limitation in instruction following rather than reasoning capability.
    \vspace{-0.5em}
\end{enumerate}

Consequently, we retained only models capable of strictly adhering to the single-letter answer format. This exclusion criteria—based on format compliance rather than performance—ensures the baseline remains representative and uninflated.

\subsubsection{Error Types}
\label{Error Types}
\begin{enumerate}[leftmargin=1.2em]
\item \textbf{Perceptual Error:} Failure to perceive fundamental visual properties, such as color, shape, or pattern structures.
\item \textbf{Spatial Transformation Error:} Failure to deduce correct spatial states after a transformation. This includes:
\begin{enumerate}
\item Rotation/Flipping: Errors in angle or axis; confusing rotation with flipping.
\item Folding/Unfolding: Incorrect mapping between 2D nets and 3D cubes; confusing adjacent or opposite faces.
\item Spatial Relationships: Misjudging object composition, internal structure, or occlusion.
\end{enumerate}
\item \textbf{Spatial Memorization Error:} Forgetting or misremembering object positions or relationships across a sequence of operations.
\item \textbf{Instruction Following Error:} Misunderstanding textual instructions, such as task rules (e.g., negation) or required output formats.
\item \textbf{Methodological Error:} Adopting a flawed or suboptimal problem-solving strategy, such as using a rigid or unnecessarily complex reasoning path.
\item \textbf{Calculation and Reasoning Error:} Errors in non-spatial logic or mathematical calculations.
\end{enumerate}

\subsubsection{Inter-Annotator Agreement Analysis}
To ensure the reliability and reproducibility of our error taxonomy (detailed in Appendix~\ref{Error Types}), we conducted a rigorous inter-annotator agreement study.

\begin{table}[h]
\centering
\caption{\textbf{Inter-Annotator Agreement.} Cohen's $\kappa$ calculated via binary decomposition for multi-label error classification.}
\label{tab:iaa_main}
\setlength{\tabcolsep}{5pt} 
\resizebox{0.7\linewidth}{!}{%
    \begin{tabular}{l|cccccc|c}
    \toprule
    \textbf{Category} & \textbf{Perc.} & \textbf{Trans.} & \textbf{Meth.} & \textbf{Instr.} & \textbf{Memo.} & \textbf{Calc.} & \textbf{Avg.} \\
    \midrule
    \textbf{Cohen's $\kappa$} & 0.90 & 0.81 & 0.75 & 0.96 & 0.89 & 1.00 & \textbf{0.88} \\
    \bottomrule
    \end{tabular}%
}
\vspace{-0.2em}
\end{table}

\textbf{Methodology} Since our error analysis involves a multi-label classification task (i.e., a single failure case may stem from multiple error sources simultaneously), the traditional global Cohen's $\kappa$ is not directly applicable. Instead, we adopted a standard binary decomposition approach for multi-label agreement. Specifically, we decomposed the multi-label task into 6 independent binary classification tasks, treating each error category as a "Yes/No" decision.

\textbf{Calculation} We randomly sampled 100 failure cases from the evaluation set. Two authors independently annotated these cases based on the defined taxonomy. We then calculated Cohen's $\kappa$ separately for each error category. The results, presented in~\autoref{tab:iaa_main}, demonstrate high reliability. The Methodological category showed substantial agreement ($\kappa=0.75$), while all other categories achieved almost perfect agreement ($\kappa > 0.81$), with Calculation \& Reasoning reaching perfect consensus ($\kappa=1.00$). The macroscopic average Cohen's $\kappa$ across all categories is 0.8847, indicating an almost perfect level of inter-annotator consistency.

\section{Detailed Results}
In this section, we provide more evaluation results and test cases from Gemini-2.5-pro for each task.

\subsection{Intra-Category Comparisons Across Levels}
\label{intra-category_comparisons}
To provide deeper insight into the spatial visualization reasoning capabilities of Multi-modal Large Language Models (MLLMs), this section presents comprehensive experimental results that complement the aggregate performance assessment in Section \ref{5.2}. This analysis details the accuracy of each evaluated model across the four core sub-abilities—\textit{mental rotation, mental folding, visual penetration, and mental animation}—defined in the SpatialViz-Bench benchmark, with results stratified by task type and difficulty level. This granular performance breakdown reveals specific strengths and weaknesses of the models when confronting various spatial reasoning challenges, offering targeted insights to guide future model improvements.

\subsubsection{Mental Rotation}
\autoref{tab:mental_rotation} documents model performance on 3 sub-tasks within the mental rotation category—2D Rotation (2DR), 3D Rotation (3DR), and 3-View Projection (3VP)—across different difficulty levels. 

In the 2D Rotation (2DR) task, several models demonstrate foundational capabilities at Level 0, with ol (72.5\%) and Gemini-2.5-pro (62.5\%) achieving notable results. As difficulty increases to Level 1, most models show performance decline, though leading models maintain relatively high accuracy (ol: 52.5\%, Gemini-2.5-pro: 42.5\%). 

For 3D Rotation (3DR), performance degradation with increased difficulty is more pronounced. At Level 0, ol (42.5\%) and Gemini-2.5-pro (45.0\%) perform adequately, but their accuracies decrease substantially to 15.0\% and 20.0\%, respectively, at Level 1. Many open-source models perform at or below random chance (25\%-30\%) at this higher difficulty level, highlighting the challenge of mental rotation in complex 3D space. 

Interestingly, the 3-View Projection (3VP) task reveals a different pattern: when transitioning from Level 0 (cube stacks) to Level 1 (DeepCAD engineering models), some top-tier models like ol (improving from 40.0\% to 58.0\%) and Gemini-2.5-pro (increasing from 28.0\% to 66.0\%) demonstrate enhanced performance. This suggests certain Level 1 image features may be more amenable to these models' processing mechanisms, despite the presumed increase in complexity. Nevertheless, many other models show decreased performance from Level 0 to Level 1 in this sub-task. Overall, mental rotation tasks reveal a clear performance gradient across dimensions and geometric complexity while highlighting significant capability variations among model families.

\subsubsection{Mental Folding}
\autoref{tab:mental_folding} documents model performance on 3 sub-tasks within the mental folding category—Paper Folding (PF), Cube Unfolding (CU), and Cube Reconstruction (CR)—at varying difficulty levels. These tasks assess models' capacity for continuous reasoning and dynamic visualization of 3D information throughout transformation processes.  

In the Paper Folding (PF) task, as folding steps and hole-punching complexity increase (Level 0 to Level 2), most models perform near random chance, indicating significant challenges in tracking multi-step geometric operations and performing subsequent spatial reasoning. 

The more complex Cube Unfolding (CU) and Cube Reconstruction (CR) tasks proved challenging for all models. These tasks require understanding the correspondence between 2D nets and 3D cubes, while also assessing the ability to mentally execute folding operations and continuously reason about transforming 3D structures. Even at Level 0, most models demonstrate low accuracy, often below random chance. In the CU task, Gemini-2.5-pro scored 37.5\% (L0), 27.5\% (L1), and 30.0\% (L2), while ol achieved 37.5\% (L0), 37.5\% (L1), and 27.5\% (L2). 

For CR, Gemini-2.5-pro performed at 45.0\% (L0), 10.0\% (L1), and 35.0\% (L2), and ol at 42.5\% (L0), 12.5\% (L1), and 25.0\% (L2), both experiencing significant performance drops at Level 1. However, the surprising performance improvement at Level 2 contradicts human intuition, as Level 2 patterns are objectively more complex for humans. Analysis of sample solutions reveals that models approached these tasks by employing clear textual descriptions to define patterns composed of differently colored dots, representing their positions in matrix form. Conversely, line patterns proved more challenging for models to describe, and internal rotations could not be easily represented through matrix transposition operations, which . This insight provides valuable direction for designing more challenging tests that effectively evaluate model limitations. The overall results reveal a severe deficiency in reasoning and visualization capabilities when finer-grained correspondence and transformation tracking are required. The introduction of asymmetric patterns further challenges models' ability to maintain precise visual perception and spatial-topological understanding. These results highlight current MLLMs' core weaknesses in handling spatial tasks involving geometric correspondence, topological transformations, and dynamic 3D reasoning.

\subsubsection{Visual Penetration}
\autoref{tab:visual_penetration} documents model performance on 3 sub-tasks within the Visual Penetration category—Cross-Section (CS), Cube Counting (CC), and Cube Assembly (CA)—at varying difficulty levels. This ability requires models to infer internal object structures from visible external features. 

In the Cross-Section (CS) task, which requires models to visualize sectional shapes produced by cutting composite geometric solids with various planes, Gemini-2.5-pro and ol maintained relatively stable performance across Levels 0, 1, and 2, while most other models performed near random chance. 

For the Cube Counting (CC) task, increasing constraints from two-view (Level 0) to three-view (Level 1), and subsequently expanding spatial dimensions (Level 2), progressively challenged models' view integration and counting inference capabilities. Gemini-2.5-pro's accuracy declined sharply from 80.0\% (L0) to 52.5\% (L1) and 32.5\% (L2). Interestingly, ol's performance followed a pattern of 45.0\% (L0), 32.5\% (L1), and 45.0\% (L2), recovering at Level 2 to match its Level 0 score. Most models struggled to effectively integrate multi-view information in this task. 

The Cube Assembly (CA) task, which assesses the ability to identify complementary parts forming a complete structure, showed increasing difficulty as structures enlarged and constituent parts increased (Level 0 to Level 1). For example, Gemini-2.5-pro's accuracy dropped from 45.0\% (L0) to 27.5\% (L1), and ol's from 35.0\% (L0) to 32.5\% (L1). Collectively, these results reveal current models' limitations in inferring global internal structures and spatial occupancy from local surface information.

\subsubsection{Mental Animation}
\autoref{tab:mental_animation} documents model performance on 3 sub-tasks within the Mental Animation category—Arrow Moving (AM), Block Moving (BM), and Mechanical System (MS)—at varying difficulty levels. These tasks assess understanding of dynamic state changes and causal propagation among system components. 

In the Arrow Moving (AM) task, which requires understanding ego-centric movement rules and tracking state changes, the transition from simple single-arrow movements (Level 0) to multi-arrow environments involving swaps (Level 1) increasingly challenges models' rule comprehension and state tracking. A notable performance disparity exists between closed-source models (e.g., Gemini-2.5-pro and ol) and open-source counterparts: the former maintain high accuracy across both difficulty levels (almost 100\% accuracy by Gemini-2.5-pro), while most open-source models perform significantly worse (near random), particularly in complex multi-arrow Level 1 scenarios. This suggests a capability gap, potentially stemming from differences in architecture or training data, when precise instruction following and multi-step dynamic spatial reasoning are required. 

The Block Moving (BM) task combines directional movement with gravity simulation, increasing spatial complexity and operational sequence length, thereby challenging models' intuitive physics and 3D dynamic spatial reasoning. Gemini-2.5-pro's accuracy declined sharply from 95\% to 35\%, showing the difficulty in dealing with 3D scene. 

For the Mechanical System (MS) task, which evaluates understanding of motion transmission and component linkage in complex mechanical systems, questions were designed to minimize reliance on formal physics formulas while emphasizing comprehension through observation and spatial imagination. Interestingly, some open-source models performed better than expected based on their performance in other 3D imagination tasks. This suggests these models may transform such problems into more formalized reasoning processes similar to physical rule application, rather than relying solely on intuitive 3D mental simulation. While this strategy may yield relatively good scores in certain instances, it potentially deviates from the primary goal of assessing pure spatial visualization capabilities. Overall, mental animation tasks—especially those involving complex dynamic interactions and implicit physical laws—continue to pose significant challenges for current MLLMs, with models exhibiting considerable diversity in performance strategies and capabilities.

\subsection{Performance Comparison between Different Question Format}
\label{ComparisonFormat}
This benchmark primarily uses MCAs, a deliberate and justified design choice. MCAs are particularly effective for tasks with complex answers (e.g., 3D Rotation Task) that are difficult to express textually or match automatically. Moreover, well-crafted distractors can increase task difficulty and test a model's fine-grained discrimination.

Our rationale for using the MCA format is threefold:

\begin{itemize}[leftmargin=1.2em]
    \item MCAs align with human qualitative intuition. Humans often rely on estimation rather than precise calculation in spatial reasoning. This format assesses a model's grasp of core transformation logic ("qualitatively correct" reasoning) without penalizing minor deviations.
    \item Converting some tasks to a direct-answer format is technically challenging. For instance, in 3D Rotation and Paper Folding, the answers are complex images. Requiring models to generate these images is a frontier research problem beyond the scope of current multimodal evaluation.
    \item We quantitatively measured the difficulty gap. When the Cube Counting task was converted to a fill-in-the-blank format, all models showed a significant performance drop. As shown in \autoref{tab:cube_counting_drop}, GPT-4o’s accuracy dropped by 32.50\%, while even the top-performing Gemini-2.5-pro’s declined by 14.17\%. This indicates the direct-answer format is more demanding of a model's independent reasoning, even with options like "All three other options are incorrect" to reduce guessing. Consequently, for a comprehensive assessment, we provide both formats for the Cube Counting and parts of the Cube Reconstruction tasks. This performance gap demonstrates that MCA options provide clues or "error-correction" opportunities, helping models select a best-fit answer. In contrast, the direct-answer format more authentically exposes deficits in precise reasoning.
\end{itemize}

\begin{table}[ht]
\centering
\caption{Performance Drop on Cube Counting: Multiple-Choice vs. Fill-in-the-Blank.
The "Performance Drop" column quantifies the accuracy degradation when switching from the discriminative (Multiple-Choice) to the more challenging generative (Fill-in-the-Blank) task format.}
\label{tab:cube_counting_drop}
\resizebox{\textwidth}{!}{%
\renewcommand{\arraystretch}{1.2}
\begin{tabular}{@{}l|ccc|ccc|c@{}}
\toprule
\multirow{2}{*}{\textbf{Model}} & \multicolumn{3}{c|}{\textbf{Multiple-Choice Acc. (\%)}} & \multicolumn{3}{c|}{\textbf{Fill-in-the-Blank Acc. (\%)}} & \multicolumn{1}{c}{\textbf{Avg Performance}} \\
\cmidrule(lr){2-4} \cmidrule(lr){5-7}
& \textbf{L0} & \textbf{L1} & \textbf{L2} & \textbf{L0} & \textbf{L1} & \textbf{L2} & \multicolumn{1}{c}{\textbf{Drop (\%)}} \\
\midrule
\multicolumn{8}{c}{\textit{Open Source Models}} \\
\midrule
Qwen2.5-VL-7B-Instruct   & 32.50 & 50.00 & 27.50 & 15.00 & 2.50  & 0.00  & -30.83 \\
Qwen2.5-VL-72B-Instruct  & 32.50 & 50.00 & 42.50 & 25.00 & 32.50 & 5.00  & -20.83 \\
\midrule
\multicolumn{8}{c}{\textit{Closed Source Models}} \\
\midrule
GPT-4o                   & 40.00 & 45.00 & 37.50 & 10.00 & 12.50 & 2.50  & -32.50 \\
o1                       & 45.00 & 32.50 & 45.00 & 20.51 & 22.50 & 10.00 & -23.16 \\
Gemini-2.5-pro           & 80.00 & 52.50 & 32.50 & 55.00 & 52.50 & 15.00 & \textbf{-14.17} \\
\bottomrule
\end{tabular}%
}
\end{table}

\clearpage
\begin{table}[ht]
\centering
\caption{Comparison of model performances on Mental Rotation tasks. The \captionbest{first} and \captionsecond{second} highest accuracy of MLLMs are marked in red and blue, with open-source and closed-source models marked separately.}
\label{tab:mental_rotation} 
\resizebox{0.95\linewidth}{!}{%
    \renewcommand{\arraystretch}{1.4}
    \begin{tabular}{l|c|ccc|ccc|ccc} 
    \toprule
    \multirow{2}{*}{\textbf{Model}} & \multirow{2}{*}{\textbf{Overall}} & \multicolumn{3}{c|}{\textbf{2DRotation}} & \multicolumn{3}{c|}{\textbf{3DRotation}} & \multicolumn{3}{c}{\textbf{3ViewProjection}} \\ 
    \cline{3-11} 
    & & {L0} & {L1} & {Avg} & {L0} & {L1} & {Avg} & {L0} & {L1} & {Avg} \\ 
    \midrule
    Human & 85.56 & 92.50 & 87.50 & 90.00 & 83.33 & 75.00 & 79.17 & 91.67 & 83.33 & 87.50 \\
    \midrule
    Random & 27.69 & 25.00 & 22.50 & 23.75 & 25.00 & 30.00 & 27.50 & 30.00 & 32.00 & 31.00 \\
    \hline
    \multicolumn{11}{c}{\textbf{Open Source MLLMs}} \\ 
    \hline
    \multicolumn{11}{c}{3B} \\
    \hline
    SAIL-VL-1.5-2B & 22.31 & 20.00 & 25.00 & 22.50 & 17.50 & 27.50 & 22.50 & 20.00 & 24.00 & 22.00 \\
    InternVL3-2B & 27.31 & 12.50 & 20.00 & 16.25 & 32.50 & 35.00 & 33.75 & 24.00 & 38.00 & 31.00 \\
    Deepseek-VL2-tiny(3B) & 22.69 & 10.00 & 25.00 & 17.50 & 20.00 & 25.00 & 22.50 & 22.00 & 32.00 & 27.00 \\
    Qwen2.5-VL-3B-Instruct & 20.00 & 25.00 & 15.00 & 20.00 & 15.00 & 22.50 & 18.75 & 16.00 & 26.00 & 21.00 \\
    \hline
    \multicolumn{11}{c}{7B} \\ 
    \hline
    Qwen2.5-VL-7B-Instruct & 23.85 & 25.00 & 25.00 & 25.00 & 20.00 & 12.50 & 16.25 & 14.00 & 44.00 & 29.00 \\
    Qwen2.5-Omni-7B & 24.23 & 32.50 & 12.50 & 22.50 & 25.00 & 15.00 & 20.00 & 22.00 & 36.00 & 29.00 \\
    SAIL-VL-1.6-8B & 21.92 & 25.00 & 12.50 & 18.75 & 27.50 & 15.00 & 21.25 & 24.00 & 26.00 & 25.00 \\
    InternVL3-8B & 28.85 & 22.50 & 17.50 & 20.00 & 35.00 & \markbest{42.50} & \marksecond{38.75} & 18.00 & 38.00 & 28.00 \\
    \hline
    \multicolumn{11}{c}{16B} \\ 
    \hline
    Kimi-VL-A3B-Instruct(16B) & 28.08 & 15.00 & 17.50 & 16.25 & 32.50 & 27.50 & 30.00 & 24.00 & 48.00 & 36.00 \\
    Kimi-VL-A3B-thinking(16B) & 20.00 & 10.00 & 17.50 & 13.75 & 17.50 & 22.50 & 20.00 & 20.00 & 30.00 & 25.00 \\
    Deepseek-VL2-small(16B) & 24.62 & \markbest{40.00} & 22.50 & \marksecond{31.25} & 10.00 & 22.50 & 16.25 & 22.00 & 30.00 & 26.00 \\
    \hline
    \multicolumn{11}{c}{32B} \\ 
    \hline
    Deepseek-VL2(27B) & 29.62 & 20.00 & \marksecond{30.00} & 25.00 & 35.00 & 32.50 & 33.75 & 20.00 & 40.00 & 30.00 \\
    Qwen2.5-VL-32B-Instruct & \marksecond{35.00} & \marksecond{35.00} & 27.50 & \marksecond{31.25} & 32.50 & \marksecond{37.50} & 35.00 & 22.00 & \marksecond{54.00} & 38.00 \\
    InternVL3-38B & 28.46 & 25.00 & 20.00 & 22.50 & 32.50 & 35.00 & 33.75 & 22.00 & 36.00 & 29.00 \\
    \hline
    \multicolumn{11}{c}{72B} \\ 
    \hline
    Qwen2.5-VL-72B-Instruct & 29.23 & 25.00 & \markbest{32.50} & 28.75 & \marksecond{40.00} & 22.50 & 31.25 & 22.00 & 34.00 & 28.00 \\
    QvQ-72B-preview & 27.69 & 15.00 & 27.50 & 21.25 & 27.50 & 32.50 & 30.00 & \markbest{32.00} & 30.00 & 31.00 \\
    InternVL3-78B & 28.46 & 20.00 & \marksecond{30.00} & 25.00 & 25.00 & 25.00 & 25.00 & 20.00 & 48.00 & 34.00 \\
    \hline
    \multicolumn{11}{c}{108B} \\ 
    \hline
    Llama-4-Maverick-17B-128E-Instruct & 33.85 & 25.00 & 15.00 & 20.00 & \markbest{45.00} & 35.00 & \markbest{40.00} & 26.00 & \marksecond{54.00} & \marksecond{40.00} \\
    LLama-4-Scout-17B-16E-Instruct & \markbest{37.31} & 32.50 & \markbest{32.50} & \markbest{32.50} & 32.50 & \marksecond{37.50} & 35.00 & \marksecond{28.00} & \markbest{58.00} & \markbest{43.00} \\
        \hline
    \multicolumn{11}{c}{\textbf{Closed Source MLLMs}} \\ 
    \hline
    GPT-4o & 31.15 & 20.00 & \marksecond{45.00} & 32.50 & 30.00 & 25.00 & 27.50 & 20.00 & 46.00 & 33.00 \\
    o1 & \markbest{46.92} & \markbest{72.50} & \markbest{52.50} & \markbest{62.50} & \marksecond{42.50} & 15.00 & 28.75 & \markbest{40.00} & 58.00 & \markbest{49.00} \\
    Claude-3.5-sonnet & 34.62 & 27.50 & 35.00 & 31.25 & 32.50 & 17.50 & 25.00 & \marksecond{36.00} & 54.00 & 45.00 \\
    Claude-3.7-sonnet & 38.08 & 40.00 & 25.00 & 32.50 & 40.00 & \markbest{32.50} & \markbest{36.25} & 34.00 & 54.00 & 44.00 \\
    Gemini-2.5-flash & 35.77 & 55.00 & 30.00 & 42.50 & 40.00 & 20.00 & 30.00 & 18.00 & 52.00 & 35.00 \\
    Gemini-2.5-pro & \marksecond{44.23} & \marksecond{62.50} & 42.50 & \marksecond{52.50} & \markbest{45.00} & 20.00 & 32.50 & 28.00 & \markbest{66.00} & \marksecond{47.00} \\
    Doubao-1-5-vision-pro & 30.38 & 7.50 & 7.50 & 7.50 & \marksecond{42.50} & 
    \marksecond{27.50} & \marksecond{35.00} & 28.00 & \marksecond{62.00} & 45.00 \\
    Qwen-VL-max & 28.08 & 12.50 & 35.00 & 23.75 & 30.00 & 22.50 & 26.25 & 22.00 & 44.00 & 33.00 \\
    \bottomrule
    \end{tabular}
}
\end{table}

\clearpage
\begin{table}[ht]
\centering
\caption{Comparison of model performances on Mental Folding tasks.}
\label{tab:mental_folding} 
\resizebox{\linewidth}{!}{%
    \renewcommand{\arraystretch}{1.4}
    \begin{tabular}{l|c|cccc|cccc|cccc} 
    \toprule
    \multirow{2}{*}{\textbf{Model}} & \multirow{2}{*}{\textbf{Overall}} & \multicolumn{4}{c|}{\textbf{PaperFolding}} & \multicolumn{4}{c|}{\textbf{CubeUnfolding}} & \multicolumn{4}{c}{\textbf{CubeReconstruction}} \\ 
    \cline{3-14} 
    & & {L0} & {L1}& {L2} & {Avg} & {L0} & {L1}& {L2} & {Avg} & {L0} & {L1}& {L2} & {Avg} \\ 

    \midrule
    Human & 80.56 & 100.00 & 93.75 & 87.50 & 93.75 & 87.50 & 75.00 & 62.50 & 75.00 & 81.25 & 75.00 & 62.50 & 72.92 \\
    \midrule
    Random & 21.67 & 17.50 & 20.00 & 20.00 & 19.17 & 15.00 & 27.50 & 17.50 & 20.00 & 30.00 & 25.00 & 22.50 & 25.83 \\
    \hline
    \multicolumn{14}{c}{\textbf{Open Source}} \\ 
    \hline
    \multicolumn{14}{c}{3B} \\
    \hline
    SAIL-VL-1.5-2B & 22.50 & 12.50 & 25.00 & 22.50 & 20.00 & 30.00 & 27.50 & 25.00 & 27.50 & 22.50 & 20.00 & 17.50 & 20.00 \\
    InternVL3-2B & 24.44 & 25.00 & 27.50 & 15.00 & 22.50 & \markbest{35.00} & 12.50 & 30.00 & 25.83 & \marksecond{35.00} & 22.50 & 17.50 & 25.00 \\
    Deepseek-VL2-tiny(3B) & 20.56 & 27.50 & 17.50 & 20.00 & 21.67 & 20.00 & 25.00 & 17.50 & 20.83 & 15.00 & 20.00 & 22.50 & 19.17 \\
    Qwen2.5-VL-3B-Instruct & 24.17 & 20.00 & \markbest{37.50} & 17.50 & 25.00 & 25.00 & 25.00 & 27.50 & 25.83 & 25.00 & \markbest{32.50} & 7.50 & 21.67 \\
    \hline
    \multicolumn{14}{c}{7B} \\ 
    \hline
    Qwen2.5-VL-7B-Instruct & \markbest{28.61} & \markbest{35.00} & \marksecond{35.00} & 32.50 & \markbest{34.17} & 17.50 & 30.00 & 17.50 & 21.67 & 27.50 & \marksecond{30.00} & 32.50 & 30.00 \\
    Qwen2.5-Omni-7B & 24.17 & 27.50 & 30.00 & 17.50 & 25.00 & \marksecond{32.50} & \markbest{37.50} & 12.50 & 27.50 & 17.50 & 27.50 & 15.00 & 20.00 \\
    SAIL-VL-1.6-8B & 23.89 & \markbest{35.00} & 17.50 & 32.50 & 28.33 & 25.00 & 30.00 & 20.00 & 25.00 & 17.50 & 25.00 & 12.50 & 18.33 \\
    InternVL3-8B & 25.56 & 25.00 & 20.00 & \markbest{40.00} & 28.33 & 25.00 & 20.00 & 25.00 & 23.33 & 25.00 & 27.50 & 22.50 & 25.00 \\
    \hline
    \multicolumn{14}{c}{16B} \\ 
    \hline
    Kimi-VL-A3B-Instruct(16B) & 24.17 & \marksecond{27.50} & 22.50 & 27.50 & 25.83 & 22.50 & 15.00 & 22.50 & 20.00 & 15.00 & 27.50 & \marksecond{37.50} & 26.67 \\
    Kimi-VL-A3B-thinking(16B) & 24.72 & 10.00 & 25.00 & 35.00 & 23.33 & 20.00 & 20.00 & \markbest{32.50} & 24.17 & \marksecond{35.00} & 17.50 & 27.50 & 26.67 \\
    Deepseek-VL2-small(16B) & 24.72 & 25.00 & 22.50 & 20.00 & 22.50 & 27.50 & 25.00 & 22.50 & 25.00 & 22.50 & 25.00 & 32.50 & 26.67 \\
    \hline
    \multicolumn{14}{c}{32B} \\ 
    \hline
    Deepseek-VL2(27B) & 26.39 & 22.50 & \marksecond{35.00} & \marksecond{37.50} & \marksecond{31.67} & \marksecond{32.50} & 15.00 & 27.50 & 25.00 & 17.50 & \marksecond{30.00} & 20.00 & 22.50 \\
    Qwen2.5-VL-32B-Instruct & 24.72 & 15.00 & \markbest{37.50} & 12.50 & 21.67 & 17.50 & \marksecond{35.00} & 22.50 & 25.00 & 30.00 & 10.00 & \markbest{42.50} & 27.50 \\
    InternVL3-38B & \marksecond{26.94} & 22.50 & 20.00 & 20.00 & 20.83 & 25.00 & \marksecond{35.00} & 27.50 & \marksecond{29.17} & 22.50 & \markbest{32.50} & \marksecond{37.50} & \marksecond{30.83} \\
    \hline
    \multicolumn{14}{c}{72B} \\ 
    \hline
    Qwen2.5-VL-72B-Instruct & 24.17 & 12.50 & 27.50 & 27.50 & 22.50 & 15.00 & 17.50 & 27.50 & 20.00 & 30.00 & 25.00 & 35.00 & 30.00 \\
    QvQ-72B-preview & 21.11 & 15.00 & 12.50 & 22.50 & 16.67 & 22.50 & 15.00 & 20.00 & 19.17 & 30.00 & 25.00 & 27.50 & 27.50 \\
    InternVL3-78B & 22.22 & 15.00 & 30.00 & 12.50 & 19.17 & \markbest{35.00} & 22.50 & 17.50 & 25.00 & 30.00 & 20.00 & 17.50 & 22.50 \\
    \hline
    \multicolumn{14}{c}{108B} \\ 
    \hline
    Llama-4-Maverick-17B-128E-Instruct & 25.00 & 15.00 & 17.50 & 17.50 & 16.67 & 30.00 & 25.00 & \markbest{32.50} & \marksecond{29.17} & 30.00 & \markbest{32.50} & 25.00 & 29.17 \\
    LLama-4-Scout-17B-16E-Instruct & \markbest{28.61} & 15.00 & 17.50 & 17.50 & 16.67 & \markbest{35.00} & 32.50 & \marksecond{30.00} & \markbest{32.50} & \markbest{42.50} & \markbest{32.50} & 35.00 & \markbest{36.67} \\
    \hline
    \multicolumn{14}{c}{\textbf{Closed Source}} \\ 
    \hline
    GPT-4o & 25.00 & 25.00 & 35.00 & \marksecond{27.50} & 29.17 & 25.00 & 12.50 & 10.00 & 15.83 & 30.00 & 17.50 & \markbest{42.50} & 30.00 \\
    o1 & 29.72 & \marksecond{27.50} & 30.00 & \marksecond{27.50} & 28.33 & \markbest{37.50} & \markbest{37.50} & \marksecond{27.50} & \markbest{34.17} & 42.50 & 12.50 & 25.00 & 26.67 \\
    Claude-3.5-sonnet & 25.00 & 7.50 & 35.00 & 20.00 & 20.83 & 25.00 & 17.50 & 25.00 & 22.50 & 32.50 & 20.00 & \markbest{42.50} & \marksecond{31.67} \\
    Claude-3.7-sonnet & 24.72 & 20.00 & 20.00 & 15.00 & 18.33 & \marksecond{32.50} & 25.00 & 22.50 & 26.67 & 32.50 & 17.50 & 37.50 & 29.17 \\
    Gemini-2.5-flash & \marksecond{32.50} & 15.00 & \marksecond{37.50} & \marksecond{27.50} & 26.67 & \marksecond{32.50} & \marksecond{30.00} & \marksecond{27.50} & 30.00 & \markbest{55.00} & \markbest{27.50} & \marksecond{40.00} & \markbest{40.83} \\
    Gemini-2.5-pro & \markbest{35.00} & \markbest{57.50} & \markbest{40.00} & \markbest{32.50} & \markbest{43.33} & \markbest{37.50} & 27.50 & \markbest{30.00} & \marksecond{31.67} & \marksecond{45.00} & 10.00 & 35.00 & 30.00 \\
    Doubao-1-5-vision-pro & 28.06 & 25.00 & \marksecond{37.50} & \markbest{32.50} & \marksecond{31.67} & 22.50 & 22.50 & 25.00 & 23.33 & \marksecond{45.00} & 17.50 & 25.00 & 29.17 \\
    Qwen-VL-max & 24.44 & \marksecond{27.50} & 25.00 & 20.00 & 24.17 & 12.50 & 15.00 & 25.00 & 17.50 & 42.50 & 22.50 & 30.00 & \marksecond{31.67} \\
    \bottomrule
    \end{tabular}
}
\end{table}

\clearpage
\begin{table}[ht]
\centering
\caption{Comparison of model performances on Visual Penetration tasks.}
\label{tab:visual_penetration} 
\resizebox{\linewidth}{!}{%
    \renewcommand{\arraystretch}{1.4}
    \begin{tabular}{l|c|cccc|cccc|ccc} 
    \toprule
    \multirow{2}{*}{\textbf{Model}} & \multirow{2}{*}{\textbf{Overall}} & \multicolumn{4}{c|}{\textbf{CrossSection}} & \multicolumn{4}{c|}{\textbf{CubeCounting}} & \multicolumn{3}{c}{\textbf{CubeAssembly}} \\ 
    \cline{3-13} 
    & & {L0} & {L1}& {L2} & {Avg} & {L0} & {L1}& {L2} & {Avg} & {L0} & {L1} & {Avg} \\ 
    \midrule
    Human & 75.42 & 75.00 & 75.00 & 68.75 &  72.92 & 81.25 & 75.00 & 56.25 & 70.83 & 87.50 &  75.00 & 82.50 \\
    \midrule
    Random & 28.12 & 32.50 & 27.50 & 30.00 & {30.00} & 30.00 & 20.00 & 25.00 & 25.00 & 22.50 & 37.50 & 30.00 \\
    \hline
    \multicolumn{13}{c}{\textbf{Open Source}} \\ 
    \hline
    \multicolumn{13}{c}{3B} \\
    \hline
    SAIL-VL-1.5-2B & 27.19 & \markbest{37.50} & 20.00 & 15.00 & 24.17 & \marksecond{40.00} & 20.00 & 20.00 & 26.67 & 32.50 & 32.50 & 32.50 \\
    InternVL3-2B & 26.56 & 22.50 & 22.50 & 15.00 & 20.00 & 22.50 & 32.50 & 37.50 & 30.83 & 27.50 & 32.50 & 30.00 \\
    Deepseek-VL2-tiny(3B) & 20.94 & 17.50 & 25.00 & 20.00 & 20.83 & 25.00 & 25.00 & 17.50 & 22.50 & 17.50 & 20.00 & 18.75 \\
    Qwen2.5-VL-3B-Instruct & 25.94 & 25.00 & 25.00 & \marksecond{27.50} & \marksecond{25.83} & 17.50 & 35.00 & 17.50 & 23.33 & 30.00 & 30.00 & 30.00 \\
    \hline
    \multicolumn{13}{c}{7B} \\ 
    \hline
    Qwen2.5-VL-7B-Instruct & 27.19 & 12.50 & 12.50 & 25.00 & 16.67 & 32.50 & \marksecond{50.00} & 27.50 & 36.67 & 35.00 & 22.50 & 28.75 \\
    Qwen2.5-Omni-7B & 27.19 & 15.00 & 22.50 & 25.00 & 20.83 & 37.50 & 27.50 & 35.00 & 33.33 & 25.00 & 30.00 & 27.50 \\
    SAIL-VL-1.6-8B & 21.25 & 17.50 & 22.50 & 25.00 & 21.67 & 22.50 & 17.50 & 17.50 & 19.17 & 30.00 & 17.50 & 23.75 \\
    InternVL3-8B & 30.94 & 17.50 & 15.00 & 15.00 & 15.83 & 25.00 & 45.00 & \markbest{52.50} & \marksecond{40.83} & 45.00 & 32.50 & 38.75 \\
    \hline
    \multicolumn{13}{c}{16B} \\ 
    \hline
    Kimi-VL-A3B-Instruct(16B) & 17.19 & 17.50 & 25.00 & 22.50 & 21.67 & 7.50 & 2.50 & 5.00 & 5.00 & 27.50 & 30.00 & 28.75 \\
    Kimi-VL-A3B-thinking(16B) & 29.38 & 27.50 & 17.50 & \markbest{30.00} & 25.00 & \markbest{45.00} & 40.00 & 25.00 & 36.67 & 20.00 & 30.00 & 25.00 \\
    Deepseek-VL2-small(16B) & 25.31 & 7.50 & 12.50 & 7.50 & 9.17 & 30.00 & 32.50 & \marksecond{42.50} & 35.00 & 30.00 & 40.00 & 35.00 \\
    \hline
    \multicolumn{13}{c}{32B} \\ 
    \hline
    Kimi-VL-A3B-Instruct(16B) & 17.19 & 17.50 & 25.00 & 22.50 & 21.67 & 7.50 & 2.50 & 5.00 & 5.00 & 27.50 & 30.00 & 28.75 \\
    Kimi-VL-A3B-thinking(16B) & 29.38 & 27.50 & 17.50 & \markbest{30.00} & 25.00 & \markbest{45.00} & 40.00 & 25.00 & 36.67 & 20.00 & 30.00 & 25.00 \\
    Deepseek-VL2-small(16B) & 25.31 & 7.50 & 12.50 & 7.50 & 9.17 & 30.00 & 32.50 & \marksecond{42.50} & 35.00 & 30.00 & 40.00 & 35.00 \\
    \hline
    \multicolumn{13}{c}{72B} \\ 
    \hline
    Qwen2.5-VL-72B-Instruct & \markbest{39.06} & 27.50 & \markbest{40.00} & 22.50 & \markbest{30.00} & 32.50 & \marksecond{50.00} & \marksecond{42.50} & \markbest{41.67} & \markbest{55.00} & 42.50 & \marksecond{48.75} \\
    QvQ-72B-preview & 27.81 & \marksecond{32.50} & \marksecond{30.00} & \marksecond{27.50} & \markbest{30.00} & 35.00 & 25.00 & 7.50 & 22.50 & 40.00 & 25.00 & 32.50 \\
    InternVL3-78B & \marksecond{35.00} & 17.50 & 25.00 & 20.00 & 20.83 & 37.50 & \markbest{52.50} & 30.00 & 40.00 & 42.50 & \marksecond{55.00} & \marksecond{48.75} \\
    \hline
    \multicolumn{13}{c}{108B} \\ 
    \hline
    Llama-4-Maverick-17B-128E-Instruct & 32.19 & 27.50 & 15.00 & 15.00 & 19.17 & 27.50 & 47.50 & 30.00 & 35.00 & \marksecond{52.50} & 42.50 & 47.50 \\
    LLama-4-Scout-17B-16E-Instruct & 34.06 & 17.50 & 17.50 & 17.50 & 17.50 & 35.00 & 47.50 & 30.00 & 37.50 & 50.00 & \markbest{57.50} & \markbest{53.75} \\
    \hline
    \multicolumn{13}{c}{\textbf{Closed Source}} \\ 
    \hline
    GPT-4o & 32.50 & 25.00 & 25.00 & 7.50 & 19.17 & 40.00 & 45.00 & 37.50 & 40.83 & \markbest{52.50} & 27.50 & 40.00 \\
    o1 & 37.81 & \markbest{40.00} & \markbest{42.50} & \marksecond{30.00} & \markbest{37.50} & 45.00 & 32.50 & \markbest{45.00} & 40.83 & 35.00 & 32.50 & 33.75 \\
    Claude-3.5-sonnet & 33.44 & \marksecond{35.00} & 20.00 & 12.50 & 22.50 & 35.00 & 45.00 & 27.50 & 35.83 & 47.50 & \marksecond{45.00} & \markbest{46.25} \\
    Claude-3.7-sonnet & 31.56 & 20.00 & \marksecond{35.00} & 17.50 & 24.17 & 30.00 & 32.50 & 30.00 & 30.83 & 40.00 & \markbest{47.50} & \marksecond{43.75} \\
    Gemini-2.5-flash & 32.81 & 32.50 & \marksecond{35.00} & 22.50 & 30.00 & 52.50 & 32.50 & 30.00 & 38.33 & 30.00 & 27.50 & 28.75 \\
    Gemini-2.5-pro & \markbest{42.19} & 32.50 & \marksecond{35.00} & \markbest{32.50} & \marksecond{33.33} & \markbest{80.00} & 52.50 & 32.50 & \marksecond{55.00} & 45.00 & 27.50 & 36.25 \\
    Doubao-1-5-vision-pro & \marksecond{39.69} & \marksecond{35.00} & 30.00 & 25.00 & 30.00 & \marksecond{62.50} & \markbest{65.00} & \marksecond{40.00} & \markbest{55.83} & 42.50 & 17.50 & 30.00 \\
    Qwen-VL-max & 38.44 & 32.50 & 20.00 & 27.50 & 26.67 & 57.50 & \marksecond{62.50} & 22.50 & 47.50 & \marksecond{50.00} & 35.00 & 42.50 \\
    \bottomrule
    \end{tabular}
}
\end{table}

\clearpage
\begin{table}[ht]
\centering
\caption{Comparison of model performances on Mental Animation tasks.}
\label{tab:mental_animation} 
\resizebox{\linewidth}{!}{%
    \renewcommand{\arraystretch}{1.4}
    \begin{tabular}{l|c|ccc|ccc|ccc} 
    \toprule
    \multirow{2}{*}{\textbf{Model}} & \multirow{2}{*}{\textbf{Overall}} & \multicolumn{3}{c|}{\textbf{ArrowMoving}} & \multicolumn{3}{c|}{\textbf{BlockMoving}} & \multicolumn{3}{c}{\textbf{MechanicalSystem}} \\ 
    \cline{3-11} 
    & & {L0} & {L1} & {Avg} & {L0} & {L1} & {Avg} & {L0} & {L1} & {Avg} \\ 

    \midrule
    Human & 88.33 & 92.50& 87.5 & 90.00 & 95.83 & 79.16 & 87.5 & 87.50 & 87.50 & 87.50 \\
    \midrule
    Random & 23.33 & 32.50 & 25.00 & 28.75 & 10.00 & 22.50 & 16.25 & 30.00 & 20.00 & 25.00 \\
    \hline
    \multicolumn{11}{c}{\textbf{Open Source}} \\ 
    \hline
    \multicolumn{11}{c}{3B} \\ 
    \hline
    SAIL-VL-1.5-2B & 24.58 & 15.00 & 27.50 & 21.25 & 22.50 & 27.50 & 25.00 & 35.00 & 20.00 & 27.50 \\
    InternVL3-2B & 27.08 & 22.50 & 15.00 & 18.75 & 37.50 & 27.50 & 32.50 & 25.00 & 35.00 & 30.00 \\
    Deepseek-VL2-tiny(3B) & 21.67 & 25.00 & 12.50 & 18.75 & 25.00 & 17.50 & 21.25 & 25.00 & 25.00 & 25.00 \\
    Qwen2.5-VL-3B-Instruct & 35.83 & \markbest{35.00} & 35.00 & \markbest{35.00} & 32.50 & 27.50 & 30.00 & 57.50 & 27.50 & 42.50 \\
    \hline
    \multicolumn{11}{c}{7B} \\ 
    \hline
    Qwen2.5-VL-7B-Instruct & 32.50 & 22.50 & 22.50 & 22.50 & 22.50 & 25.00 & 23.75 & \markbest{67.50} & 35.00 & 51.25 \\
    Qwen2.5-Omni-7B & 35.42 & 27.50 & 35.00 & \marksecond{31.25} & 32.50 & 27.50 & 30.00 & \markbest{67.50} & 22.50 & 45.00 \\
    SAIL-VL-1.6-8B & 35.00 & 12.50 & \marksecond{37.50} & 25.00 & 37.50 & 32.50 & 35.00 & 52.50 & 37.50 & 45.00 \\
    InternVL3-8B & 37.08 & \marksecond{30.00} & 30.00 & 30.00 & 30.00 & 30.00 & 30.00 & 62.50 & 40.00 & 51.25 \\
    \hline
    \multicolumn{11}{c}{16B} \\ 
    \hline
    Kimi-VL-A3B-Instruct(16B) & 27.92 & 17.50 & 12.50 & 15.00 & 27.50 & 35.00 & 31.25 & 57.50 & 17.50 & 37.50 \\
    Kimi-VL-A3B-thinking(16B) & \marksecond{40.42} & 22.50 & \marksecond{37.50} & 30.00 & 35.00 & \marksecond{52.50} & \marksecond{43.75} & 62.50 & 32.50 & 47.50 \\
    Deepseek-VL2-small(16B) & 26.25 & 25.00 & 27.50 & 26.25 & 25.00 & 22.50 & 23.75 & 47.50 & 10.00 & 28.75 \\
    \hline
    \multicolumn{11}{c}{32B} \\ 
    \hline
    Deepseek-VL2(27B) & 29.17 & 20.00 & 32.50 & 26.25 & 35.00 & 25.00 & 30.00 & 40.00 & 22.50 & 31.25 \\
    Qwen2.5-VL-32B-Instruct & 37.08 & 22.50 & 35.00 & 28.75 & 27.50 & 27.50 & 27.50 & 62.50 & \marksecond{47.50} & 55.00 \\
    InternVL3-38B & 37.08 & 25.00 & 25.00 & 25.00 & 25.00 & 35.00 & 30.00 & \marksecond{65.00} & \marksecond{47.50} & \marksecond{56.25} \\
    \hline
    \multicolumn{11}{c}{72B} \\ 
    \hline
    Qwen2.5-VL-72B-Instruct & \markbest{43.75} & 27.50 & 27.50 & 27.50 & \marksecond{45.00} & 35.00 & 40.00 & \markbest{67.50} & \markbest{60.00} & \markbest{63.75} \\
    QvQ-72B-preview & 39.58 & 27.50 & 22.50 & 25.00 & 40.00 & \markbest{60.00} & \markbest{50.00} & 42.50 & 45.00 & 43.75 \\
    InternVL3-78B & 35.42 & 25.00 & 22.50 & 23.75 & 35.00 & 47.50 & 41.25 & 55.00 & 27.50 & 41.25 \\
    \hline
    \multicolumn{11}{c}{108B} \\
    \hline
    Llama-4-Maverick-17B-128E-Instruct & 39.17 & \markbest{35.00} & 35.00 & \markbest{35.00} & 40.00 & 40.00 & 40.00 & 45.00 & 40.00 & 42.50 \\
    LLama-4-Scout-17B-16E-Instruct & 39.58 & 15.00 & \markbest{42.50} & 28.75 & \markbest{47.50} & 32.50 & 40.00 & 57.50 & 42.50 & 50.00 \\
    \hline
    \multicolumn{11}{c}{\textbf{Closed Source}} \\ 
    \hline
    GPT-4o & 38.33 & 32.50 & 12.50 & 22.50 & 25.00 & 40.00 & 32.50 & 62.50 & \markbest{57.50} & \markbest{60.00} \\
    o1 & \marksecond{57.50} & 75.00 & \marksecond{60.00} & \marksecond{67.50} & \markbest{50.00} & \markbest{55.00} & \markbest{52.50} & 62.50 & 42.50 & 52.50 \\
    Claude-3.5-sonnet & 40.42 & 42.50 & 32.50 & 37.50 & 25.00 & 37.50 & 31.25 & 57.50 & \marksecond{47.50} & 52.50 \\
    Claude-3.7-sonnet & 46.25 & 72.50 & \marksecond{60.00} & 66.25 & 25.00 & 32.50 & 28.75 & 55.00 & 32.50 & 43.75 \\
    Gemini-2.5-flash & 50.00 & \marksecond{82.50} & 52.50 & \marksecond{67.50} & 32.50 & 35.00 & 33.75 & 62.50 & 35.00 & 48.75 \\
    Gemini-2.5-pro & \markbest{62.92} & \markbest{97.50} & \markbest{92.50} & \markbest{95.00} & 27.50 & \marksecond{42.50} & 35.00 & \markbest{75.00} & 42.50 & \marksecond{58.75} \\
    Doubao-1-5-vision-pro & 35.83 & 22.50 & 22.50 & 22.50 & 32.50 & \marksecond{42.50} & \marksecond{37.50} & 57.50 & 37.50 & 47.50 \\
    Qwen-VL-max & 39.17 & 25.00 & 27.50 & 26.25 & \marksecond{35.00} & 37.50 & 36.25 & \marksecond{65.00} & 45.00 & 55.00 \\
    \bottomrule
    \end{tabular}
}
\end{table}
\clearpage
\subsection{Test Cases}
\label{test_cases}
This section presents test cases from Gemini-2.5-pro across various tasks, helping us identify error sources and provide direction for model improvement.

\begin{figure}[ht]
    \centering
    \includegraphics[width=1\linewidth]{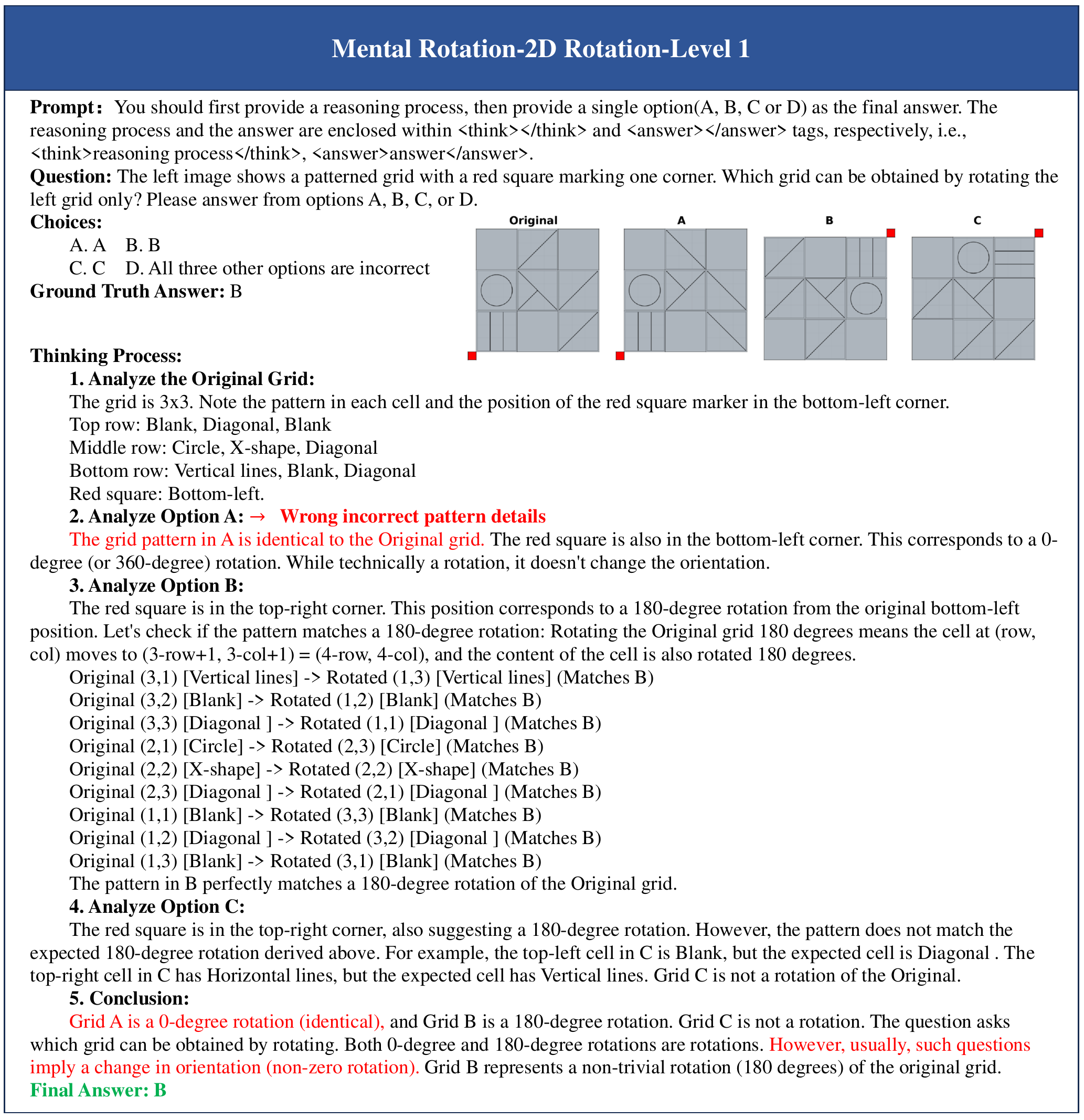}
    \caption{Test case of 2D Rotation Task from Gemini-2.5-pro.}
    \label{fig:Sample2DRotation}
\end{figure}

\newpage
\begin{figure}[ht]
    \centering
    \includegraphics[width=1\linewidth]{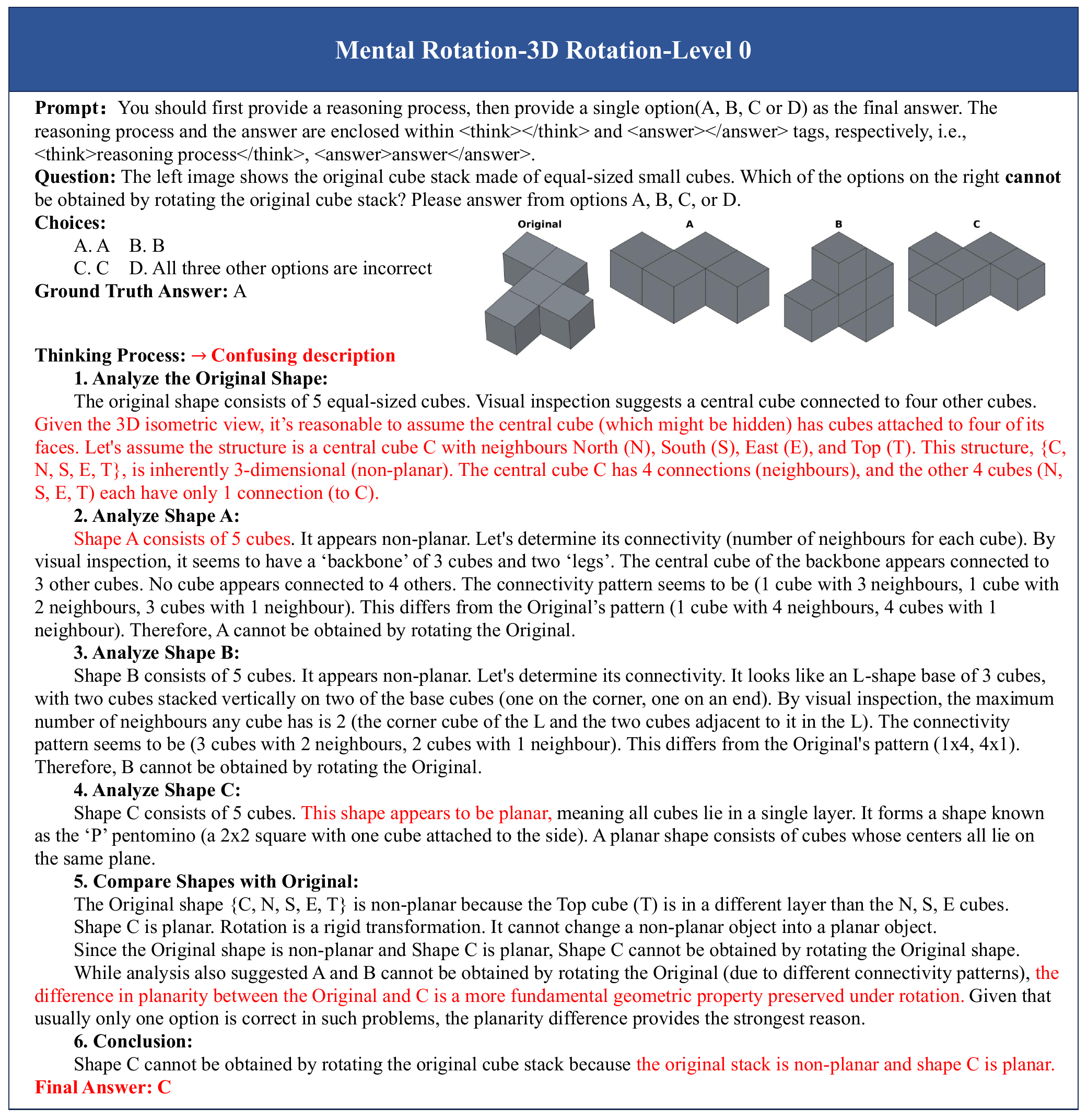}
    \caption{Test case of 3D Rotation Task from Gemini-2.5-pro.}
    \label{fig:Sample3DRotation}
\end{figure}

\newpage
\begin{figure}[ht]
    \centering
    \includegraphics[width=1\linewidth]{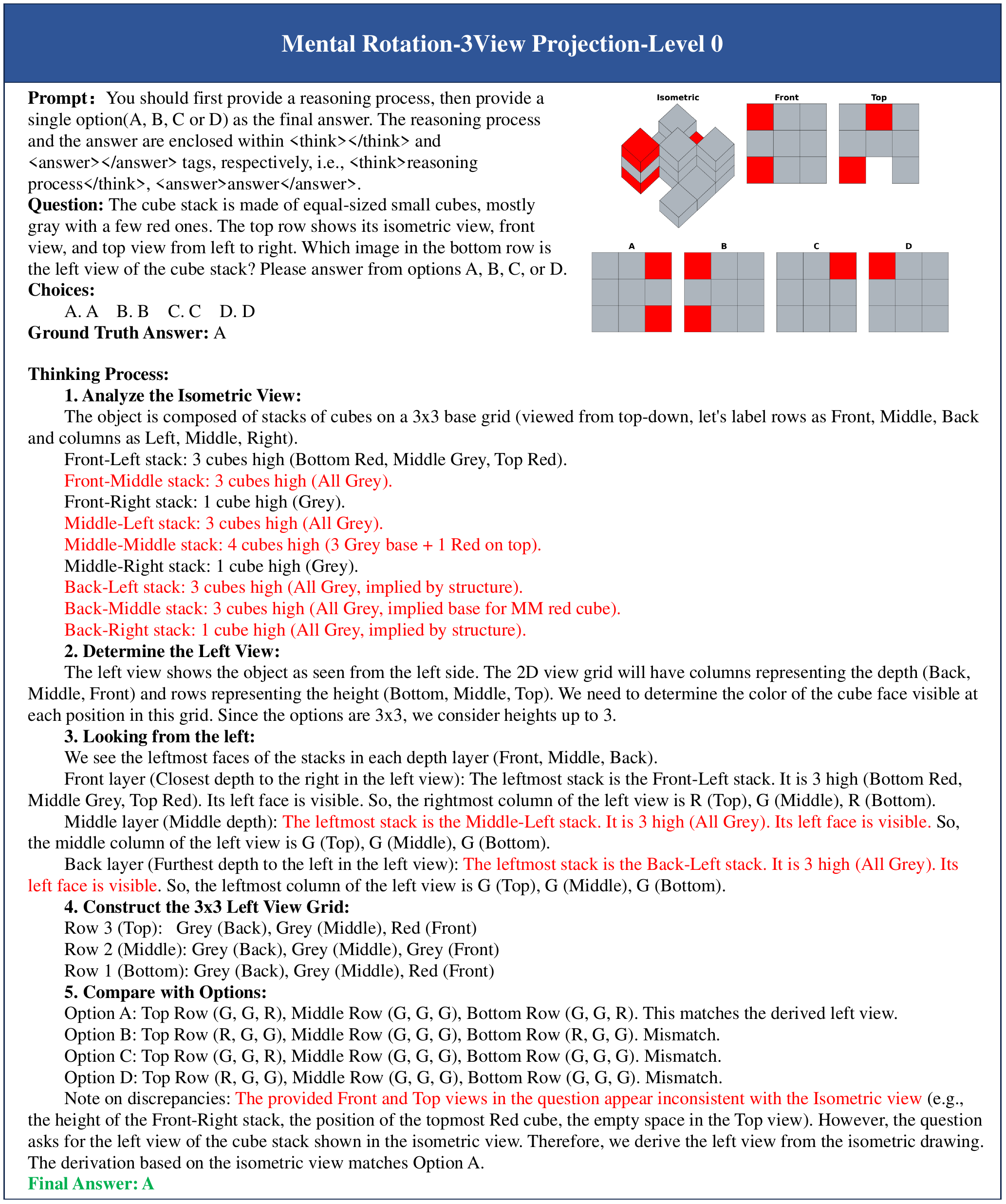}
    \caption{Test case of Three-View Projection (Level 0) Task from Gemini-2.5-pro.}
    \label{fig:Sample2DRotation}
\end{figure}

\newpage
\begin{figure}[ht]
    \centering
    \includegraphics[width=1\linewidth]{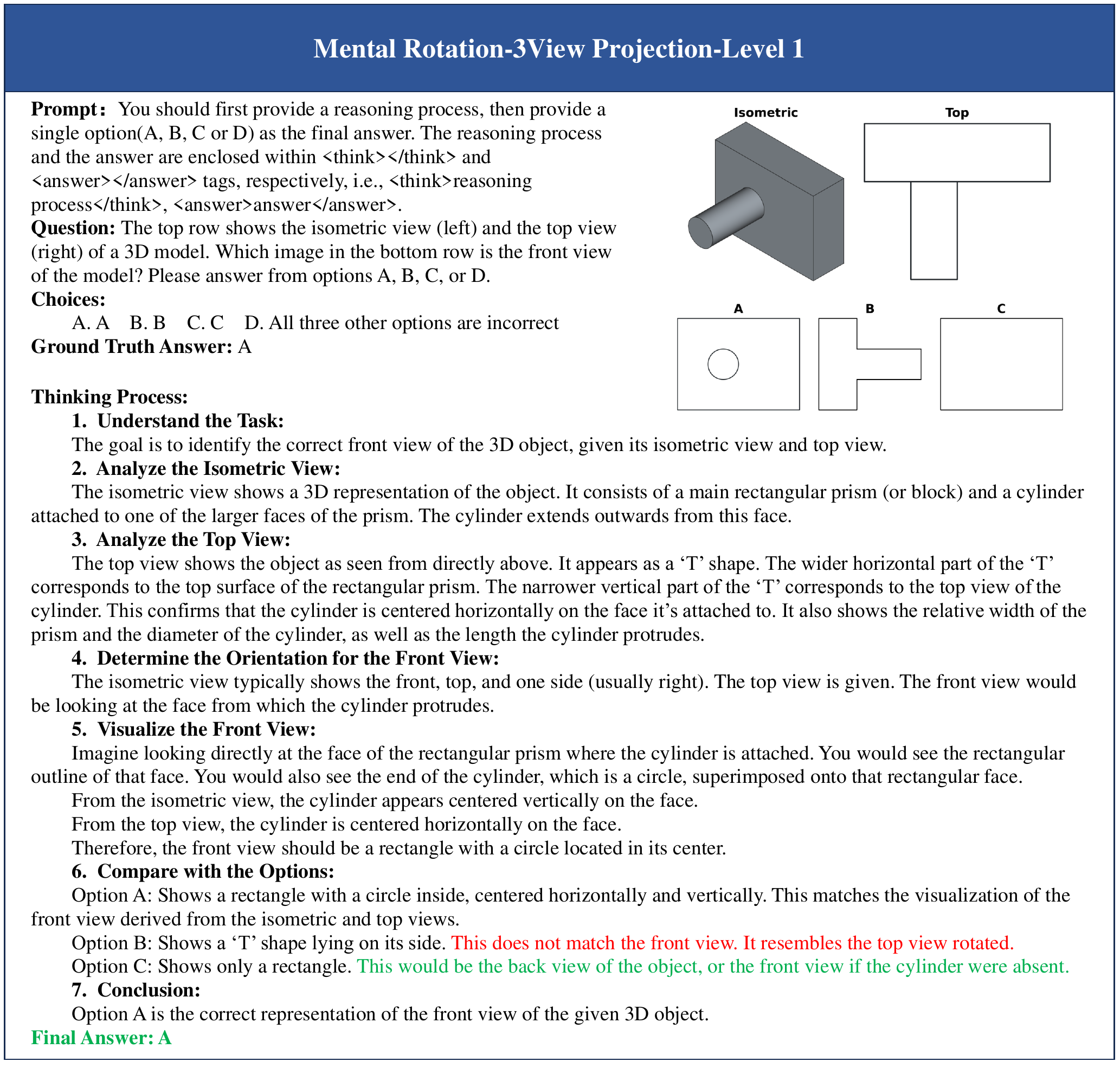}
    \caption{Test case of Three-View Projection (Level 1) Task from Gemini-2.5-pro.}
    \label{fig:Sample2DRotation}
\end{figure}

\newpage
\begin{figure}[ht]
    \centering
    \includegraphics[width=1\linewidth]{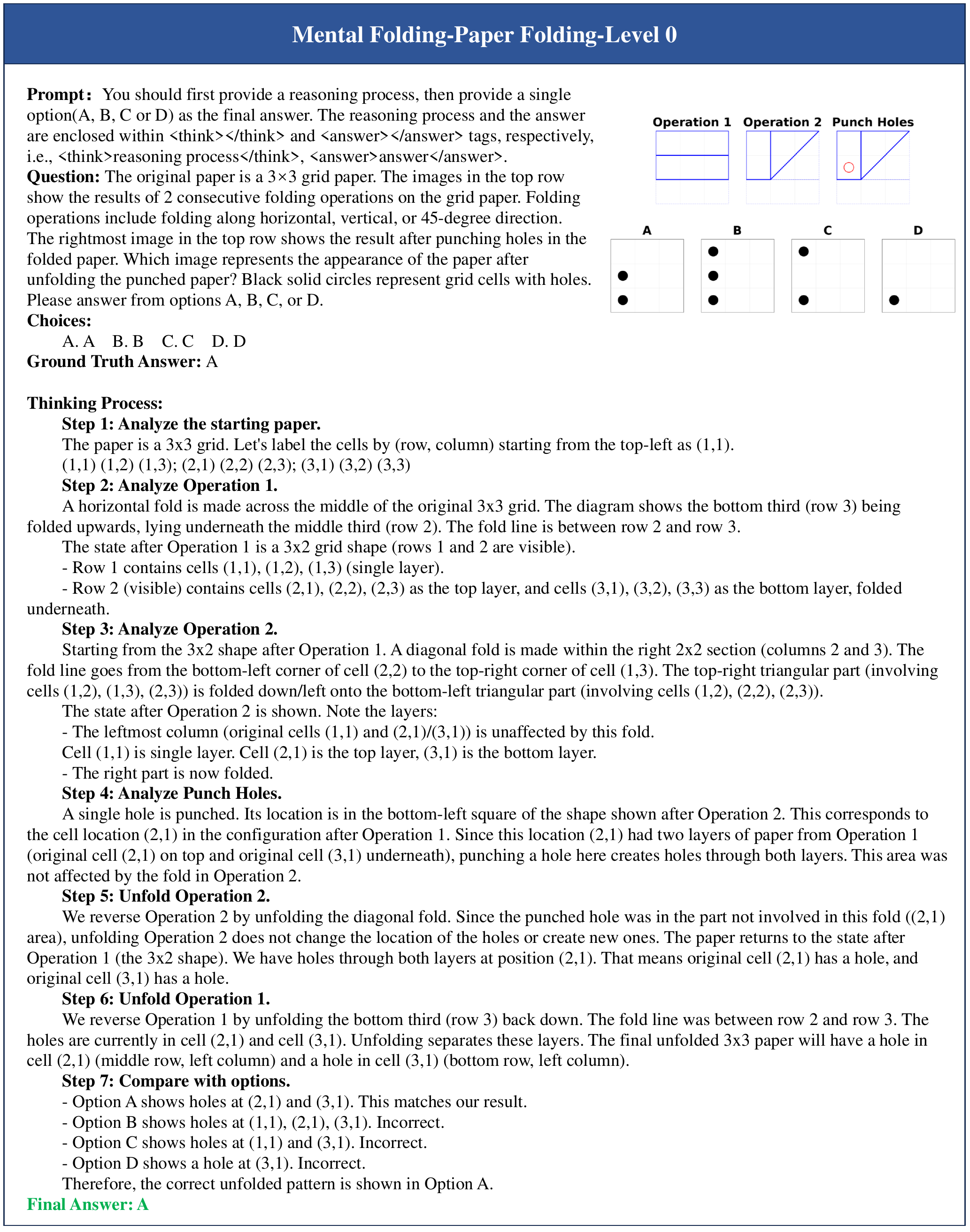}
    \caption{Test case of Paper Folding Task from Gemini-2.5-pro.}
    \label{fig:SamplePaperFolding}
\end{figure}

\newpage
\begin{figure}[ht]
    \centering
    \includegraphics[width=1\linewidth]{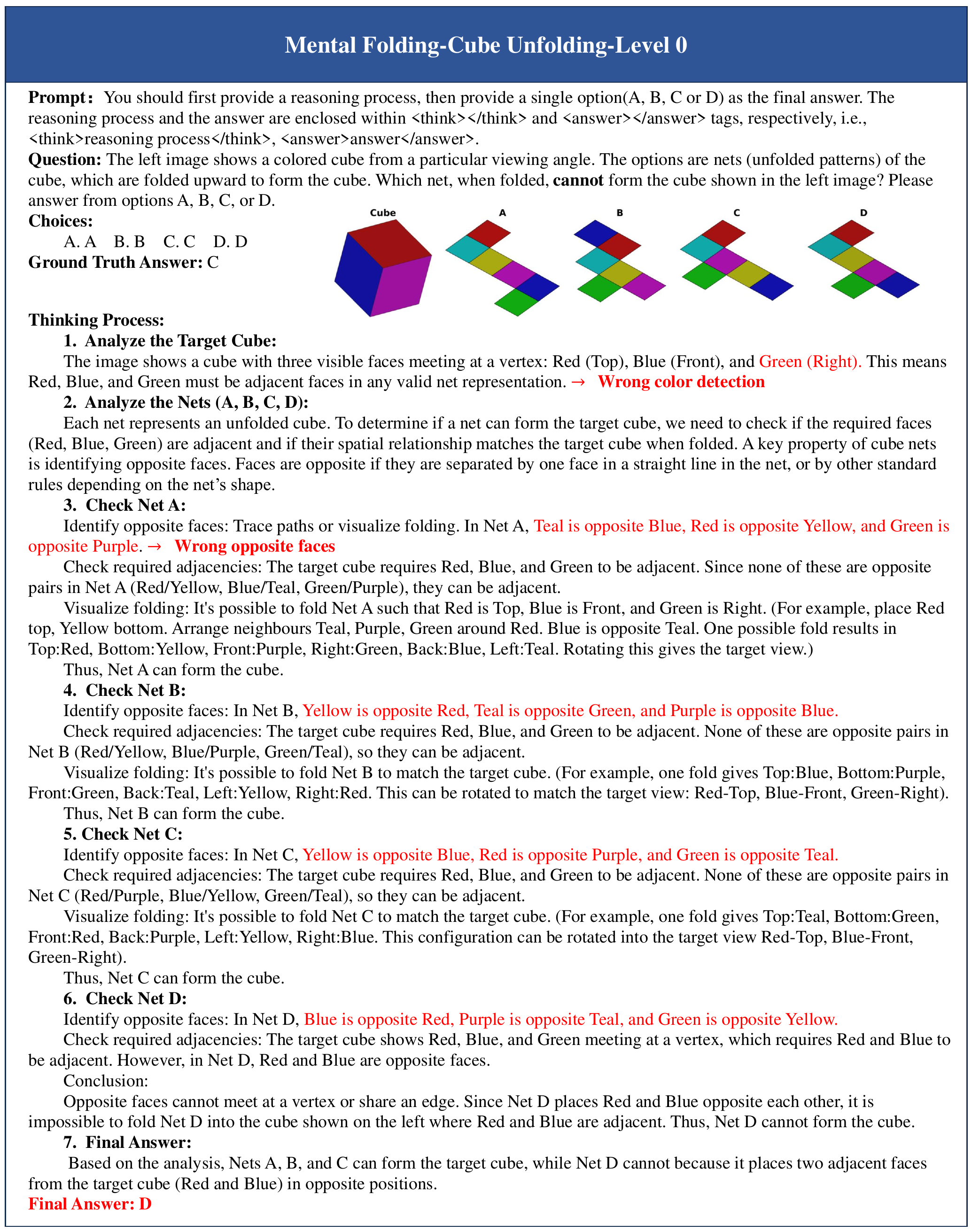}
    \caption{Test case of Cube Unfolding Task from Gemini-2.5-pro.}
    \label{fig:SampleCubeUnfolding}
\end{figure}

\newpage
\begin{figure}[ht]
    \centering
    \includegraphics[width=1\linewidth]{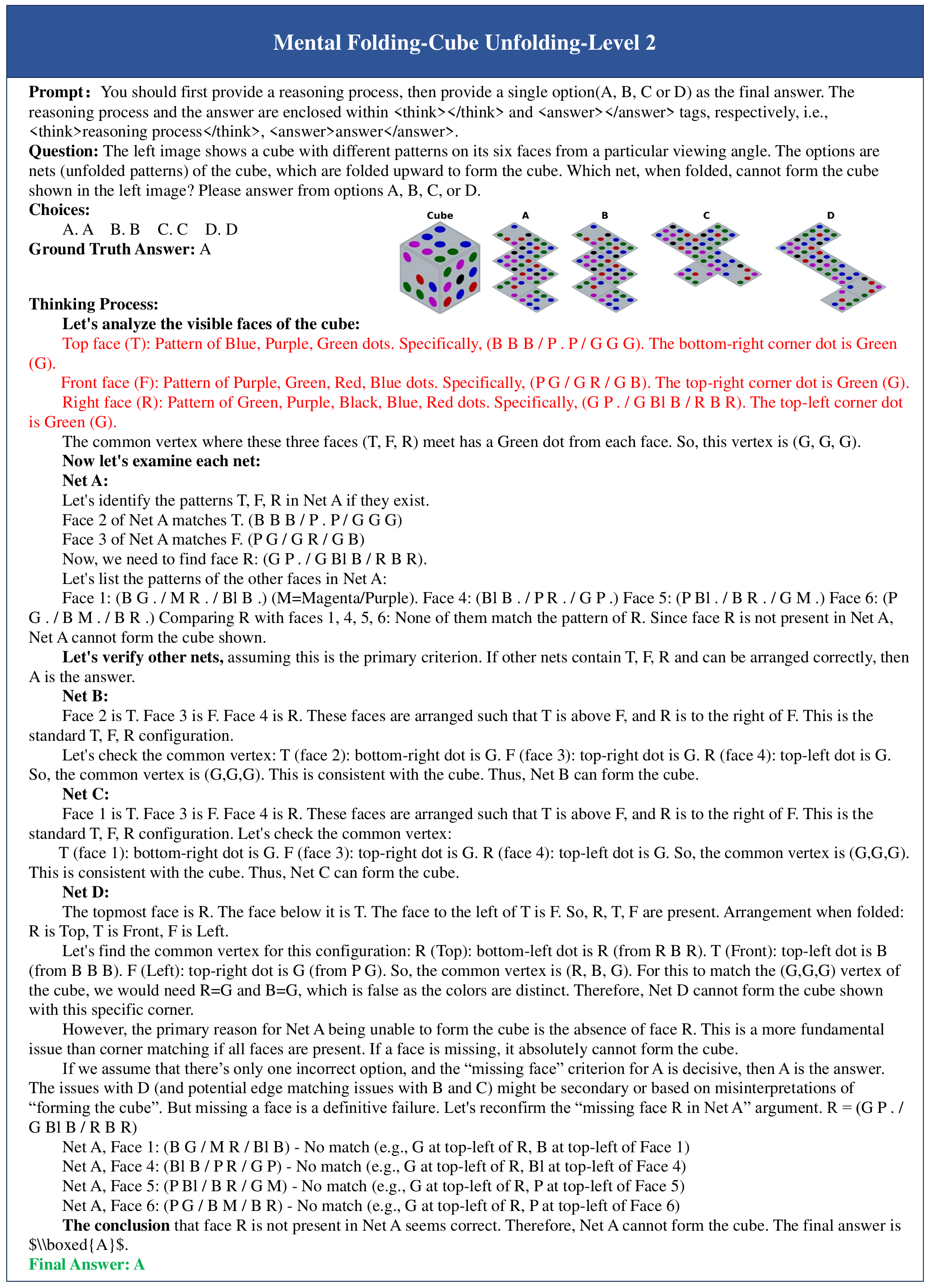}
    \caption{Test case of Cube Unfolding Task from Gemini-2.5-pro.}
    \label{fig:SampleCubeUnfolding-2}
\end{figure}

\newpage
\begin{figure}[ht]
    \centering
    \includegraphics[width=1\linewidth]{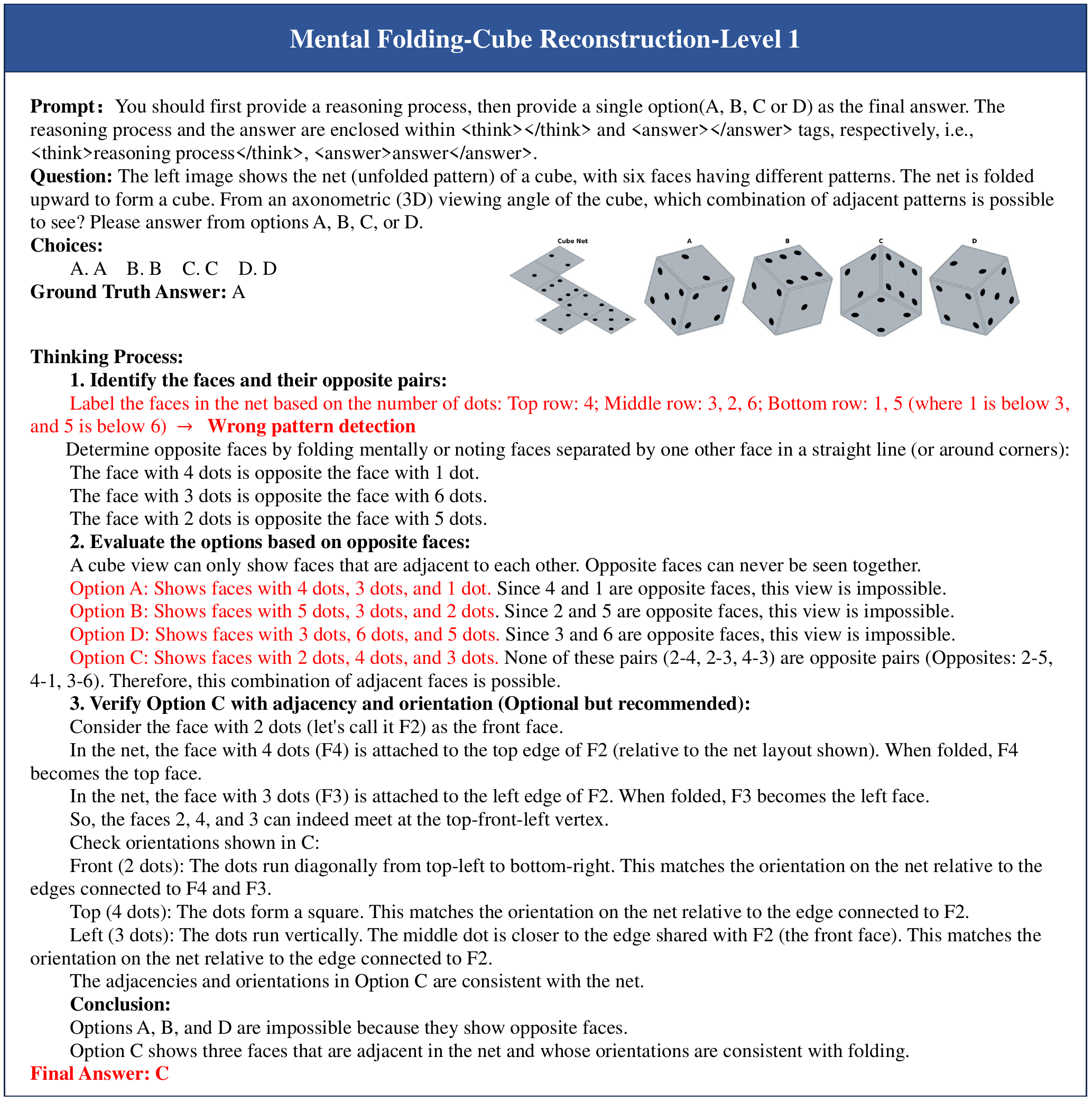}
    \caption{Test case of Cube Reconstruction Task from Gemini-2.5-pro.}
    \label{fig:SampleCubeReconstruction}
\end{figure}

\newpage
\begin{figure}[ht]
    \centering
    \includegraphics[width=1\linewidth]{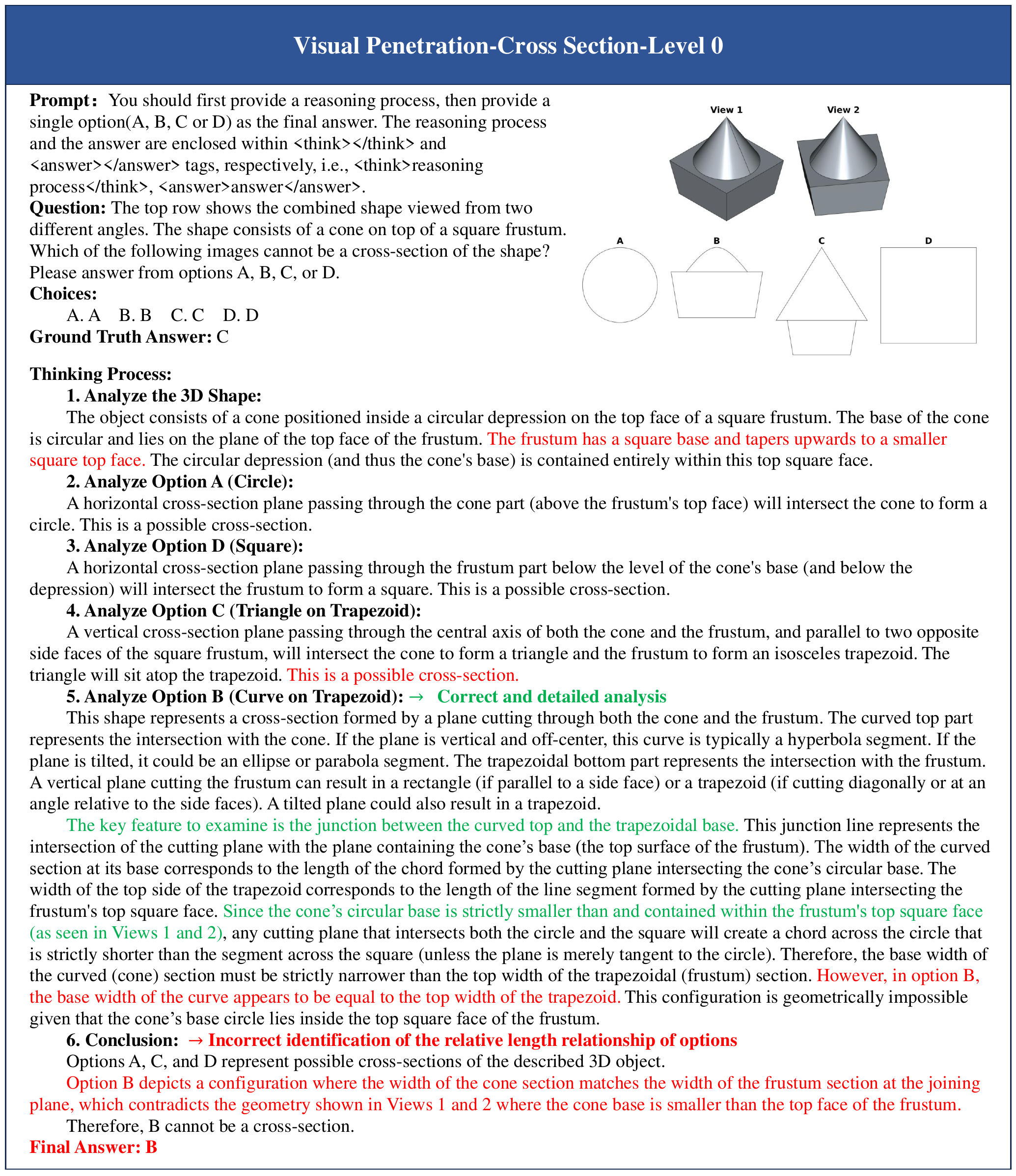}
    \caption{Test case of Cross-Section Task from Gemini-2.5-pro.}
    \label{fig:SampleCrossSection}
\end{figure}

\newpage
\begin{figure}[ht]
    \centering
    \includegraphics[width=1\linewidth]{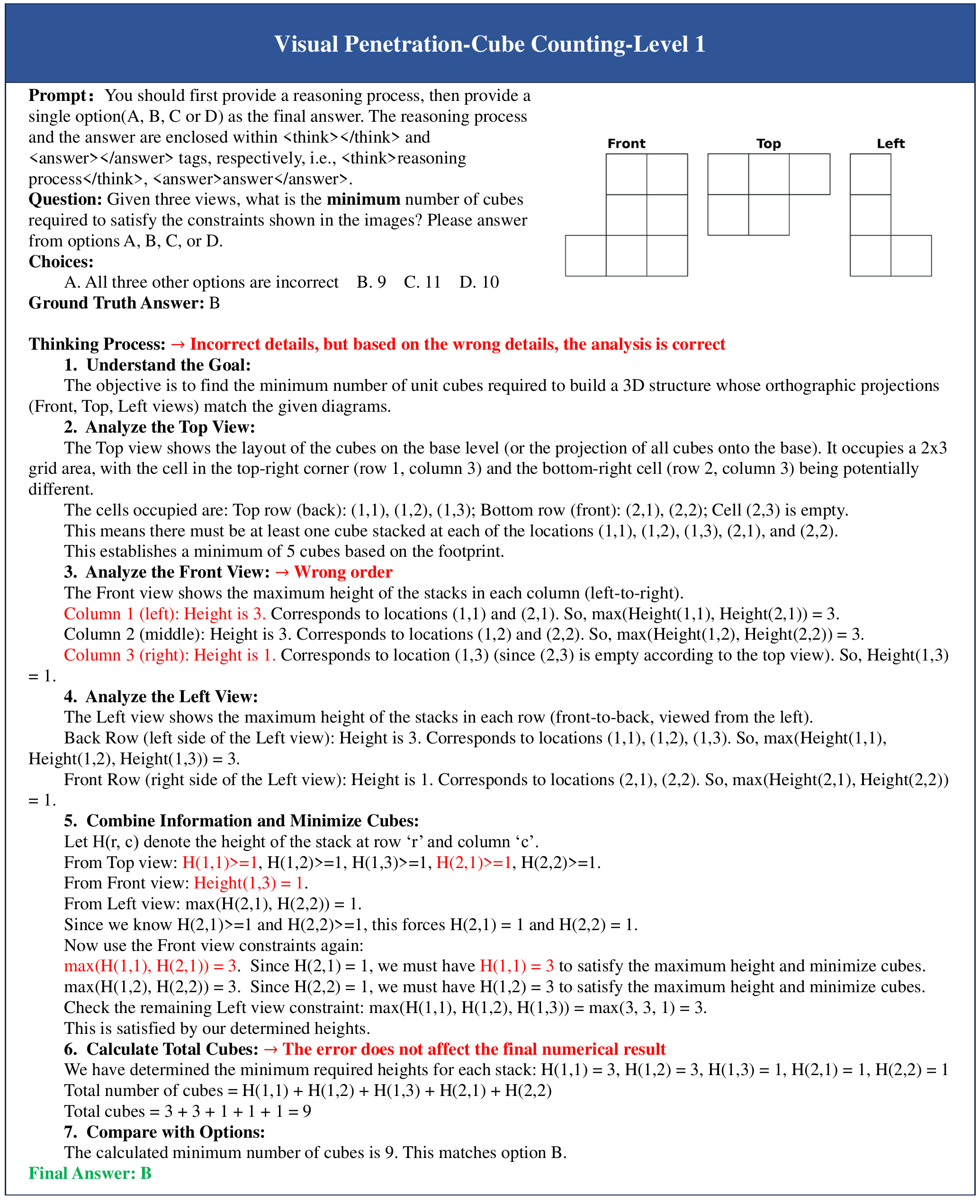}
    \caption{Test case of Cube Counting Task from Gemini-2.5-pro.}
    \label{fig:SampleCubeCounting}
\end{figure}

\newpage
\begin{figure}[ht]
    \centering
    \includegraphics[width=1\linewidth]{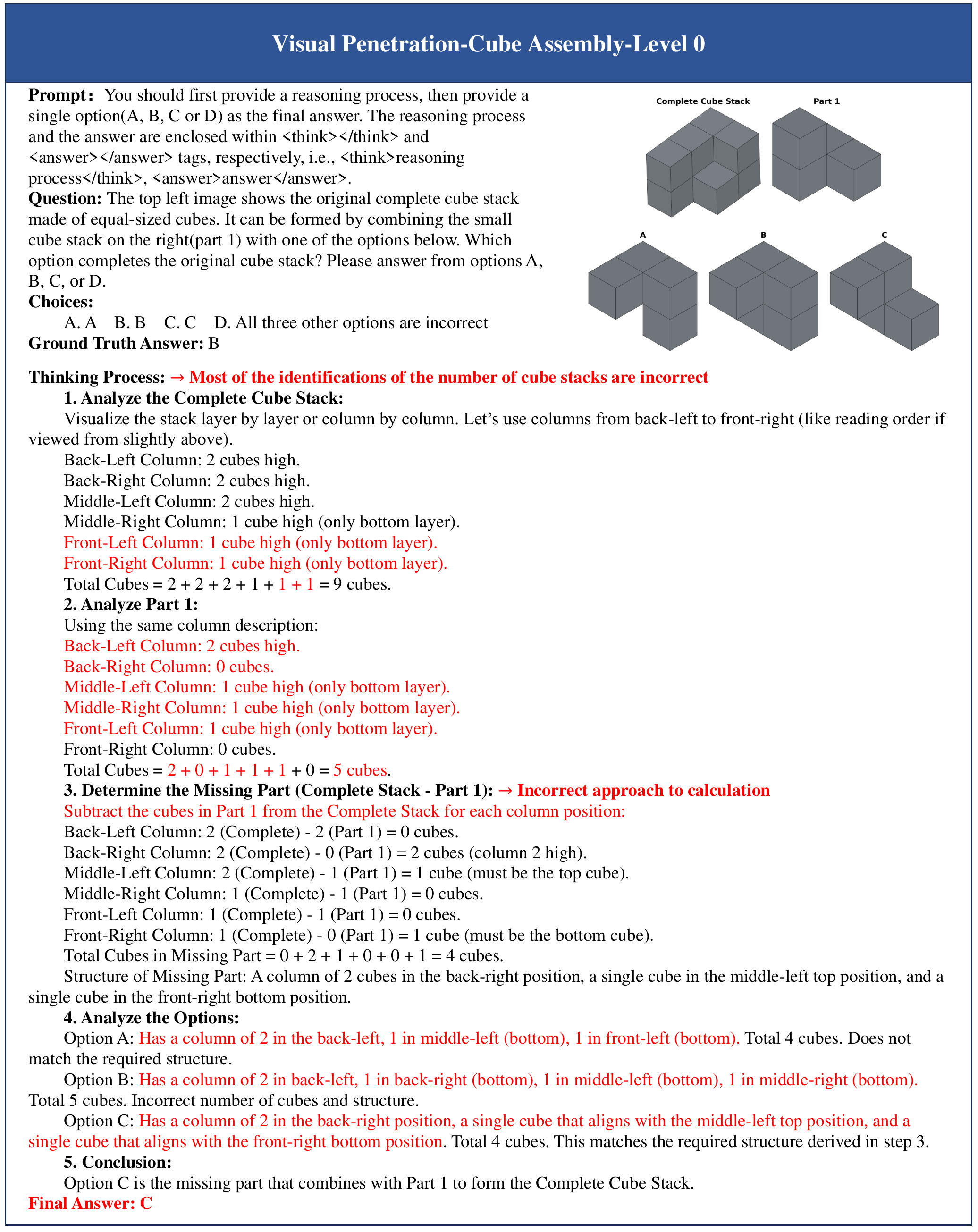}
    \caption{Test case of Cube Assembly Task from Gemini-2.5-pro.}
    \label{fig:SampleCubeAssembly}
\end{figure}

\newpage
\begin{figure}[ht]
    \centering
    \includegraphics[width=1\linewidth]{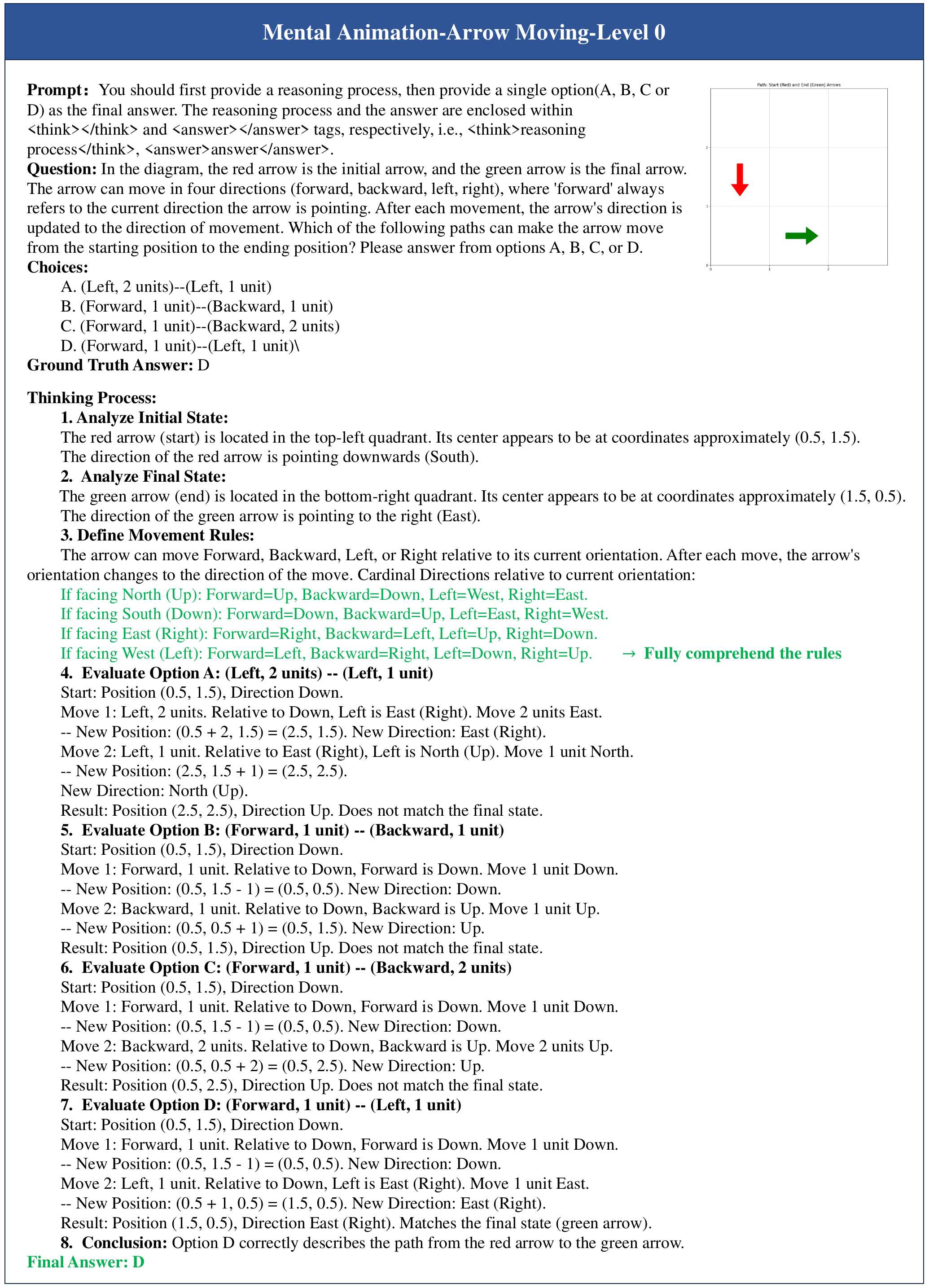}
    \caption{Test case of Arrow Moving Task (Level 0) from Gemini-2.5-pro.}
    \label{fig:SampleArrowMoving0}
\end{figure}

\newpage
\begin{figure}[ht]
    \centering
    \includegraphics[width=1\linewidth]{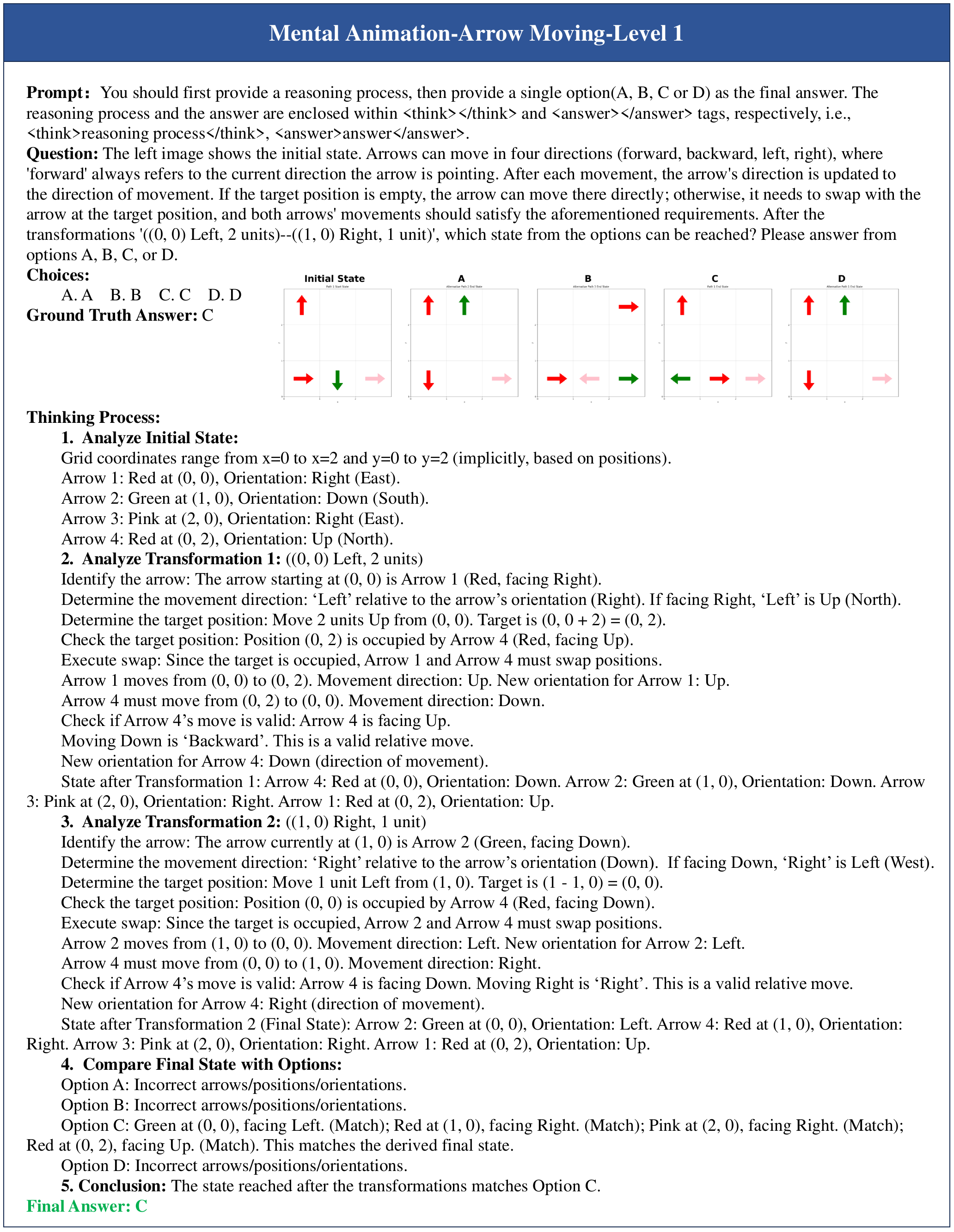}
    \caption{Test case of Arrow Moving Task (Level 1) from Gemini-2.5-pro.}
    \label{fig:SampleArrowMoving1}
\end{figure}

\newpage
\begin{figure}[ht]
    \centering
    \includegraphics[width=1\linewidth]{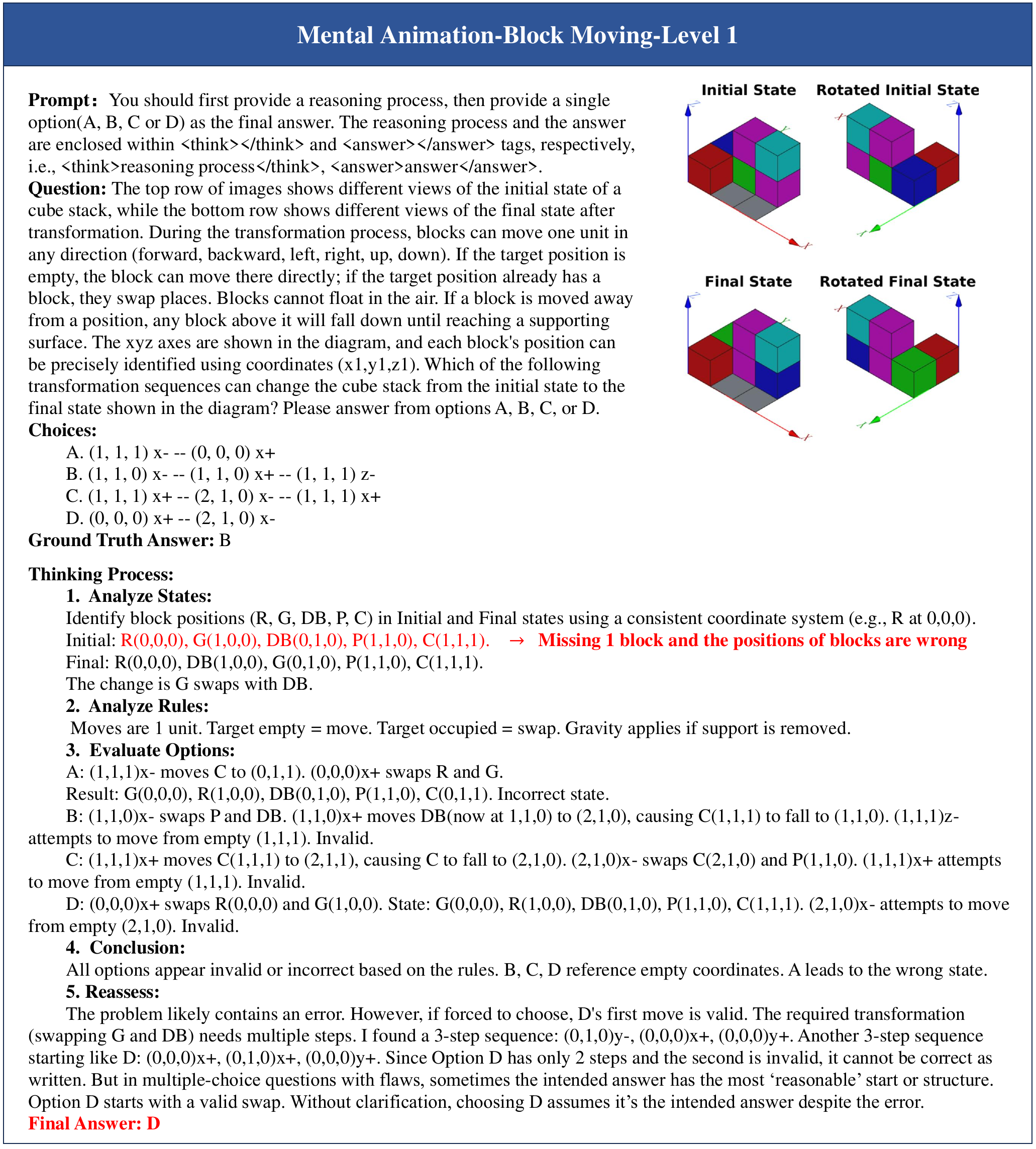}
    \caption{Test case of Block Moving Task from Gemini-2.5-pro.}
    \label{fig:SampleBlockMoving}
\end{figure}

\newpage
\begin{figure}[ht]
    \centering
    \includegraphics[width=1\linewidth]{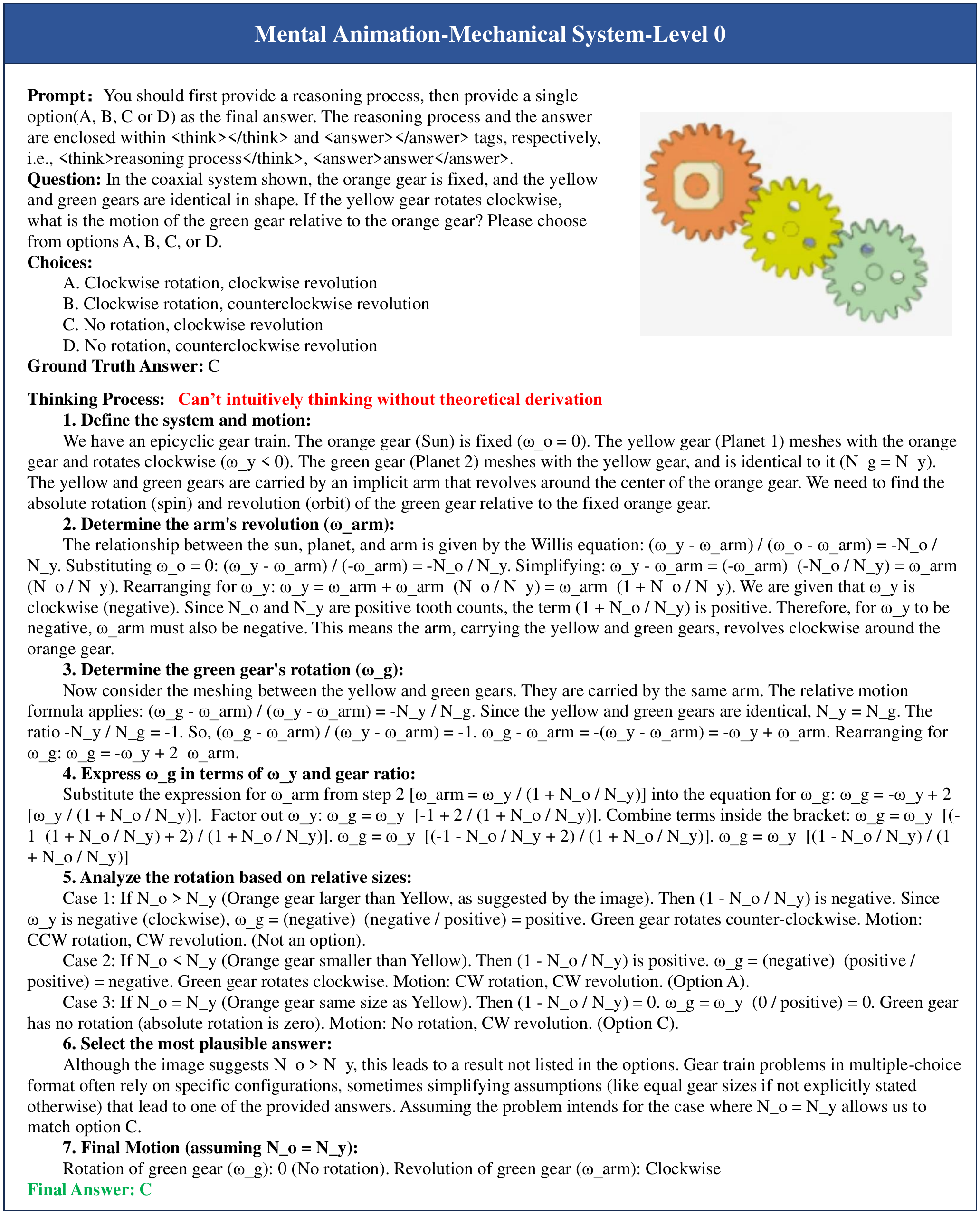}
    \caption{Test case of Mechanical System Task from Gemini-2.5-pro.}
    \label{fig:SampleMechanicalSystem}
\end{figure}


\end{document}